\documentclass[journal]{IEEEtran}
\pdfoutput=1 
\usepackage{times}
\usepackage{soul}
\usepackage{url}
\usepackage[hidelinks]{hyperref}
\usepackage[utf8]{inputenc}
\usepackage[small]{caption}
\usepackage{graphicx}
\usepackage{subfigure}
\usepackage{amsmath}
\usepackage{amsthm}
\usepackage{booktabs}
\usepackage{algorithm}
\usepackage{algorithmic}
\usepackage{color}
\usepackage{amssymb}
\usepackage{diagbox}
\usepackage{framed,multirow}

\begin{document}
\title{Label Distribution Amendment with Emotional Semantic Correlations for Facial Expression Recognition}
\author{Shasha~Mao,~%~\IEEEmembership{Member,~IEEE,}
        Guanghui~Shi,~% ~Boxin~Shi,~%~\IEEEmembership{Member,~IEEE,}
        Licheng~Jiao,~%~\IEEEmembership{Fellow,~IEEE,}
        Shuiping~Gou,~%~\IEEEmembership{Member,~IEEE,}
        Yangyang~Li,~%~\IEEEmembership{Member,~IEEE,}
        Lin~Xiong,~and~{Boxin~Shi}% <-this % stops a space
\thanks{S. Mao, G. Shi, L. Jiao, S. Gou and Y. Li are with Key Laboratory of Intelligent Perception and Image Understanding of Ministry of Education, School of Artificial Intelligence, Xidian University, Xi’an, China, e-mail: ssmao,lchjiao,shpgou,yyli@xidian.edu.cn.}% <-this % stops a space 
\thanks{X. Lin is with JD Finance America Corporation, 675 E Middlefield Rd, Mountain View, CA 94043, USA, e-mail: Lin.Xiong@jd.com.}% <-this % stops a space
\thanks{B. Shi is with National Engineering Laboratory for Video Technology, Department of Computer Science and Technology, Peking University, Beijing, China, e-mail: shiboxin@pku.edu.cn.}
%\thanks{Manuscript received April 19, 2005; revised August 26, 2015.}
}

% The paper headers
\markboth{Journal of \LaTeX\ Class Files, July~2015}%
{Mao \MakeLowercase{\textit{et al.}}: Label Distribution Amendment with Emotional Semantic Correlations for Facial Expression Recognition}

\maketitle

\begin{abstract}
By utilizing label distribution learning, a probability distribution is assigned for a facial image to express a compound emotion, which effectively improves the problem of label uncertainties and noises occurred in one-hot labels. In practice, it is observed that correlations among emotions are inherently different, such as surprised and happy emotions are more possibly synchronized than surprised and neutral. It indicates the correlation may be crucial for obtaining a reliable label distribution. 
Based on this, we propose a new method that amends the label distribution of each facial image by leveraging correlations among expressions in the semantic space. 
Inspired by inherently diverse correlations among word2vecs, the topological information among facial expressions is firstly explored in the semantic space, and each image is embedded into the semantic space. Specially, a class-relation graph is constructed to transfer the semantic correlation among expressions into the task space. By comparing semantic and task class-relation graphs of each image, the confidence of its label distribution is evaluated.  Based on the confidence, the label distribution is amended by enhancing samples with higher confidence and weakening samples with lower confidence. 
Experimental results demonstrate the proposed method is more effective than compared state-of-the-art methods. 
\end{abstract}


\IEEEpeerreviewmaketitle

\section{Introduction}
\IEEEPARstart{F}{acial} expression is one of signals that directly communicate humman's emotions and behaviors, and some latent emotions may be analyzed based on facial expressions. Therefore, facial expression recognition (FER) has attracted many attentions in the past decades, and many studies \cite{buck1974sex, smith1996spontaneous, corneanu2016survey, liu2015inspired, acharya2018CovPool} have been proposed to improve the perfromance of FER. Early, the studies focused on some lab-collected expression datasets, such as CK+ \cite{CK},  MMI \cite{MMI}, JAFFE \cite{lyons1998japanese}, Oulu-CASIA \cite{zhao2011facial}, et al, where facial expression images are generally collected with the single emotion produced factitiously. With increasing the size of facial expression datasets, some wild expresssion datasets are collected based on spontaneous emotions, such as  RAF-DB \cite{RAF-DB}, AffectNet \cite{AffectNet}, et al, and meanwhile many deep FER methods \cite{Liu_2014_CVPR,li2018occlusion,wang2018two,lo2021facial,li2020deep} have been proposed. Noticeably, in most of FER studies, the one-hot label is generally annotated as the catergary of emotions for each facial image, considering that it is tough to manually annotate multiple labels.

%%%%%%%%%%%%%%%%%%%%%%%%%%% 
\begin{figure}
	\centering
	\includegraphics[width=1\columnwidth]{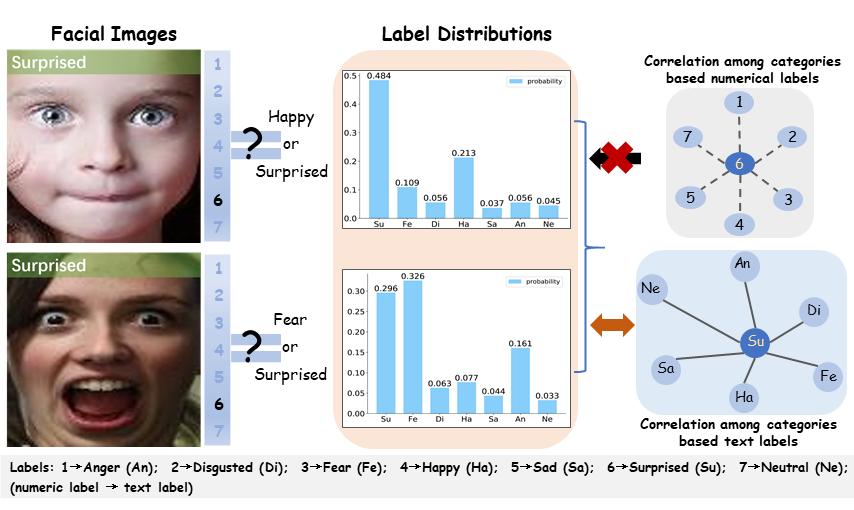}
	\caption{An illustration of the problem existing in the one-hot label and the correlation among emotions, where both facial images belong to a same category (surprised) from RAF-DB. Differently from numerical labels, the correlation among text labels is analysable and diverse, which is beneficial for generating reliable label distributions. 
	}
	\label{fig-1}
\end{figure}
%%%%%%%%%%%%%%%%%%%%%%%%%%% 

However, some facial expressions are compound even ambiguous rather than simply prototypical in practice, since human's emotion is psychologically complicated and various \cite{2000neuropsychology,li2019blended,chen2020labelAuxiliary}. 
It means that the one-hot label is insufficient to precisely express the compound emotion emerging in a cheek. Whereas, the emotions appearing in cheeks are generally ambiguous or complicated in the wild expression data \cite{zhou2015emotion, chen2020labelAuxiliary}, and Fig. \ref{fig-1} shows a simple illustration. 
In Fig.~\ref{fig-1}, two facial images are from the $6^{th}$ category (Surprised) of the wild expression data (RAF-DB), and they are given a one-hot label with the numerical value $6$. But, it is obviously observed that the emotions of two images are more possibly ambiguous and compound, rather than the single emotion (Surprised).  
For the first image, its expression may be the mixture of surprised and happy emotions, and the expression may be the mixture of surprised and fear for the second image. 
It illustrates that a label distribution is more appropriate than the given one-hot label for FER data. 
Moreover, the label noise widely exists in FER databases due to incorrect or imprecise annotations \cite{zhou2015emotion,zeng2018facial}. If we consider from the view of noisy labels, the given one-hot labels of two images may be more likely noisy ones. To address this problem, label distribution learning (LDL) \cite{xu2018label,wang2019ldl} is introduced into FER to assign a probability distribution for each facial image instead of one-hot labels. Many related studies \cite{chen2020labelAuxiliary,zhou2015emotion,barsoum2016training,jia2019facial,fu2020semantic} have demonstrated the label distribution is effective for presenting label uncertainties and noises in FER.      

From human's emotions, it is observed that some emotions are intrinsically connected with each other. For instances, the surprised expression is possibly appeared while happening a happy thing, and the angry expression is possibly occurred while happening a disgusted thing. As shown in Fig. \ref{fig-1}, for the first image, its label distribution implies that she surprised because she might be happy, where the label distribution is given by the proposed method.  
It indicates that there are intrinsic correlations between different emotions in FER, since some emotions are causally generated. 
Except the mentioned connection, there are also irrelevant emotions, for examples, the happy expression is less possibly occurred with the angry or sadness emotion, and surprised and neutral emotions also difficultly appear in one facial action. 

Based on the above analyses, it indicates that the intension of correlations (connection or irrelevance) among different emotions are inherent and meanwhile diverse. Conceivably, these diversities of intrinsic correlations among expressions are significant for insuring the reliability of label distributions. 
However, in most studies \cite{chen2020labelAuxiliary,zhou2015emotion,jia2019facial}, the label distribution is only utilized to achieve the conversion of the one-hot label to a vector with seven or eight values corresponding to expression categories, and the intrinsic correlation among expressions was barely focused or analyzed in generating the label distribution of expressions.  
Additionally, it is worth noting that the correlation among expressions is unable to be analyzed based on the numerical labels, since the information of numeric values is limited.  
As shown in the top right corner of Fig.~\ref{fig-1}, the correlations between the value $6$ (denoting the $6^{th}$ category) and others are equal and negligible, representing by dotted lines. 

Interestingly, we find that the correlation among different emotions can be explored based on the text information corresponding to each expression, differently from numerical labels.
As shown in the bottom right corner of Fig.~\ref{fig-1}, the correlations between surprised and other categories are different based on the analysis for text labels, and more detailed analyses are found in the following section (shown in Fig.~\ref{inter-class}).  
Obviously, when expression categories are represented by corresponding words instead of numerical values, their correlations are analysable and different.  
Specially, considering the characteristic that the correlation among emotions is inherent, it means the correlation is consistent between different feature spaces. 
Inspired by this, we consider exploring the topological information among different emotions with text labels and expecting that the semantic correlation among expressions is able to assist the learning for FER with numerical labels. 

Hence, we propose a new FER method that leverages {\it S}emantic {\it C}relations among {\it E}xpressions to {\it A}mend the {\it L}abel {\it D}istribution of facial expression, named by SCE-ALD. In SCE-ALD, the word2vecs corresponding to expression categories is firstly utilized to explore the topological information among expressions and construct the semantic space, and each image is embedded into the semantic space to attain its semantic feature.  Considering the consistency of correlations among expressions in different spaces, the class-relation graph is constructed as the bridge to transfer the semantic correlations among expressions into the task space and meanwhile evaluate the confidence of each sample. 
Based on the obtained confidences, the prototype is regenerated and the label distribution of each sample is amended in the task space. 
Noticeably, the features of samples with higher confidence are boosted and the features of samples with lower confidences are weakened in the training phase of SCE-ALD.  

Compared with the state-of-the-art methods, the contributions of the proposed method are mainly two-fold:
\begin{itemize}
	\item We innovatively pose the problem of diversely intrinsic correlations among emotions and propose SCE-ALD method that modifies label distributions based on semantic correlations among expressions. 
	To our knowledge, this is the first work that leverages text labels to explore the correlation among expressions and assist the learning of FER with numerical labels.  
	
	\item We construct the class-relation graph to achieve the relation transference from the semantic space to the task space. Based on the class-relation graph, the confidence of label distribution is assessed for each sample, and samples with higher confidence are enhanced in feature learning to modify the label distribution.    
\end{itemize} 

The rest of this manuscript is organized as follows. Section \ref{relatedworks} firstly introduces related works about facial expression recognition. Secondly, Section \ref{ProposedModelSection} introduces the detail of the proposed method. Then, experimental results and analyses are demonstrated to validate the performance of the proposed method in Section \ref{Experiments}. Finally, Section \ref{Conclusion} provides the conclusion as well as the prospects on future works.

%%%%%%%%%%%%%%%%%%%%%%% Fig.3
\begin{figure*}[t] 
	\centering
	\includegraphics[width=1\textwidth]{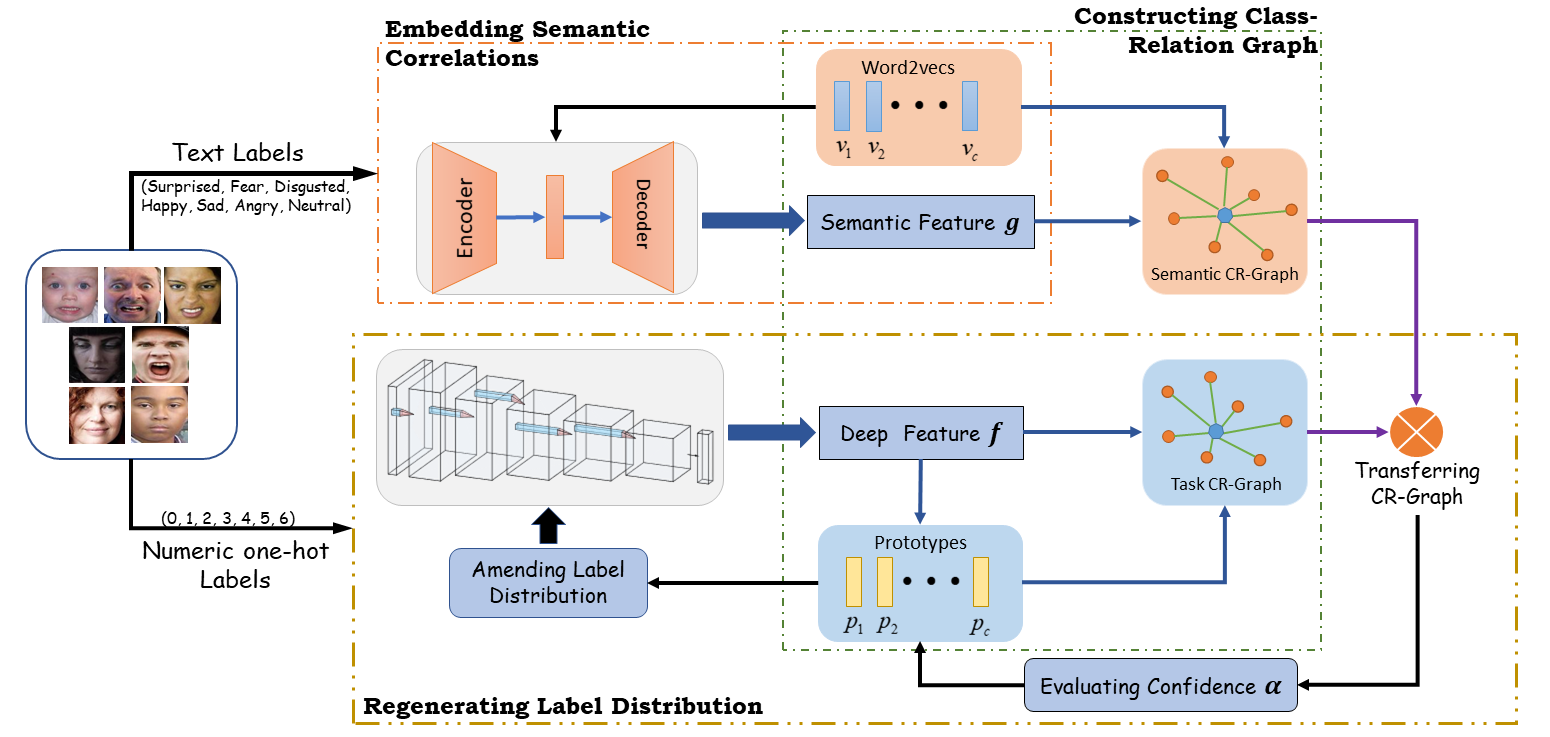}
	\caption{The framework of the proposed method (SCE-ALD).}
	\label{FrameworkProposedModel}
\end{figure*} 
%%%%%%%%%%%%%%%%%%%%%%%

\section{Related Works}\label{relatedworks}
This section reviews related works on FER with label distribution learning and label noise. 
In practice, label ambiguity, label inconsistency and label noise easily exists in real-world data of FER, caused by compound emotions and incorrect annotations. 
To address the problem of label noise, one strategy is to utilize a small clean data assess the quality of labels, such as relabeling data with noise, assigning a label confidence to each sample, or label enhancement, and many researches have been proposed \cite{li2017dis,lee2018cleannet,wang2020SCN}.  
Recently, Wang et al. \cite{wang2020SCN} proposes SCN which suppresses the uncertainties and prevent networks from over-fitting uncertain facial images. SCN introduces a self-attention importance weighting module to capture the contributions of instances and then relabels instances with lower confidences. 

On the other hand, for ambiguity and compound labels, label distribution learning is introduced into FER to assign a probability distribution for facial expression.     
Zhou et al.\cite{zhou2015emotion} introduced an Emotion Distribution Learning to learn the mapping from the expression image to the emotion distribution that has a similar data form to probability distribution. 
IPA2LT \cite{zeng2018facial} employs an inconsistent pseudo annotating framework to solve the inconsistent annotations between different facial expression databases, which takes the label-side problem of FER as an annotating error. 
However, the problem of label distribution is not taken into account in FER where expressions are ambiguous and blended with several basic expressions. 
LDL-ALSG \cite{chen2020labelAuxiliary} leverages the topological information of labels from related but more distinct tasks, such as action unit recognition and facial landmark detection, to learn label distribution in FER. 
In LDL-ALSG, clustering algorithm is used to find similar samples in the auxiliary task space as the neighborhood of each sample, and the similarity between the features of each sample and those of the sample in the neighborhood of the auxiliary task space is added to the optimization loss to solve the annotation inconsistency.   

Although the existing methods \cite{chen2020labelAuxiliary, wang2020SCN} based on LDL have verified the effectiveness of LDL for facial expression recognition, the intrinsic correlation among expressions is not considered in learning label distributions of categories. Differently from the existing methods, the proposed method exploits the correlation among emotions to modify the label distribution of expression categories for each image based on both numerical labels and text labels.     

\section{The Proposed Model}\label{ProposedModelSection}
In this paper, we propose a new FER method that adaptively amends the label distribution of facial images by leveraging the semantic information of emotional categories, shortened by SCE-ALD. 
Fig. \ref{FrameworkProposedModel} shows the framework of SCE-ALD. 
In SCE-ALD, there are three crucial modules: Embedding Semantic Correlations, Constructing Class-Relation Graph, and Regenerating Label Distributions, and the details of three modules will be introduced in the following parts. 

\subsection{Embedding Semantic Correlations} 
In practice, emotions are generally complicated and compound, and the correlation among different emotions should be inherently diverse. For instances, neutral and surprised emotions are scarcely concurring in one facial action, but happy and surprised emotions are possibly concurring. However, diverse correlations among emotions are scarce and toughly analyzed in numerical labels of expressions. 

\begin{figure}[t]
	\centering
	\includegraphics[width=0.75\columnwidth]{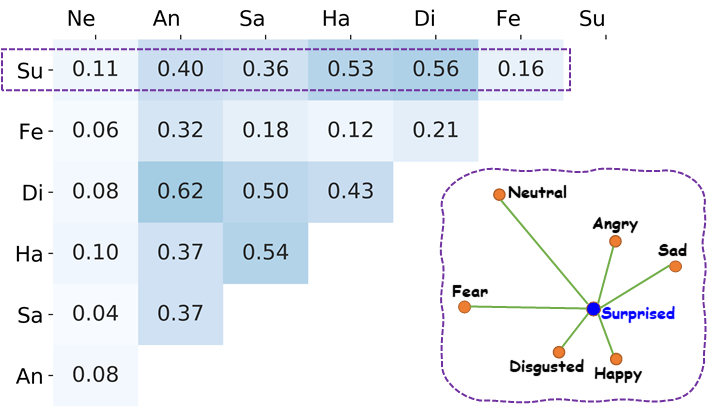}
	\caption{An analysis for similarities among word2vecs of seven expressions, where a the similarities are shown in a matrix with $6\times 7$ is used to exhibit the similarities and each element is the similarity between two expressions. In the right purple dotted box, it gives a visualization of the similarity between surprised and others based on the values in the first row of the similarity matrix. Obviously, the longer the distance between the surprised category and others is, the weaker the correlation between them is.}
	\label{inter-class}
\end{figure}

Differently from the numerical label, we find that the correlations among different words (expressing emotions) are also inherently diverse. A simple analysis of the similarity between two expressions is shown in Fig.~\ref{inter-class}, where the similarity is calculated based on the cosine similarity between seven word vectors obtained by the Google pre-training word2vec model \cite{mikolov2013word2vec}, and seven basic expressions are expressed by Surprised (Su), Fear (Fe), Disgusted (Di), Happy (Ha), Sad (Sa), Angry (An) and Neutral (Ne), respectively.  
As shown in Fig.~\ref{inter-class}, this analysis illustrates that correlations among emotions are obviously different. For examples, the similarity between Su and Ha is higher than others for the category of 'Surprised',  and the similarity between Di and An is higher than others for the category of ‘Disgust’.    
Inspired by this, the module of embedding semantic correlations is designed to explore the topological information among expressions in the semantic space, which embeds facial images into the semantic space based on semantic correlations among expressions. 
In this module, seven words ('surprised', 'fear', 'disgusted', 'happy', 'sad', 'angry' and 'neutral') are applied as text labels of seven expressions.

%%%%%%%%%%%%%%%%%%%%%%% Fig.2
%%%%%%%%%%%%%%%%%%%%%%%

Given the input data $\mathcal{I}$ including $n$ facial images and one-hot labels, $\mathcal{I}=\{({\textbf x}_1,y_1),...,({\textbf x}_n,y_n)\}$, where $y_i\in\{1,...c\}$ is the numerical label of the image ${\textbf x}_i$ and $c$ is the number of expression categories.
In order to construct the topological information among emotions, seven text labels are firstly transformed into word vectors by Google pre-trained word2vec model \cite{mikolov2013word2vec}, represented by $\mathcal{V}=\{{\textbf v}_1,...,{\textbf v}_c\}$ where ${\textbf v}_k \in {\mathbb R}^{1\times 224}$. Based on $\mathcal{V}$, the numerical label of each image is converted as a word2vec. 

Based on $\mathcal{I}$, an Auto-Encoder is trained to embed each sample into the semantic space and meanwhile obtain the feature of each sample in the semantic space. 
In order to explore the topological information among expressions, we add the cosine similarity between the semantic feature ${\textbf g}({\textbf x}_i)$ and the word2vec $V(y_i)$ of the image ${\textbf x}_i$ into the reconstruction loss function, and the loss function is formulated by  
\begin{equation}
	\mathcal{L}_{S} = \frac{1}{N} \sum_{i=0}^{N-1}({\textbf x}_i-\hat{\textbf x}_i)^2 + \gamma\cdot(1 - \textbf{CosS}(V(y_i), {\textbf g}({\textbf x}_i))),
\end{equation}
where ${\hat{\textbf x}_i}$ is the reconstructed image of ${\textbf x}_i$, and $V(y_i)$ expresses the word2vec converted from the numerical label $y_i$. When $y_i = k$, $V(y_i) =\bf v_k$,  $k\in \{1,...,c\}$. 
$\textbf{CosS(a,b)}$ denotes the cosine similarity between vectors $\textbf a$ and $\textbf b$. In Eq.(1), the former is the reconstruction loss to update the parameters of the entire Auto-Encoder, and the latter is the classification loss to update the parameters of the Encoder, where $\gamma$ is the hyperparameter that balances two parts. The former promotes ${\textbf g}({\textbf x}_i)$ to retain the intrinsic information of the image, while the latter ensures that ${\textbf g}({\textbf x}_i)$ can be correctly embedded into its corresponding category in the semantic space. 

\subsection{Constructing Class-Relation Graph} 
In SCE-ALD, the semantic correlation of expressions is introduced into the learning of FER based on numerical labels in the task space. However, the topological information in the semantic space is greatly different from the task space. It means we need construct a bridge between two spaces to transfer the topological information in the semantic space into the task space. 
Inspired by the invariability of relationships, we think the relation between each sample and categories is consistent in two spaces. Based on this, we construct the class-relation graph (CR-Graph) as a connection to convey the semantic information into the task space, and CR-Graph is obtained based on features of each image and expressions categories.  
In the following part, $\mathcal{G}_s$ and $\mathcal{G}_t$ express the semantic CR-Graph and the task CR-Graph, respectively.     

In Fig.\ref{FrameworkProposedModel}, a simple illustration of CR-Graph is shown in the upper right corner. In CR-Graph, the central node is corresponding to a sample ${\textbf x}_i$, and each edge node is corresponding to an expression category. Each edge (linking the central node and a edge node) expresses the similarity between features of ${\textbf x}_i$ and categories of expressions, where the metric of cosine similarity is employed to calculate the similarity in our model. As follows, we introduce the details of the semantic CR-Graph and the task CR-Graph, respectively.  

In the semantic space, the word2vecs ${\mathcal V}$ of expressions categories is regarded as the edge nodes, and the semantic feature ${\textbf g}({\textbf x}_i)$ is regarded as the central node, where ${\mathcal V}$ and ${\textbf g}({\textbf x}_i)$ are obtained by the module of Embedding Semantic Correlations.  
Based on ${\mathcal V}$, we calculate the cosine similarity between the feature ${\textbf g}({\textbf x}_i)$ and all word2vecs to build edges of the CR-graph. 
Then, the semantic CR-Graph $\mathcal{G}_s$ is formulated as a class-instance similarity vector $\mathcal{S}_s \in \mathbb{R}^c$ by   
\begin{equation}
	\mathcal{S}_{ki}^s =  \frac{{\textbf v}_k^T \cdot {\textbf g}({\textbf x}_i)} {{\lVert {\textbf v}_k \rVert}_2 {\lVert {\textbf g}({\textbf x}_i) \rVert}_2}, 
\end{equation} 
where $\mathcal{S}_{ki}^s$ is regarded as the similarity between the word2vec ${\textbf v}_k$ and the semantic feature ${\textbf g}({\textbf x}_i)$, which is the length of the edge liking the central node with the edge node. Larger $\mathcal{S}_{ki}^s$ indicates which category ${\textbf x}_i$ is more likely to fall into.        

Similarly to the semantic CR-Graph, the task CR-Graph $\mathcal{G}_t$ is constructed based on prototypes of expressions categories and the features obtained in the task space, and the similarity $\mathcal{S}_t \in \mathbb{R}^c$ is calculated by the formula 
\begin{equation}
	\mathcal{S}_{ki}^t =  \frac{ {{\textbf p}_k}^T \cdot {\textbf f}({\textbf x}_i)} {{\lVert {\textbf p}_k \rVert}_2 {\lVert {\textbf f}({\textbf x}_i) \rVert}_2},
\end{equation}
where ${\textbf p}_k$ is the prototype of the $k$th expression as the edge node, ${\textbf f}({\textbf x}_i)$ expresses the feature of the sample ${\textbf x}_i$ in the task space as the central node, and $\mathcal{S}_{ki}^t$ denotes the similarity between the prototype ${\textbf p}_k$ and the feature ${\textbf f}({\textbf x}_i)$ in the task space. 

\subsection{Regenerating Label Distributions}
The module of Regenerating Label Distributions is the main part of SCE-ALD, focusing on training a deep network with the modified label distributions by transferring the semantic CR-Graph into the task space for FER. 

{\textbf{Formulating Prototypes of Categories}}
To transfer the topological information in the semantic space into the task space, the class-relation graph is proposed. 
But, differently from the semantic space, the prototype of expression categories is not directly given based on numerical labels in the task space. 
Therefore, we design a formula to calculate the prototype of each category based on deep features in the task space, formulated by 
\begin{equation}
	{\textbf p}_{k} = \frac{1}{n_k} \sum\nolimits_{i=1}^{n_k} \alpha_i \cdot {\textbf f}({\textbf x}_{ki}),
\end{equation}
where ${\textbf p}_{k}$ is the prototype corresponding to the $k$th category, and ${\textbf f}({\textbf x}_{ki})$ expresses deep features of the sample ${\textbf x}_i$ that belongs to the $k$th category. $n_k$ is the number of samples which belong to the $k$th category. 
$\alpha_i$ is the weight coefficient corresponding to each image, which represents the confidence of label distribution obtained in the task space for each image. 
Noticeably, we set $\alpha_i =1$ at initialization and firstly extract deep features based on a backbone CNN as the initial feature.  

{\textbf{Transferring Semantic CR-Graph}}
Considering the consistency of relations, the CR-Graph is constructed to connect semantic and task spaces. Then, the semantic CR-Graph is transferred as a prior information into the task space to assess the confidence of obtained label distribution for each sample and meanwhile achieve the regeneration of the prototypes. By comparing the semantic CR-Graph with the task CR-Graph, the confidence $\alpha_i$ of ${\textbf x}_i$ is given based on Wasserstein distance by 
\begin{equation}       
	\alpha_i = \frac{1}{Wass(\mathcal{S}_s^i, \mathcal{S}_t^i) + \epsilon}, 
\end{equation} 
where $\mathcal{S}_s^i$ and $\mathcal{S}_t^i$ are calculated by Eqs.(2) and (3), and $\epsilon$ is a small positive constant to prevent the denominator going to zero. 
From Eq.(5), it is known that the similarity between semantic and task CR-Graphs is bigger, the confidence of ${\textbf x}_i$ is higher. 
Combining with Eq.(4), the prototype ${\textbf p}_k$ is regenerated based on the confidence $\alpha$. Noticeably, the features of samples with higher confidences are boosted in regenerating the prototype of expressions in the task space. In contrast, the features of samples with lower confidences are weakened. 

{\textbf{Amending Label Distributions}} 
In the process of training network, the label distribution is iteratively amended with the regeneration of prototypes of categories and deep features of each sample in each epoch.
Based on the prototype ${\textbf p}_k$ and the feature ${\textbf f}({\textbf x}_i)$, the label distribution is calculated by 
\begin{equation}
	l_{ik} = \frac{1}{{\lVert {\textbf f}({\textbf x}_i) - {\textbf p}_k \rVert}^2}, 
\end{equation}   
where $l_{ik}$ expresses the inverse of the distance between ${\textbf p}_k$ and ${\textbf f}({\textbf x}_i)$, which is equivalent to the probability that the sample ${\textbf x}_i$ belongs to the $k$th category. The label distribution ${\textbf l}_i$ is obtained by normalizing $\{l_{i1},...,l_{ic}\}$ for a sample.   

{\textbf{Training with Label Distribution}}   
In SCE-ALD, we use a weighted cross entropy loss for training our model, and the loss function is formulated as 
\begin{equation}
	Loss = \beta \cdot entropy({\textbf f}({\textbf x}_i), y_i) + (1-\beta) \cdot entropy({\textbf f}({\textbf x}_i),{\textbf l}_i), 
\end{equation} 
where $\beta$ is a hyperparameter to balances two losses. In Eq.(7), the former is calculated based on the original one-hot label, and the latter is calculated based on the amended label distribution. The original labels and the amended label distributions are meanwhile considered to optimize the network parameters. By adding the loss of label distribution, SCE-ALD effectively improves the performance of FER with one-hot labels, especially for noisy labels, and more analyses are given in ablation studies. 

\subsection{Implementation Details}        
In SCE-ALD, facial images are detected and aligned by MTCNN \cite{zhang2016MTCNN} and further resized to 224$\times$224 pixels. SCE-ALD is implemented with TensorFlow toolbox and use VGG16 \cite{simonyan2014vgg} as the backbone network in the task space. In the semantic space, the used Auto-Encoder consists of a forward and reverse VGG10. By default, two module are pre-trained on the MS-Celeb-1M face recognition dataset and the features are extracted from its first fully connection layer. Training the Auto-Encoder and the backbone network are conducted separately with one Nvidia 2080ti GPU, and the batch size is set as 32. 
The parameter $\beta$ is empirically set as 0.7, and more analyses for it are found in the ablation study of experiments. The leaning rate is initialized as 0.0001 and the maximum epoch is set as 40. Moreover, the loss of label distribution is included for optimization from the 10th epoch.   

\section{Experiments and Analyses}\label{Experiments}
In this section, we use three wild databases to validate the performance of the proposed method, RAF-DB \cite{RAF-DB}, AffectNet \cite{AffectNet} and FERPlus \cite{FerPlus}, considering that label-side problems easily occur in real-world data, such as label noise, label inconsistency, label ambiguity, etc. Note that lab databases are not experimented, since label-side problems are unconspicuous in lab databases of FER. The details of three used wild databases are shown as follows. 
\begin{itemize}
	\item \textbf{RAF-DB} contains around 30,000 facial images downloaded from the Internet. The facial landmarks are manually annotated by the crowdsourcing method with basic or compound expressions. In experiments, the basic RAF-DB is used and divided 12,271 images as the training set and 3,068 images as the testing set. 
	
	\item \textbf{FERPlus} is extended from FER2013 as used in the ICML 2013 Challenges, which is collected by the Google search engine. It includes 28,709 training images, 3,589 validation images and 3,589 test images, where each image is resized to 48$\times$48 resolution. FERPlus is of eight categories and only seven basic expressions are used for experiments. 
	
	\item \textbf{AffectNet} contains around 400,000 images with ten categories, where each image is annotated by one volunteer. In experiments, we use 287,401 images belonging to six basic emotions and neutral categories, where 283,901 images are selected from its training set as the training set and 3,500 images are selected from its validation set as the testing set. 
\end{itemize} 

\subsection{The Comparison Performance}
In this part, we compare the proposed method (SCE-ALD) with five state-of-the-art methods, where five methods are ACNN \cite{li2018ACNN}, RAN \cite{wang2020RAN}, IPA2LT \cite{zeng2018facial}, LDL-ALSG \cite{chen2020labelAuxiliary} and SCN \cite{wang2020SCN}. % CurriculumNet \cite{guo2018curriculumnet}, 
The details of other compared methods are shown as follows.
\begin{itemize}
	\item \textbf{ACNN} is proposed to detect the occlusion of facial regions and pays attention to the most discriminative regions. ACNN utilizes the information of 24 facial landmark points to select the key regions at the feature level, which attains a good performance and is robust to facial occlusion. 
	% focuses on the most discriminative un-occluded regions with attention mechanism in FER. 
	% A CNN with attention mechanism (ACNN) is proposed to detect the occlusion of facial regions and pay attention to the most discriminative regions. ACNN used the information of 24 facial landmark points to select the key regions at the feature level, which attain a good performance and is robust to facial occlusion. 
	
	\item \textbf{RAN} captures the discriminative facial regions for occlusion and construct a self-attention module and a relation attention module to learn coarse attention weights and refine coarse weights with global context.
	% captures the importance of facial regions for occlusion and pose robust FER.
	% The Region Attention Network (RAN) is proposed to capture the discriminative facial regions for occlusion, which comprised of a feature extraction module, a self-attention module, and a relation attention module.
	% And the later two modules are leveraged to learn coarse attention weights and refine coarse weights with global context.
	
	% {\color{green}{And the CurriculumNet \cite{guo2018curriculumnet} uses the distribution density of data in the feature space to measure its complexity to avoid noisy labels}} 
	\item \textbf{IPA2LT} employs an inconsistent pseudo annotations framework to solve the inconsistent annotations between different facial expression databases. IPA2LT takes the label-side problem of FER as an annotating error, which can be regarded as the noise existing in the label.
	
	\item \textbf{LDL-ALSG} leverages topological information of the labels from related but more distinct tasks, such as action unit recognition and facial landmark detection, to learn label distribution in FER. The similarity between features of each sample and its neighborhood from the auxiliary task space is added to the optimization loss to solve the annotation inconsistency. % Clustering algorithm is used to find similar samples in the auxiliary task space as the neighborhood of each sample. 
	
	\item \textbf{SCN} is proposed to suppress the uncertainties of labels and prevent networks from over-fitting uncertain facial images. SCN introduces the self-attention importance weighting module to capture the contributions of samples for training, which reflects the uncertainty of each sample. By SCN, the samples with low confidence are relabeled by leveraging the relabeling module.
\end{itemize}
Moreover, one baseline method (Baseline) is implemented as the comparison method, and it is equal to one case of SCE-ALD where the label distribution is not used. Noticeably, the first two compared methods (ACNN and RAN) consider the occlusion in the facial images and are not involved into label-side problems, and the last three compared methods (IPA2LT, LDL-ALSG and SCN) mainly focus on label-side problems.  
Tabel \ref{tab1} shows the comparison performance obtained by all methods.   

%%%%%%%%%%%%%%%%%%%%%%%%%%%
\begin{table}% table 1- the compared performance
	\caption{Accuracy(\%) of seven methods on three FER databases.}
	\label{tab1}
	\small 
	\begin{center}
		\begin{tabular}{|l|c|c|c|}
			\hline 
			Methods           & RAF-DB      & AffectNet   & FERPlus     \\
			\hline\hline
			Baseline          & 84.49       & 56.91       & 86.19         \\
			\hline
			ACNN              & 85.07       & 58.78       & -                \\
			RAN               & 86.90       & 59.50       & 87.85         \\
			IPA2LT            & 86.77       & 57.31       & 86.71        \\
			% CurriculumNet     & 84.63       & -           & -            \\
			LDL-ALSG          & 85.53       & 59.35       & -               \\
			SCN               & 87.03       & 60.23       & 88.01         \\
			\hline
			SCE-ALD           & \bf{87.19}  & \bf{60.31}  & \bf{88.59}   \\
			\hline
		\end{tabular}
	\end{center}	
\end{table}
%%%%%%%%%%%%%%%%% 

%%%%%%%%%%%%%%%%%%%%%%%%%%%%%%%%%%%%%%%%%%%%%%%%%%%%%%%%%%%%%%%%%%%%%%%%%%%%%%%%%%%%%%%%
\begin{figure*}[t]
	\centering
	\rule[0.2\baselineskip]{18cm}{1pt}
	\includegraphics[width=0.28\columnwidth]{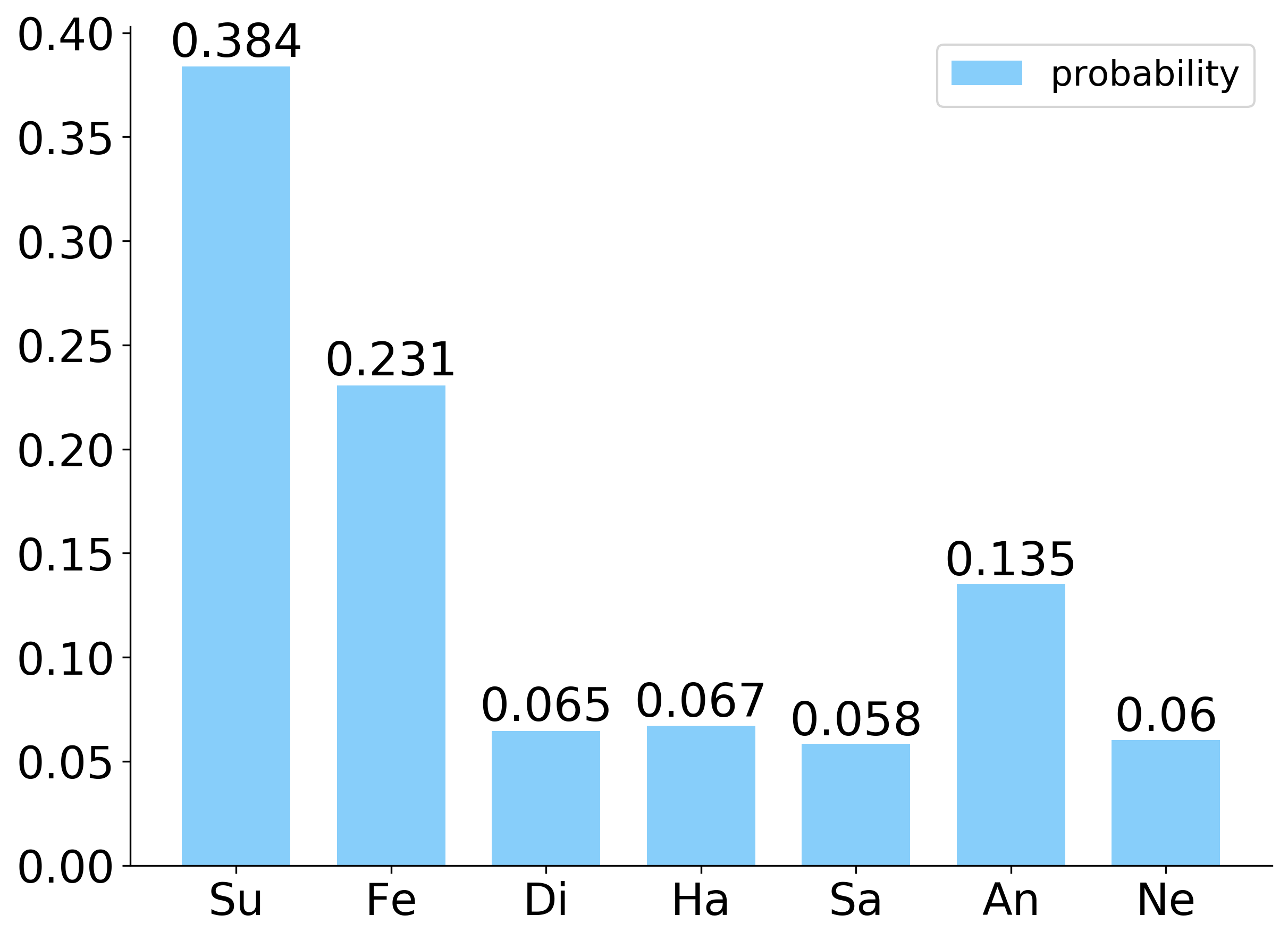}
	\includegraphics[width=0.28\columnwidth]{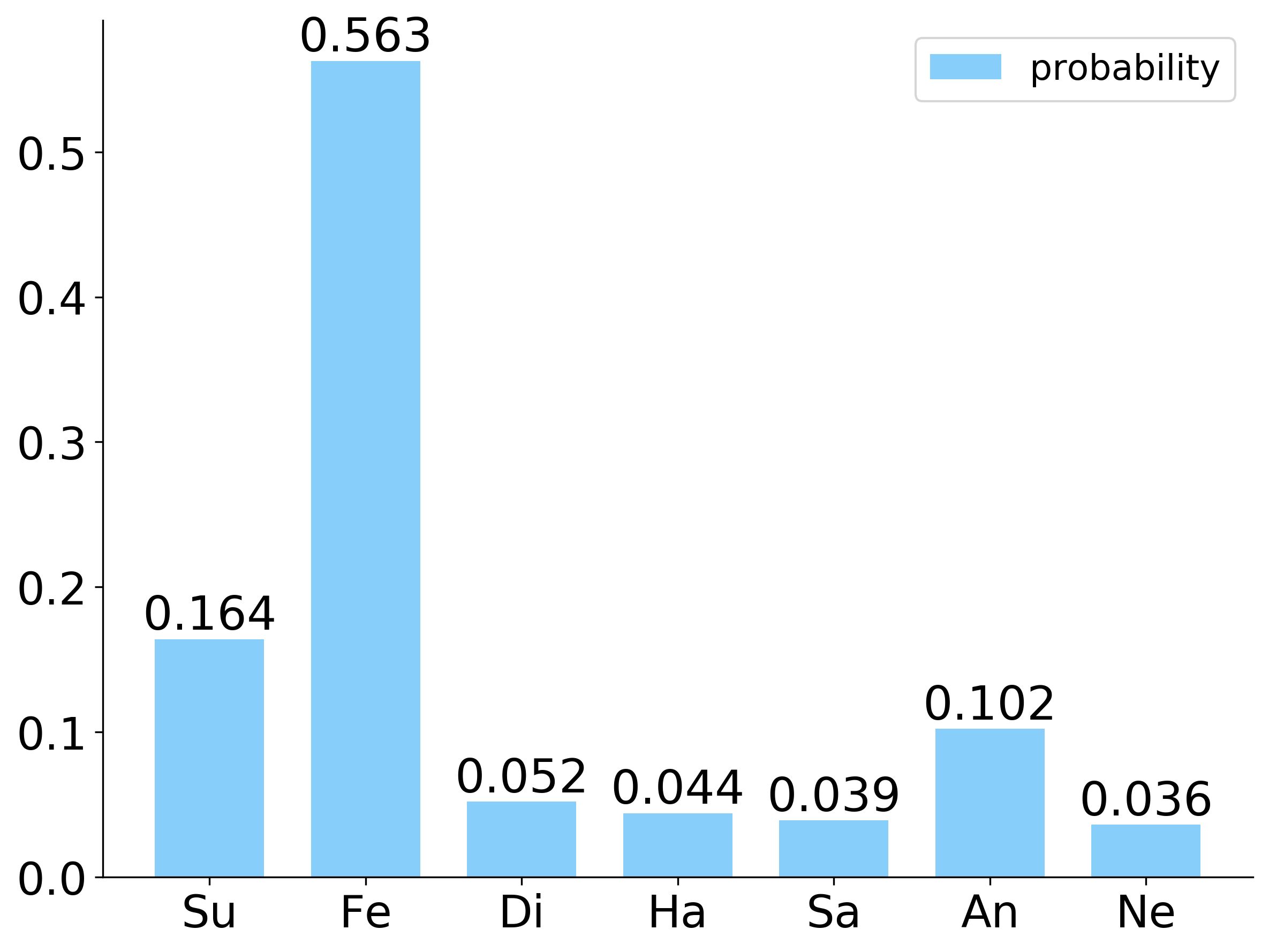}
	\includegraphics[width=0.28\columnwidth]{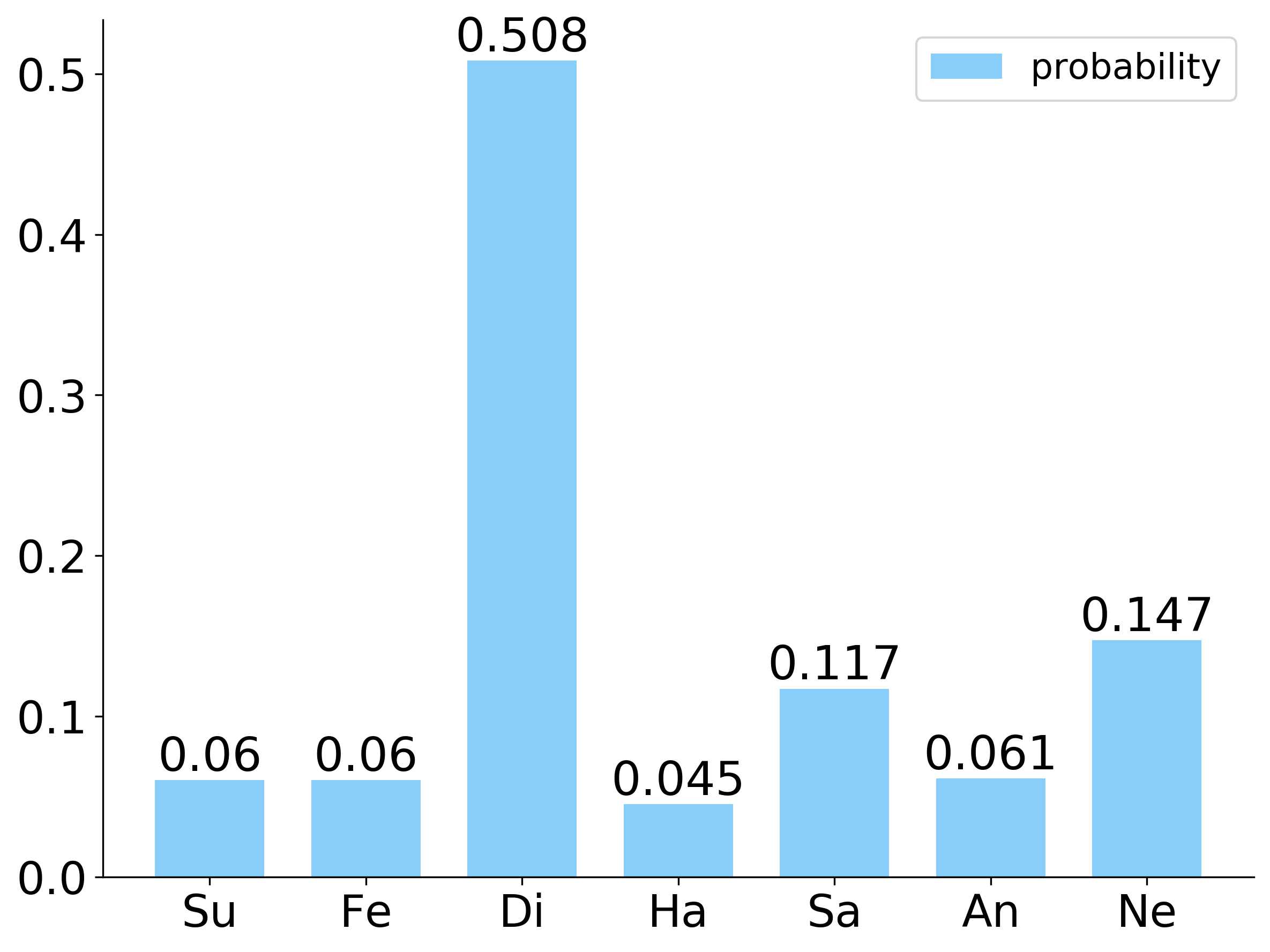}
	\includegraphics[width=0.28\columnwidth]{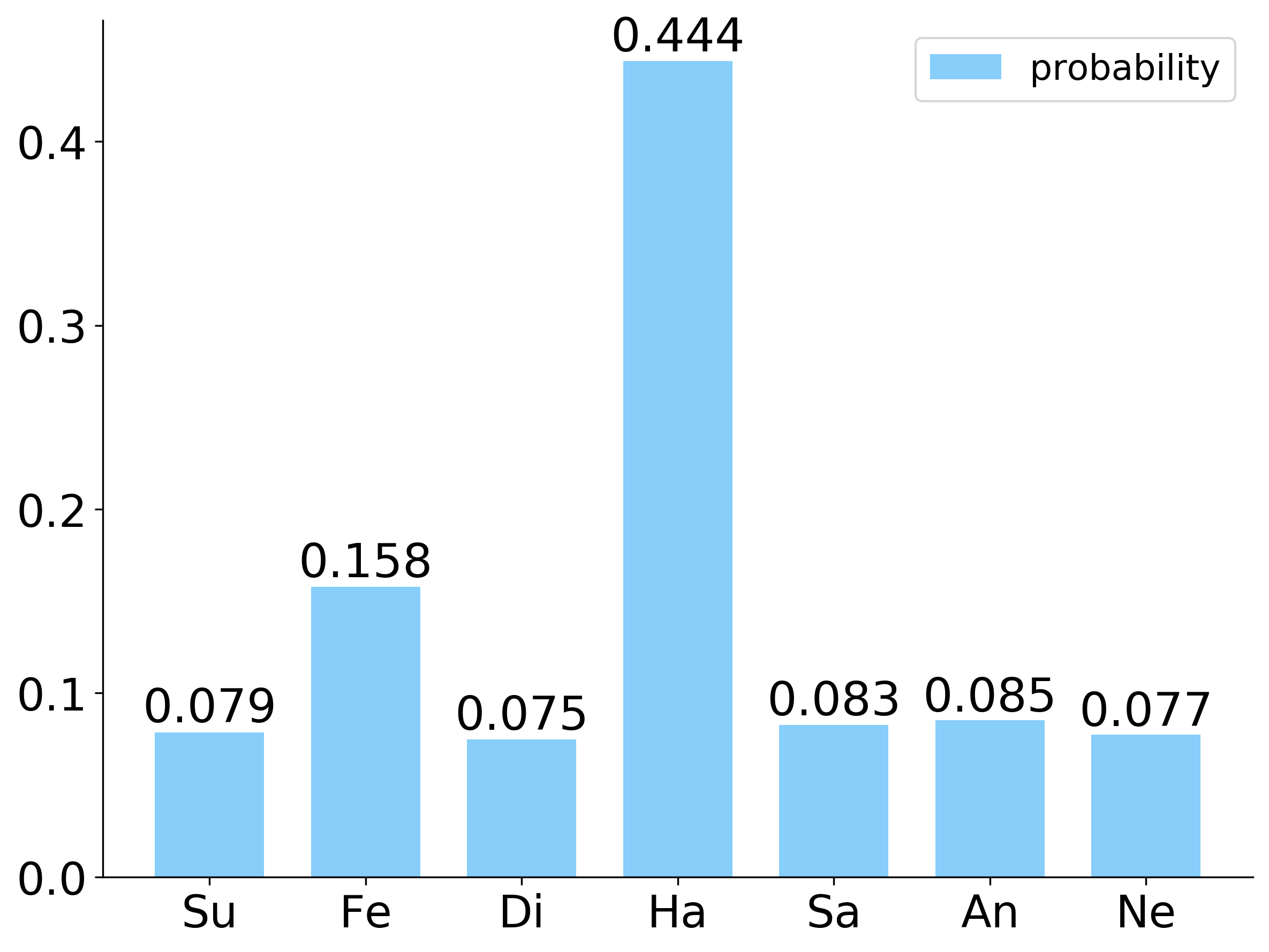}
	\includegraphics[width=0.28\columnwidth]{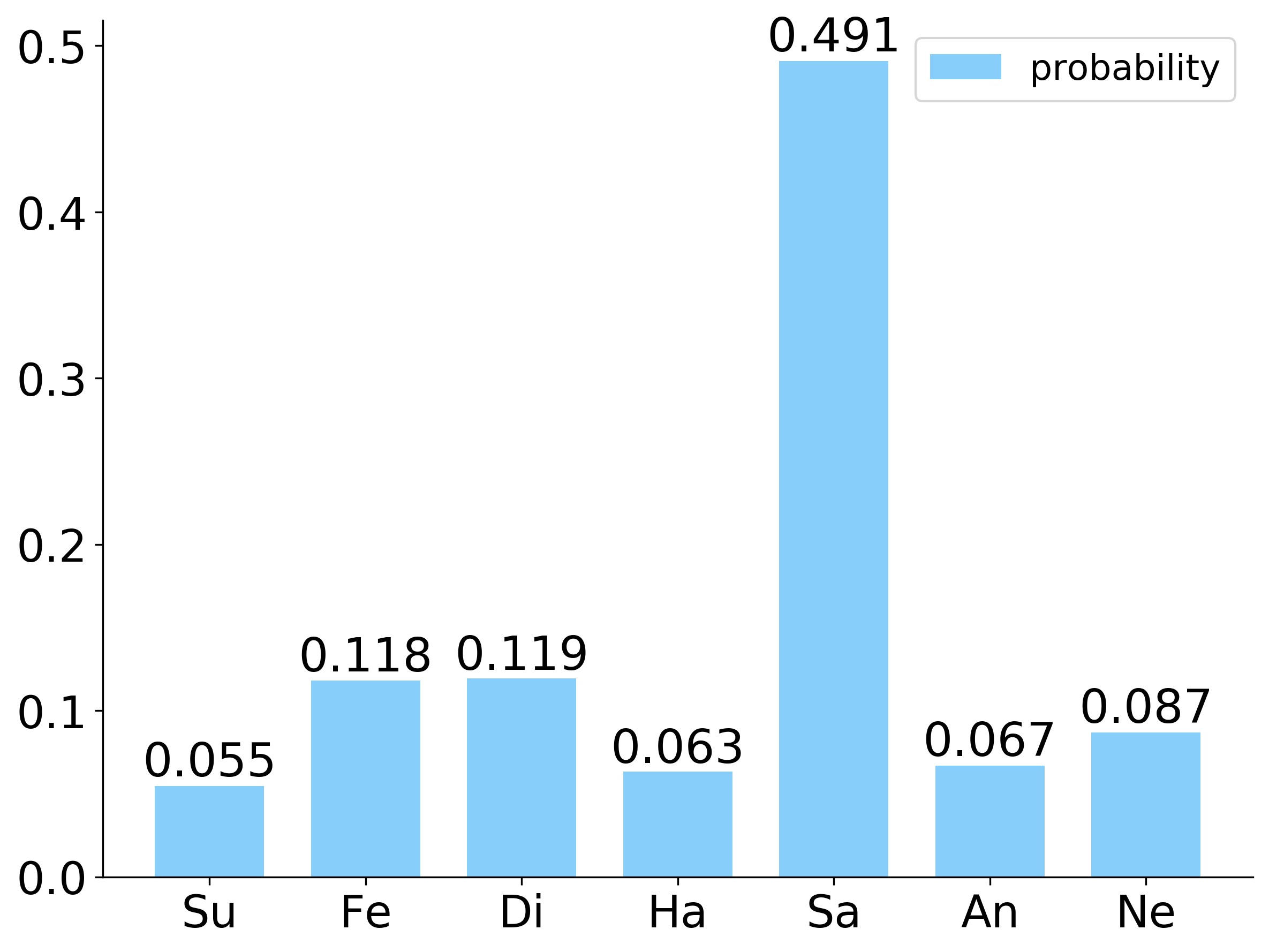}
	\includegraphics[width=0.28\columnwidth]{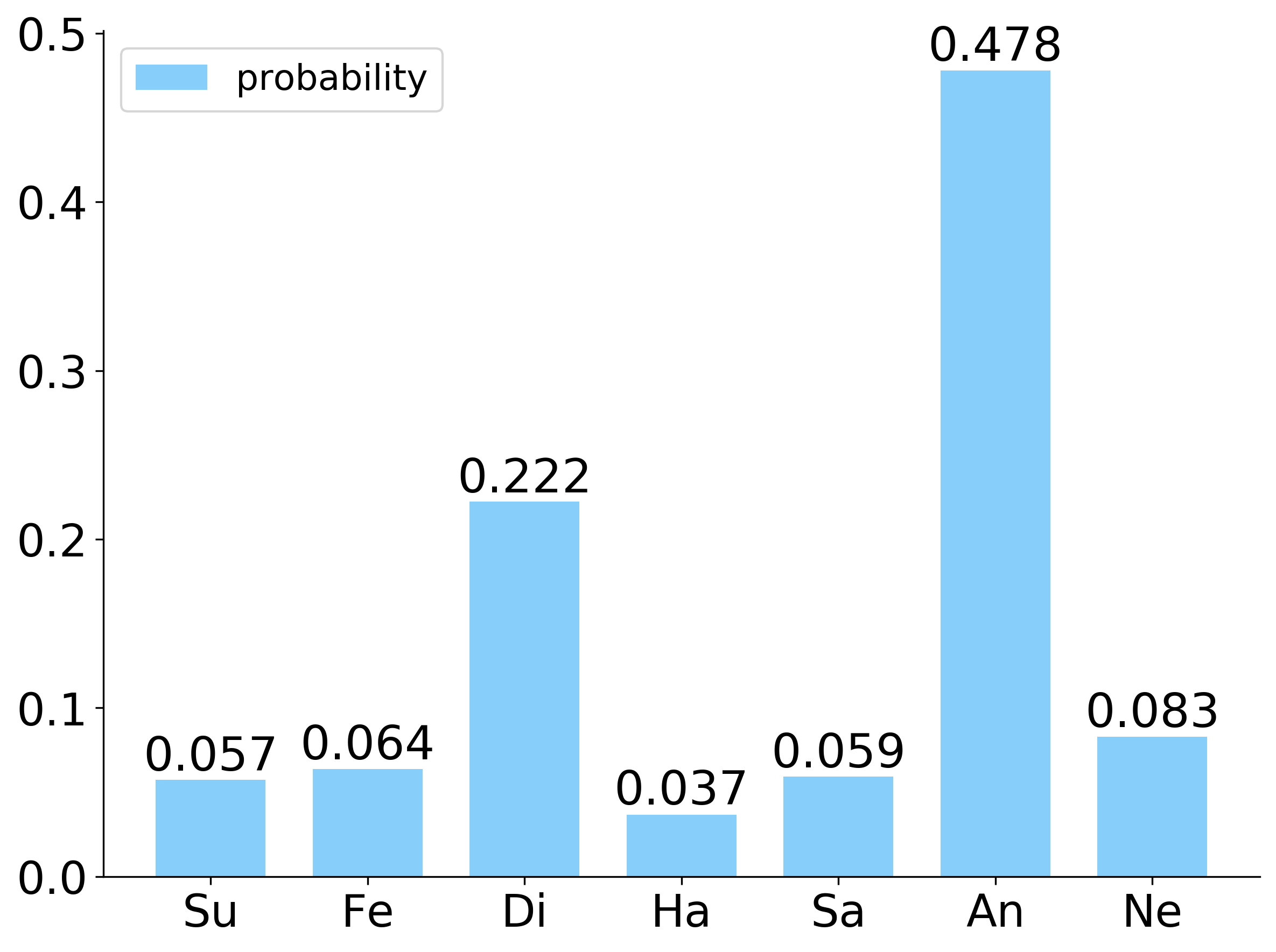}
	\includegraphics[width=0.28\columnwidth]{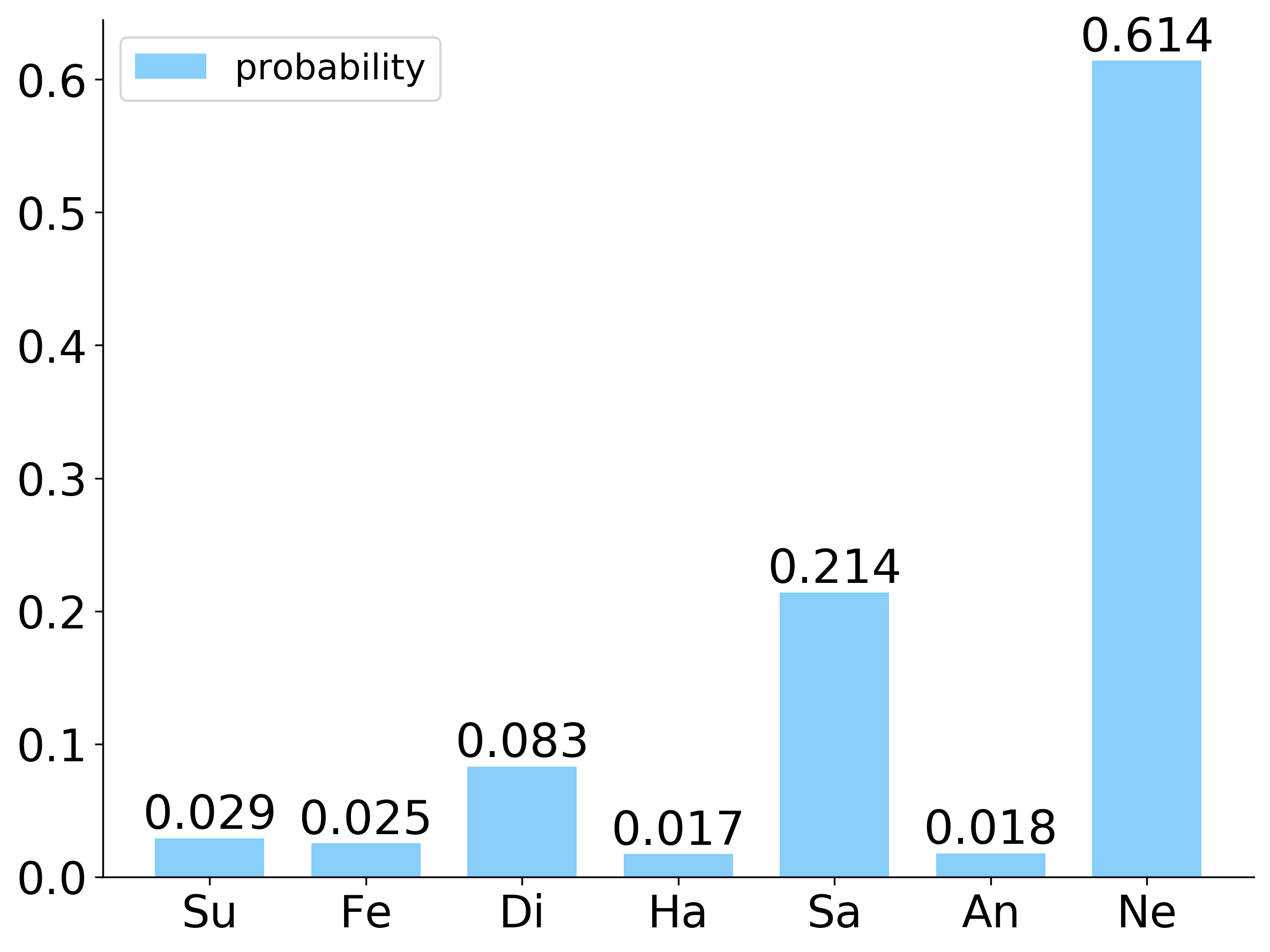}
	\\
	\includegraphics[width=0.28\columnwidth]{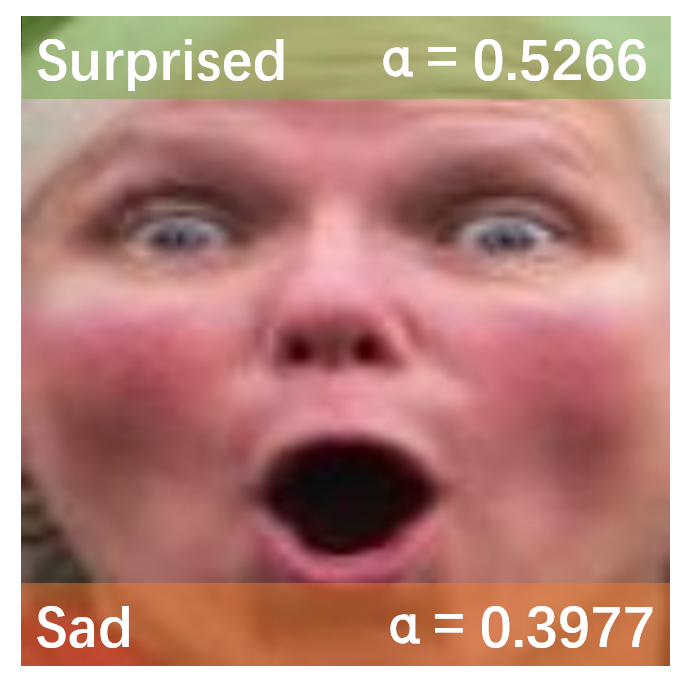}
	\includegraphics[width=0.28\columnwidth]{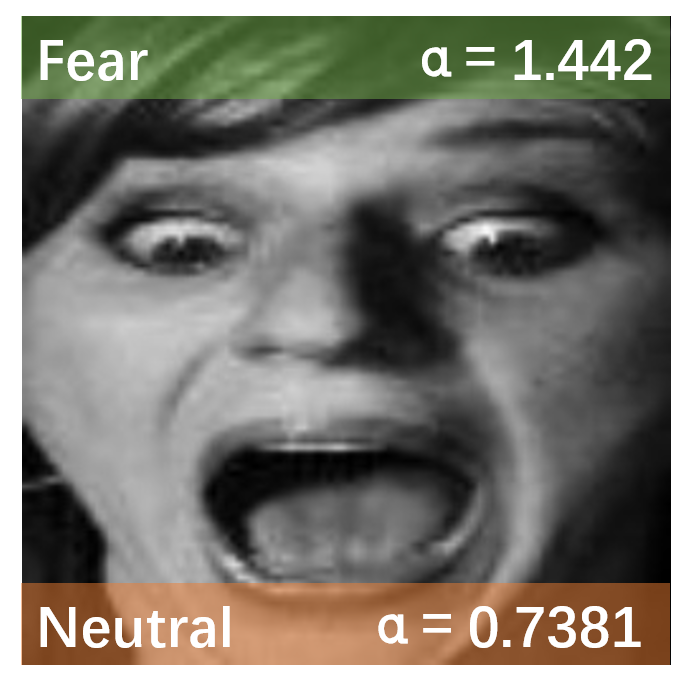}
	\includegraphics[width=0.28\columnwidth]{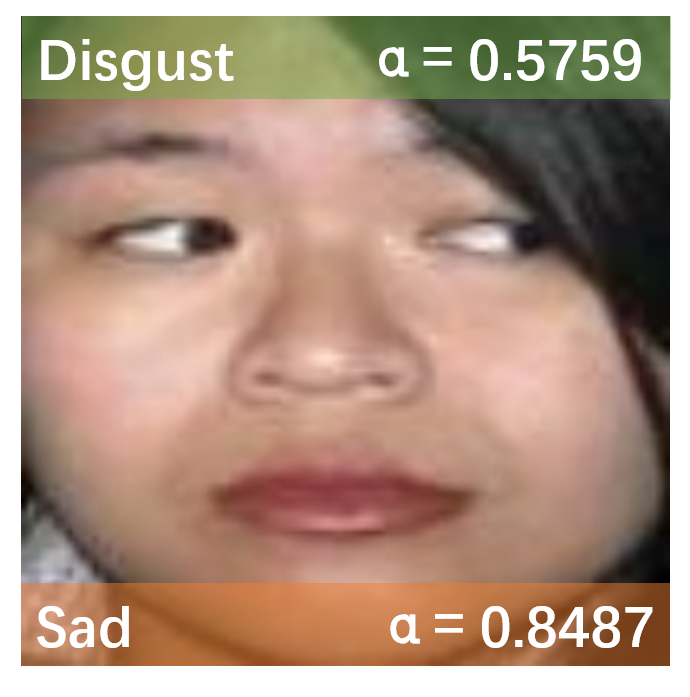}
	\includegraphics[width=0.28\columnwidth]{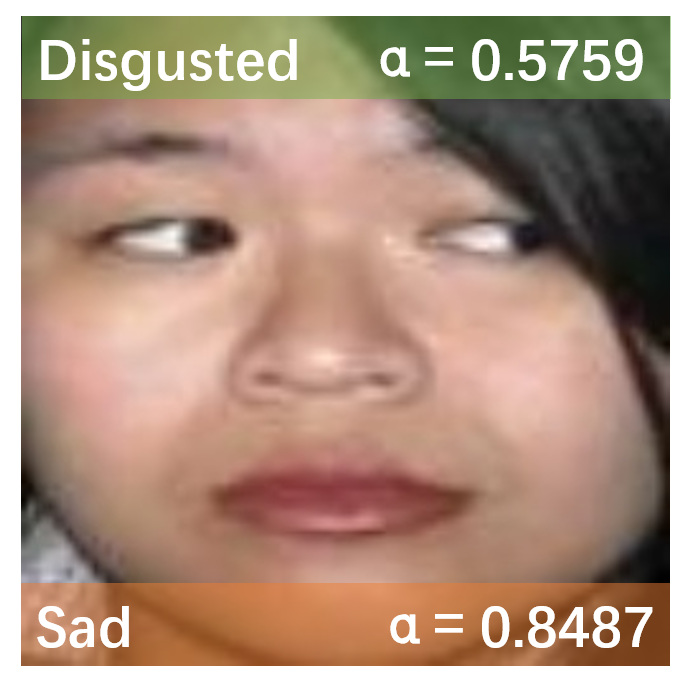}
	\includegraphics[width=0.28\columnwidth]{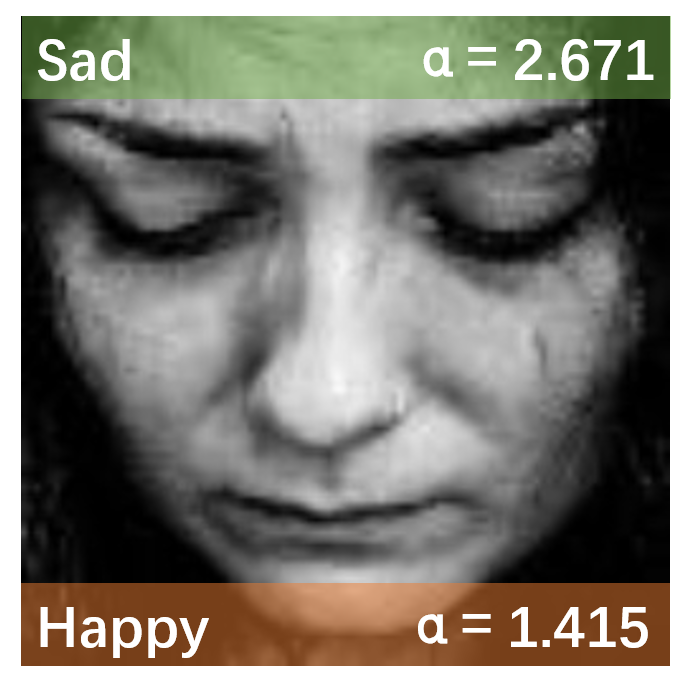}
	\includegraphics[width=0.28\columnwidth]{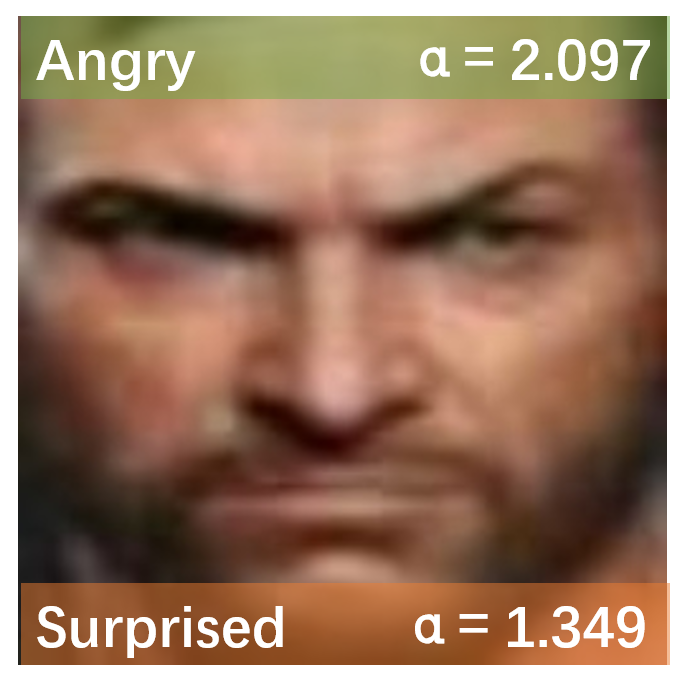}
	\includegraphics[width=0.28\columnwidth]{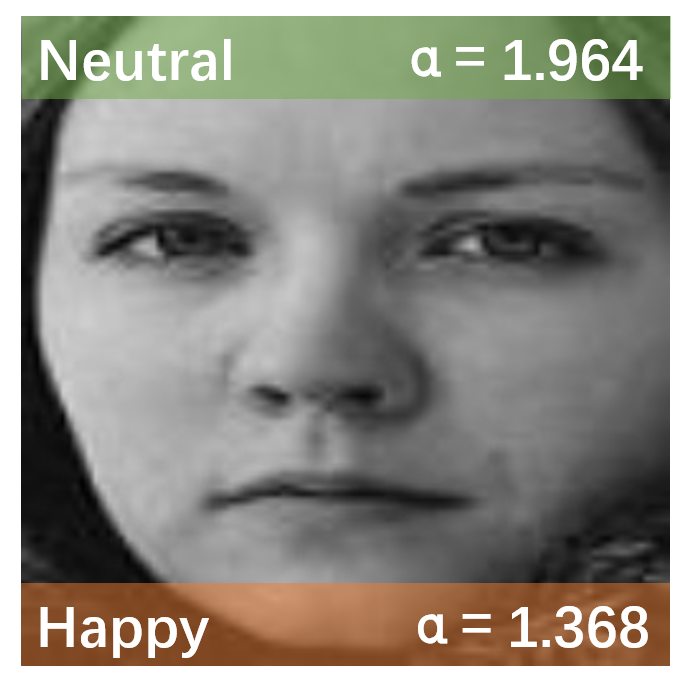}
	\\
	\includegraphics[width=0.28\columnwidth]{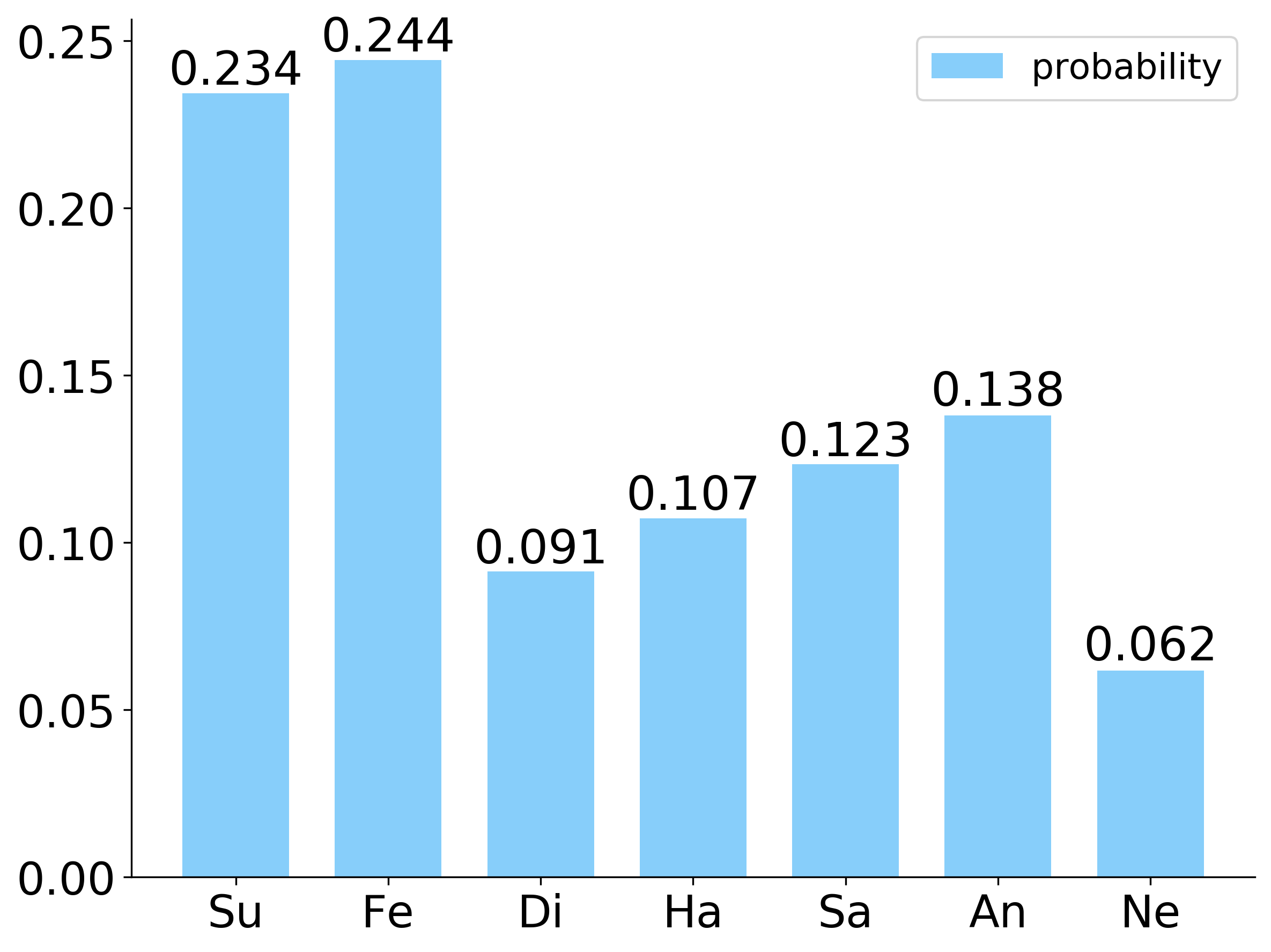}
	\includegraphics[width=0.28\columnwidth]{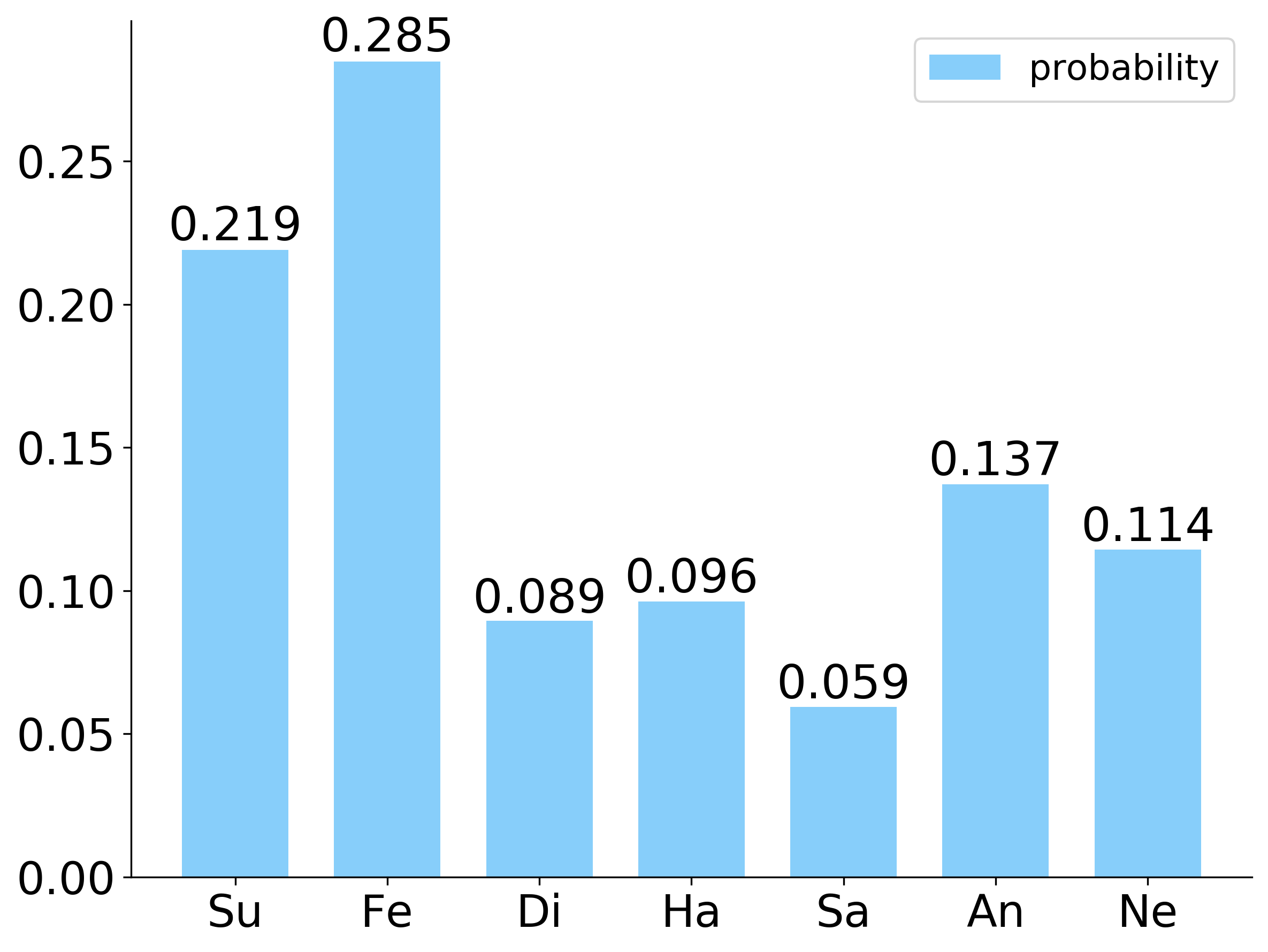}
	\includegraphics[width=0.28\columnwidth]{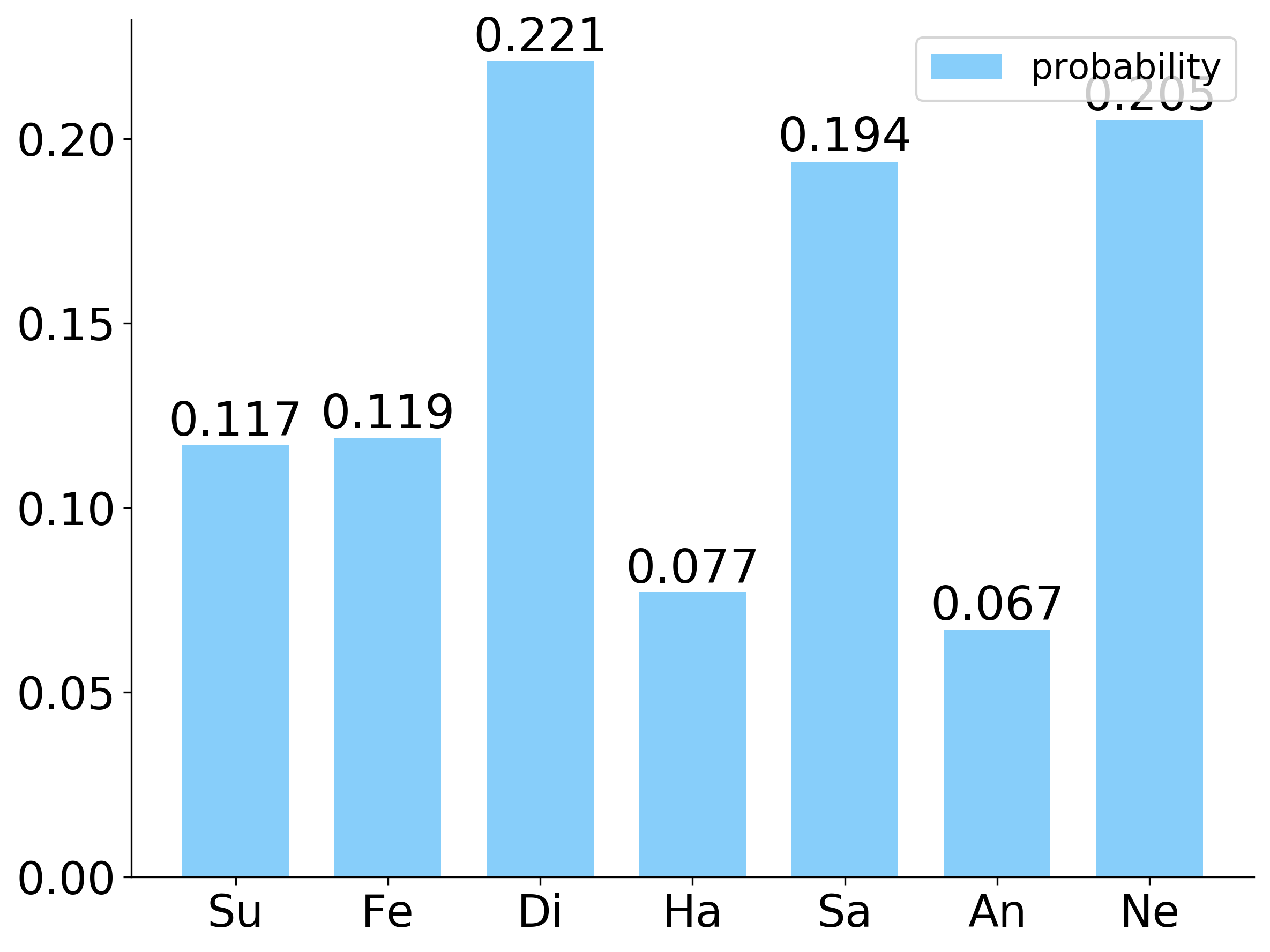}
	\includegraphics[width=0.28\columnwidth]{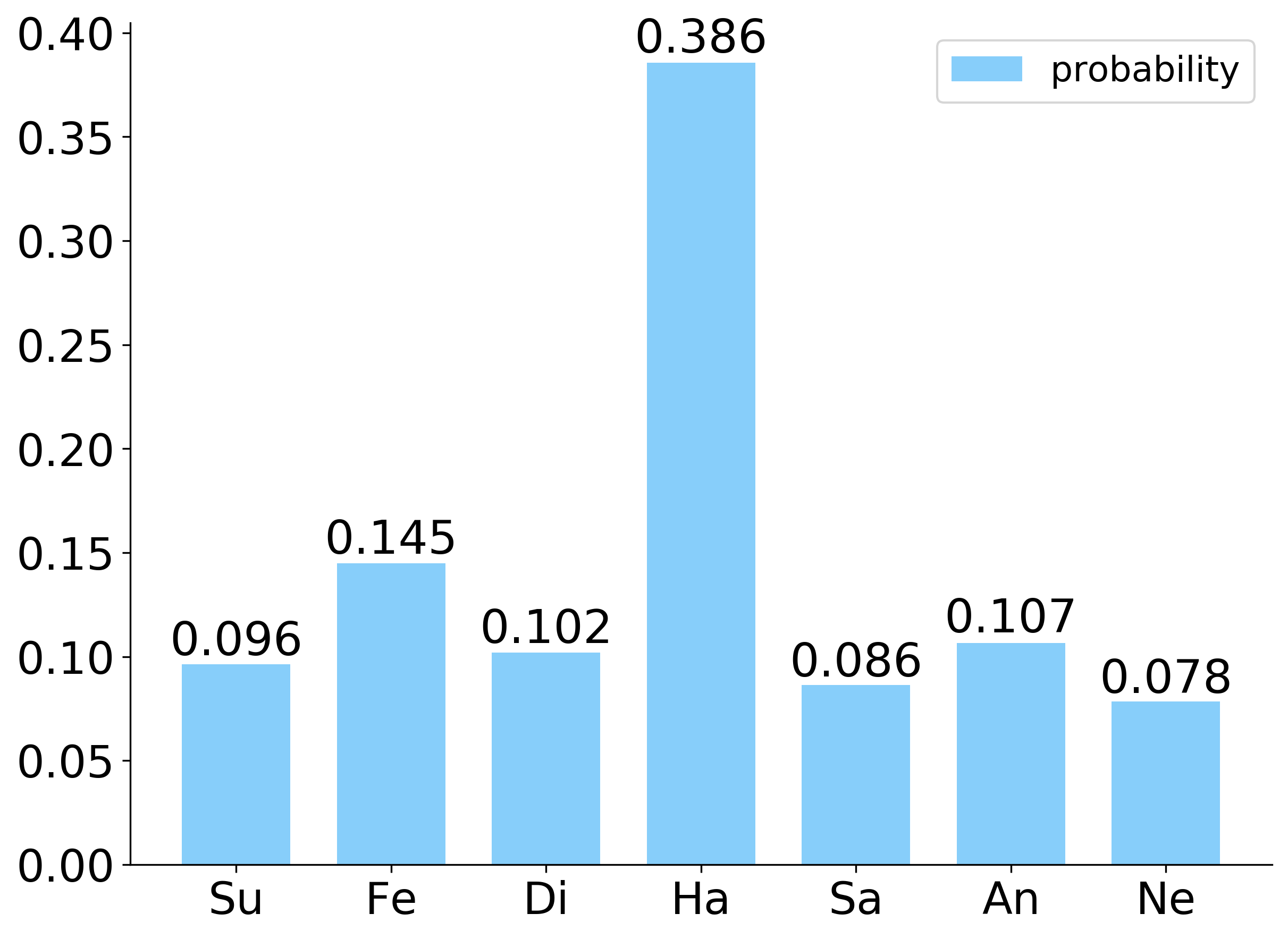}
	\includegraphics[width=0.28\columnwidth]{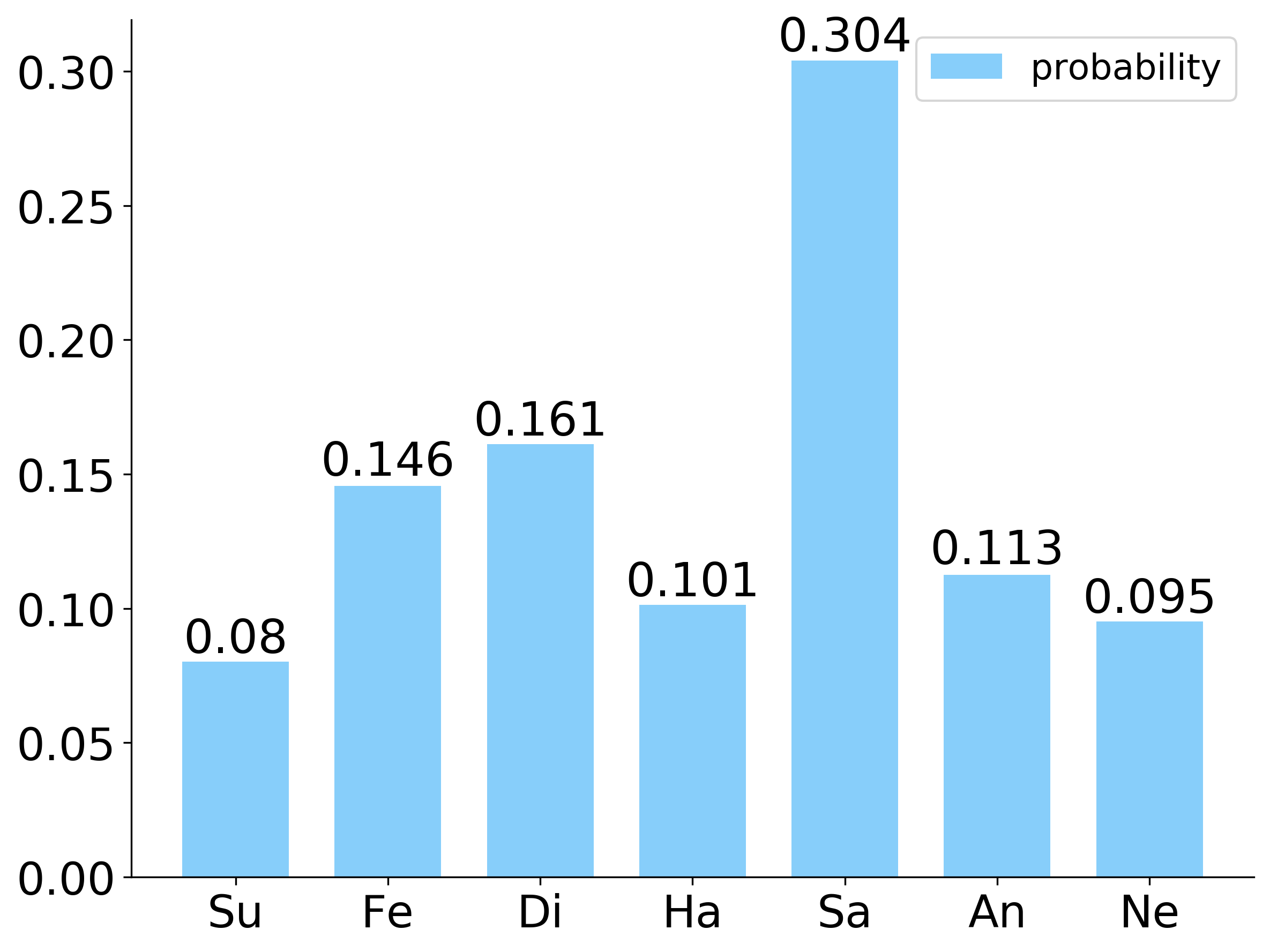}
	\includegraphics[width=0.28\columnwidth]{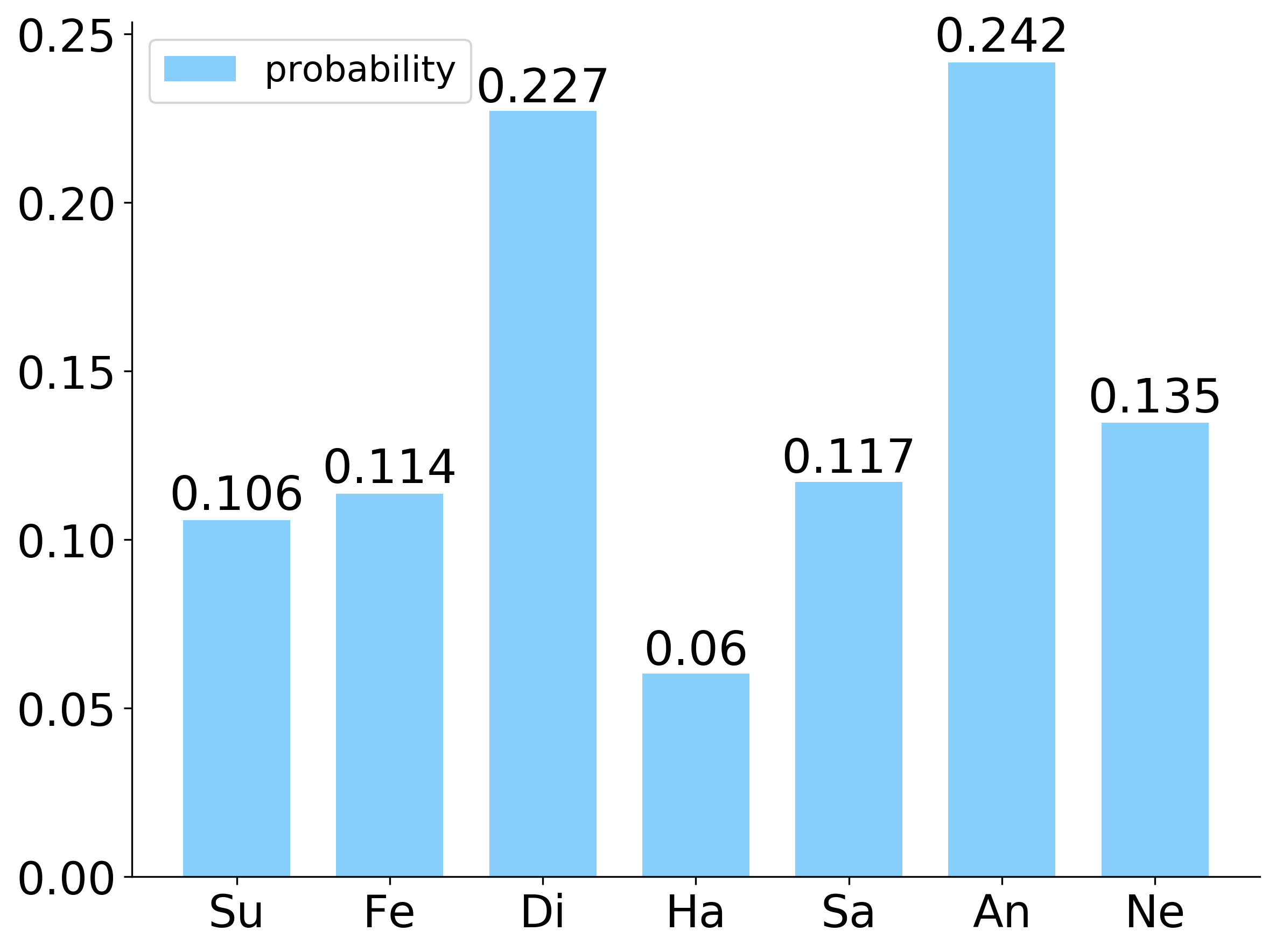}
	\includegraphics[width=0.28\columnwidth]{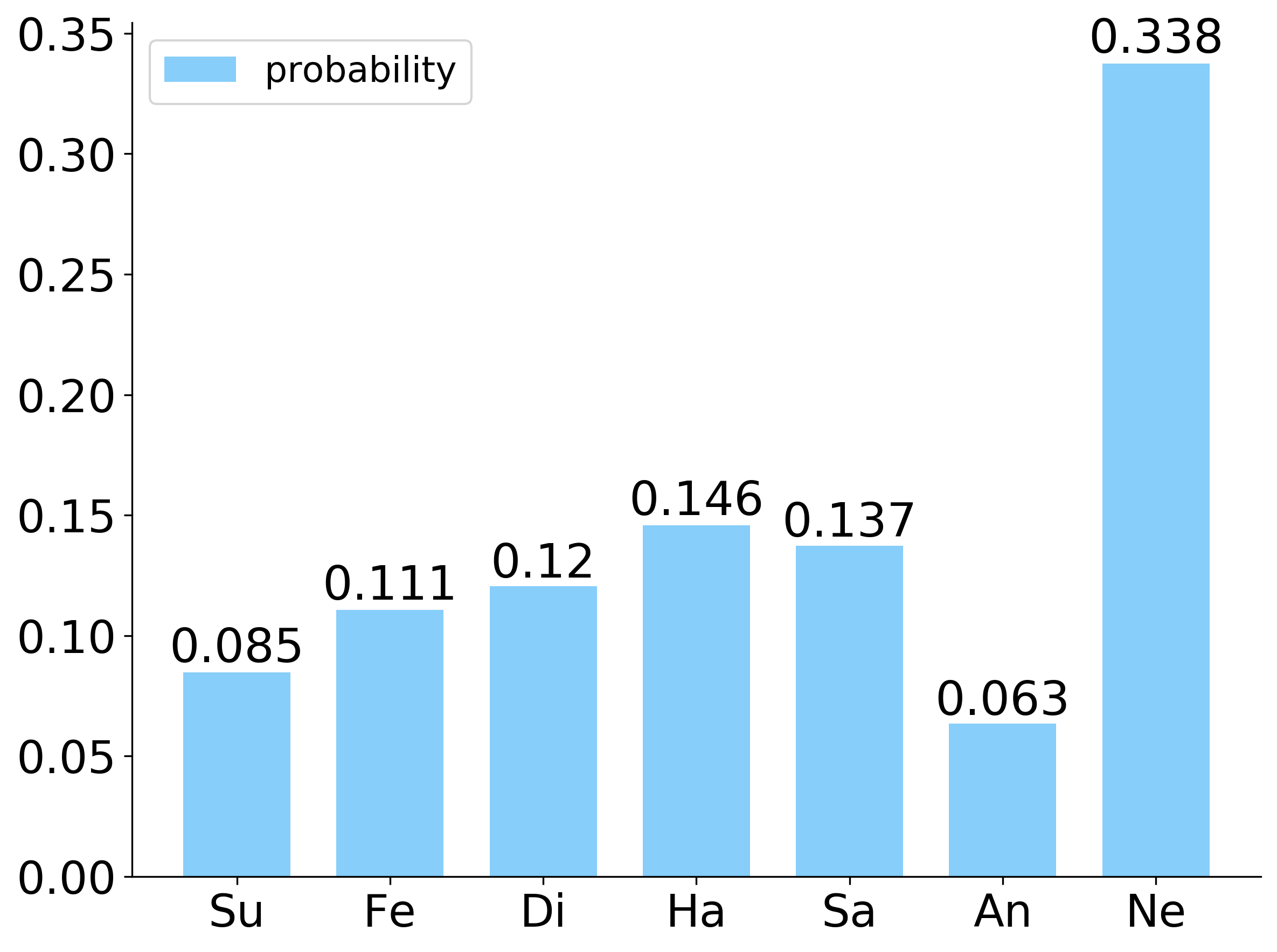}	
	\\
	\rule[0.2\baselineskip]{18cm}{1pt}
	\vspace{0.5cm}	
	\includegraphics[width=0.28\columnwidth]{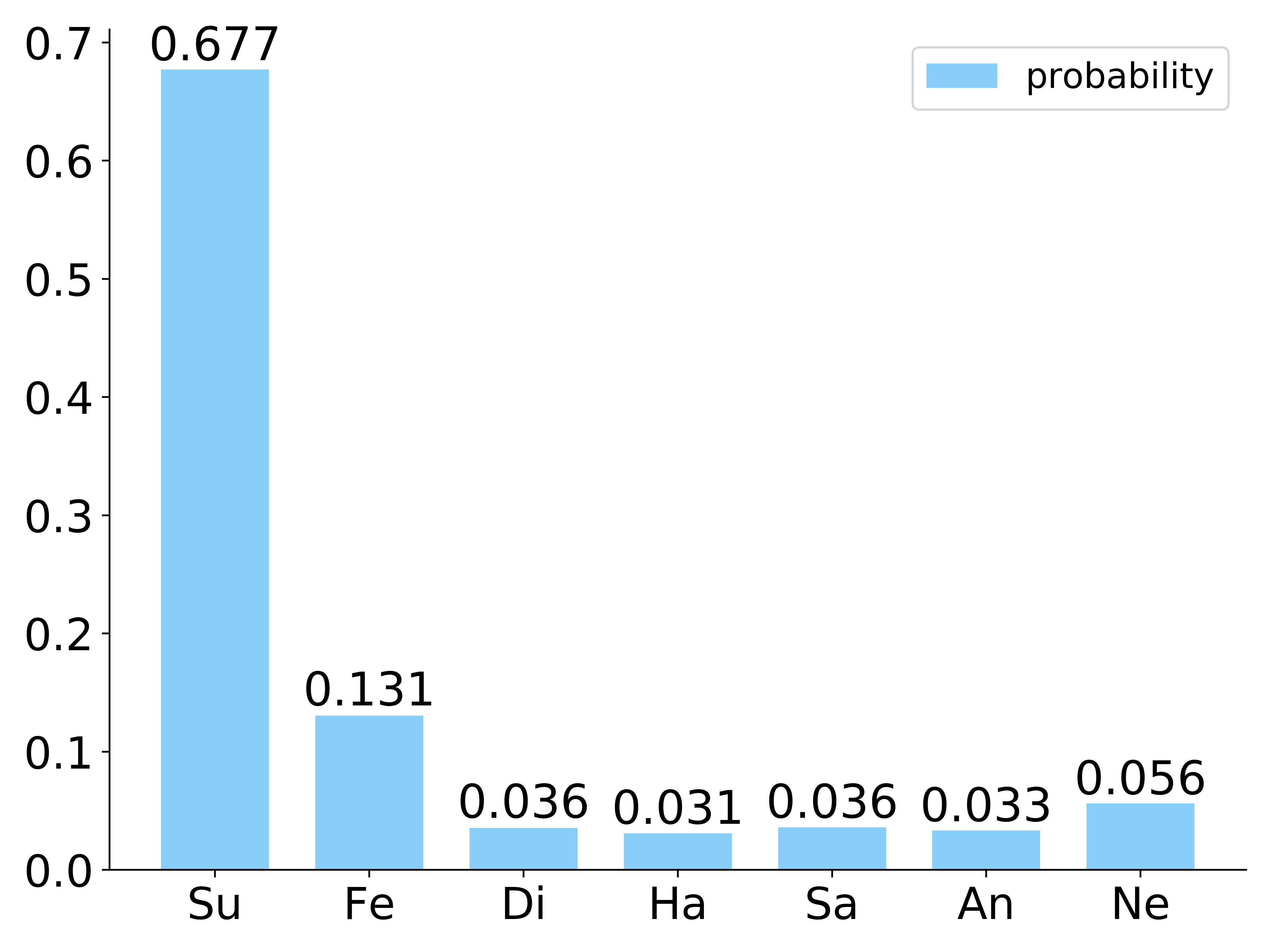}
	\includegraphics[width=0.28\columnwidth]{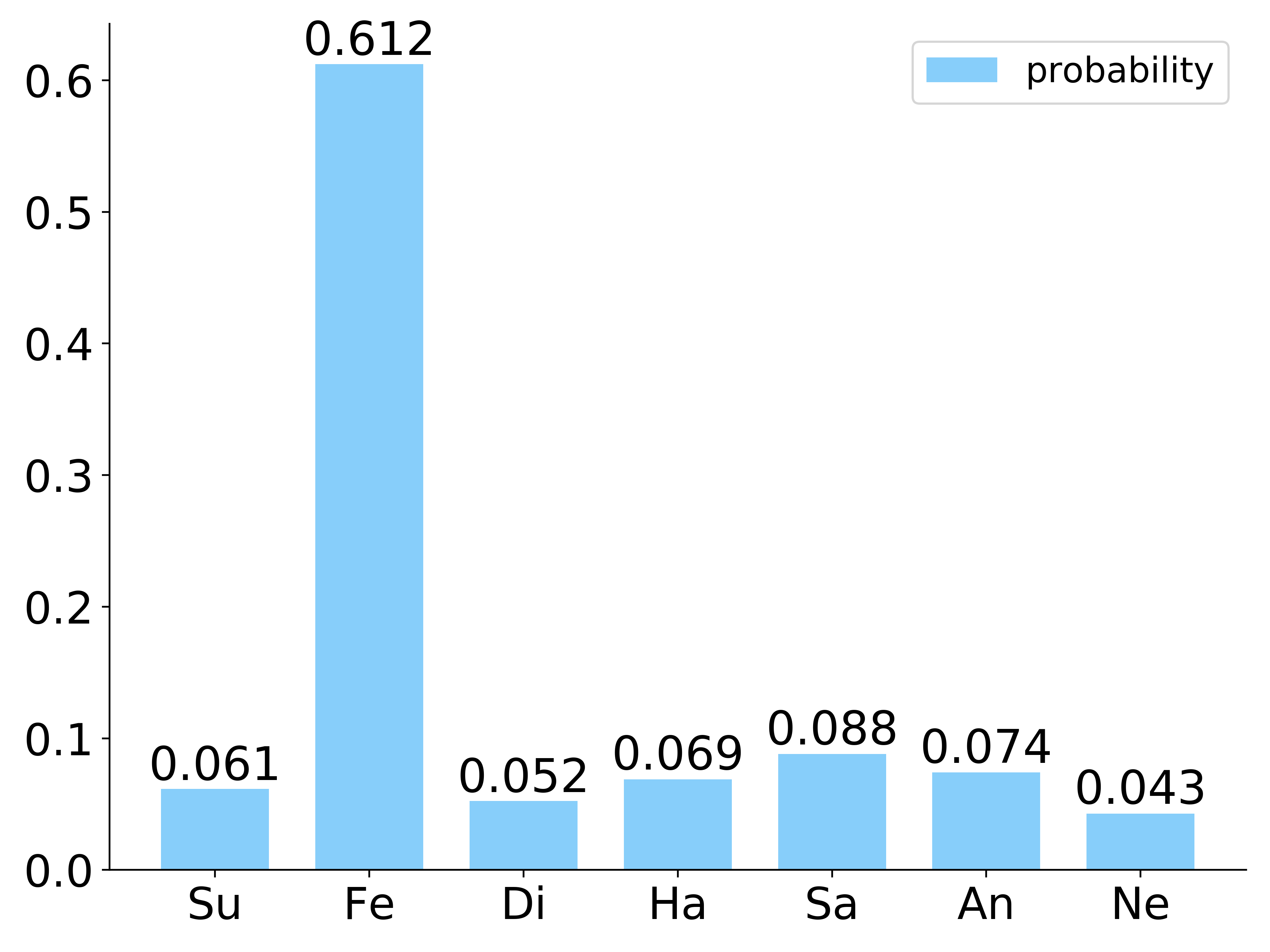}
	\includegraphics[width=0.28\columnwidth]{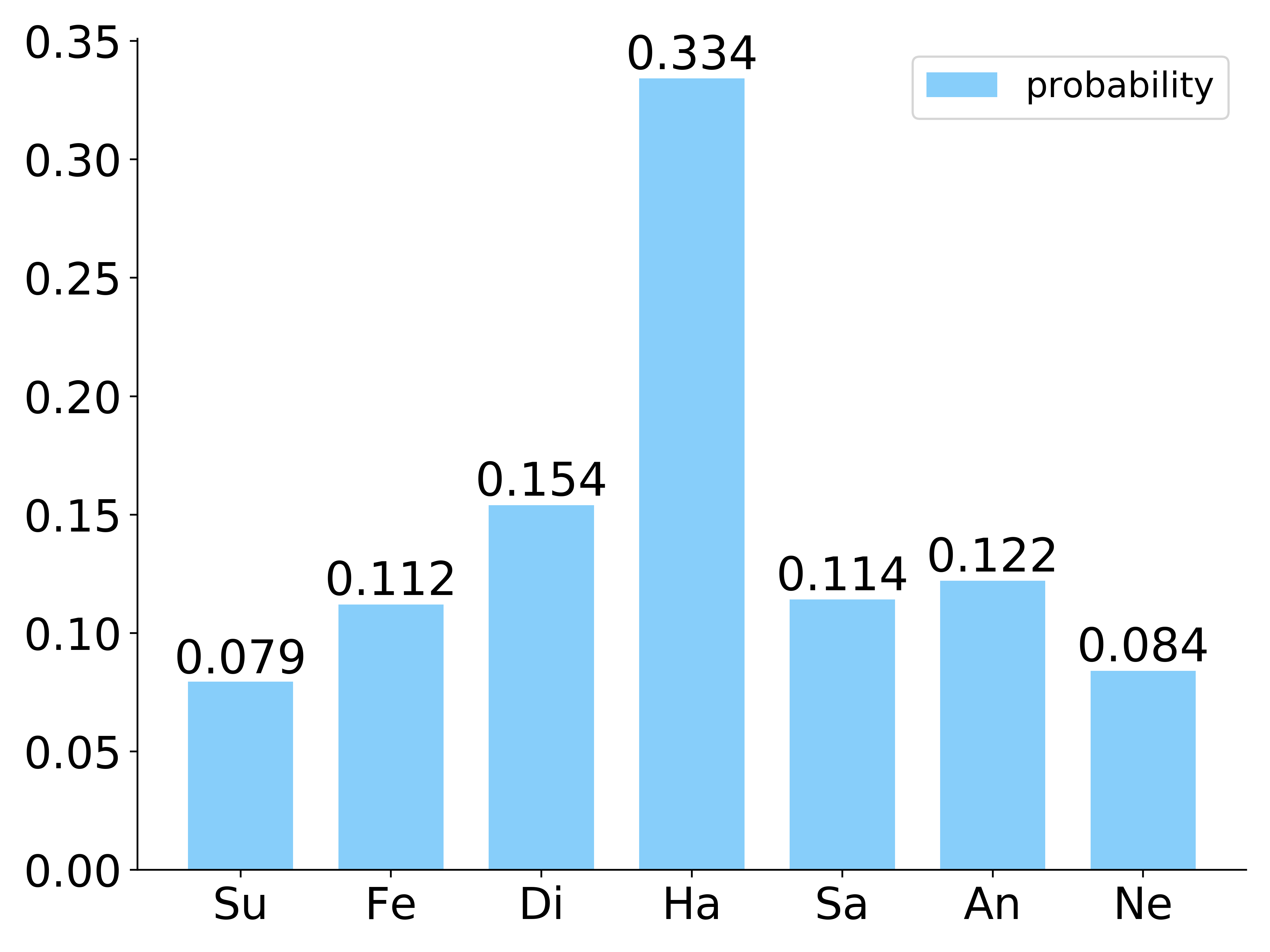}
	\includegraphics[width=0.28\columnwidth]{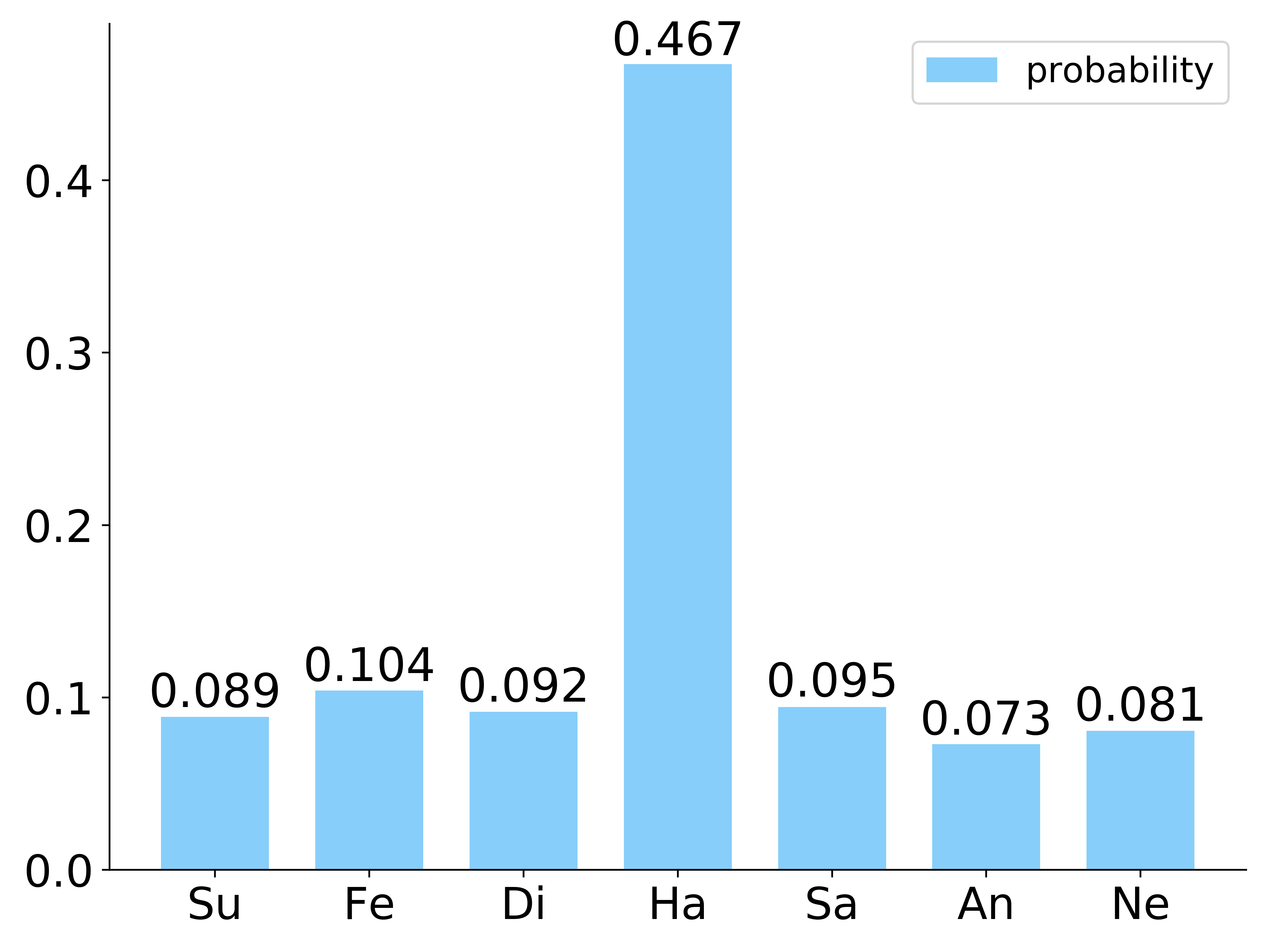}
	\includegraphics[width=0.28\columnwidth]{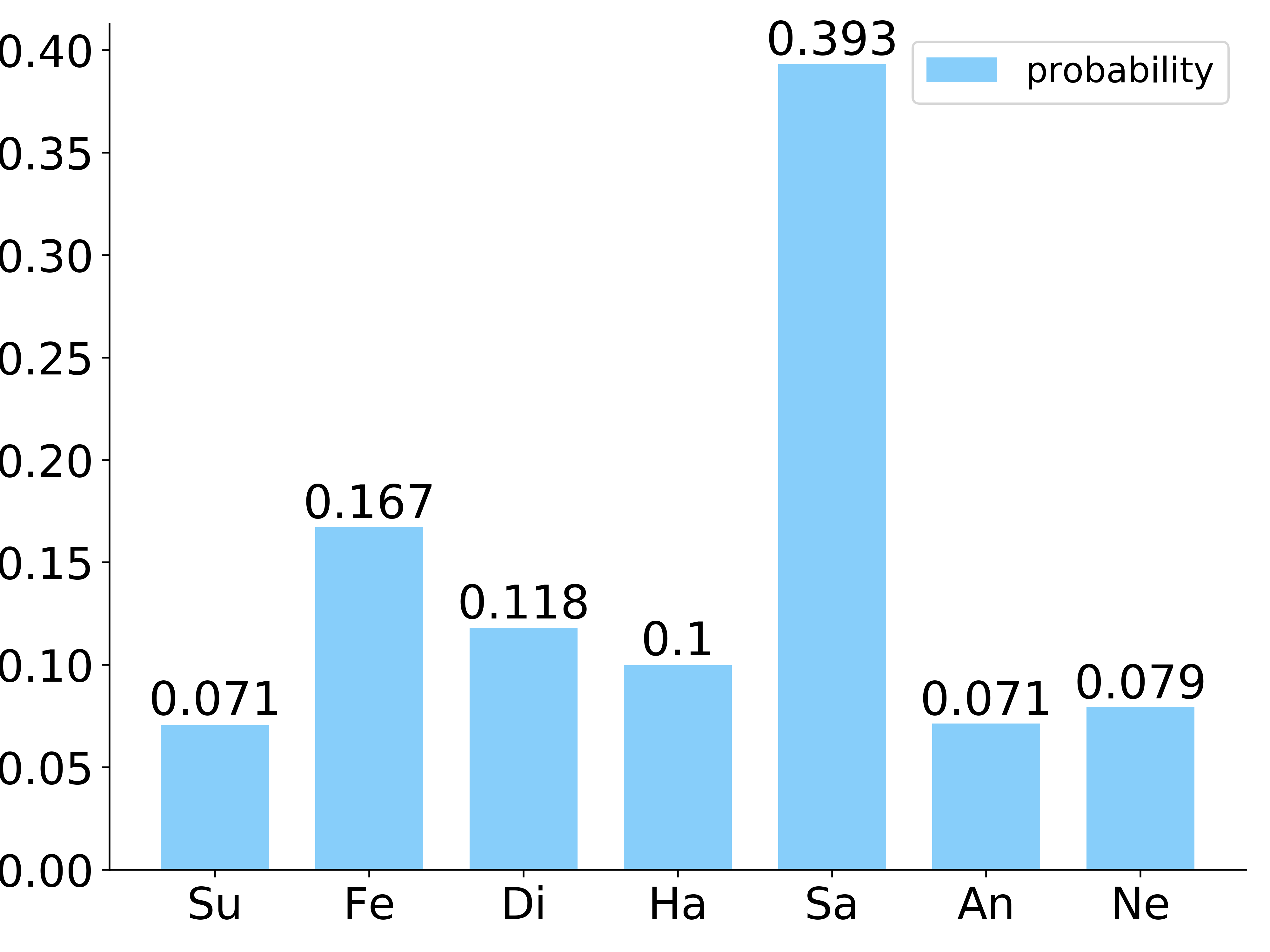}
	\includegraphics[width=0.28\columnwidth]{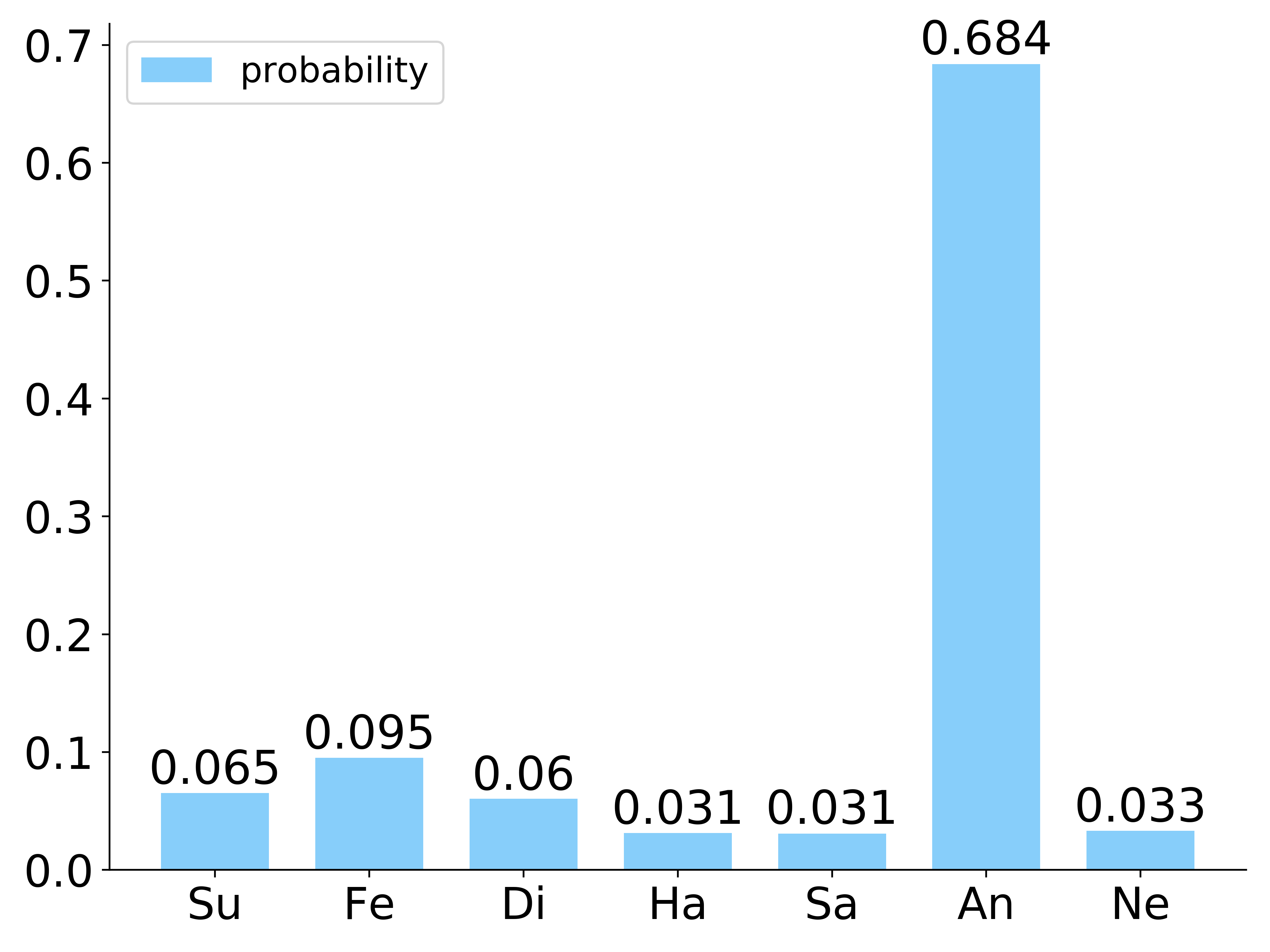}
	\includegraphics[width=0.28\columnwidth]{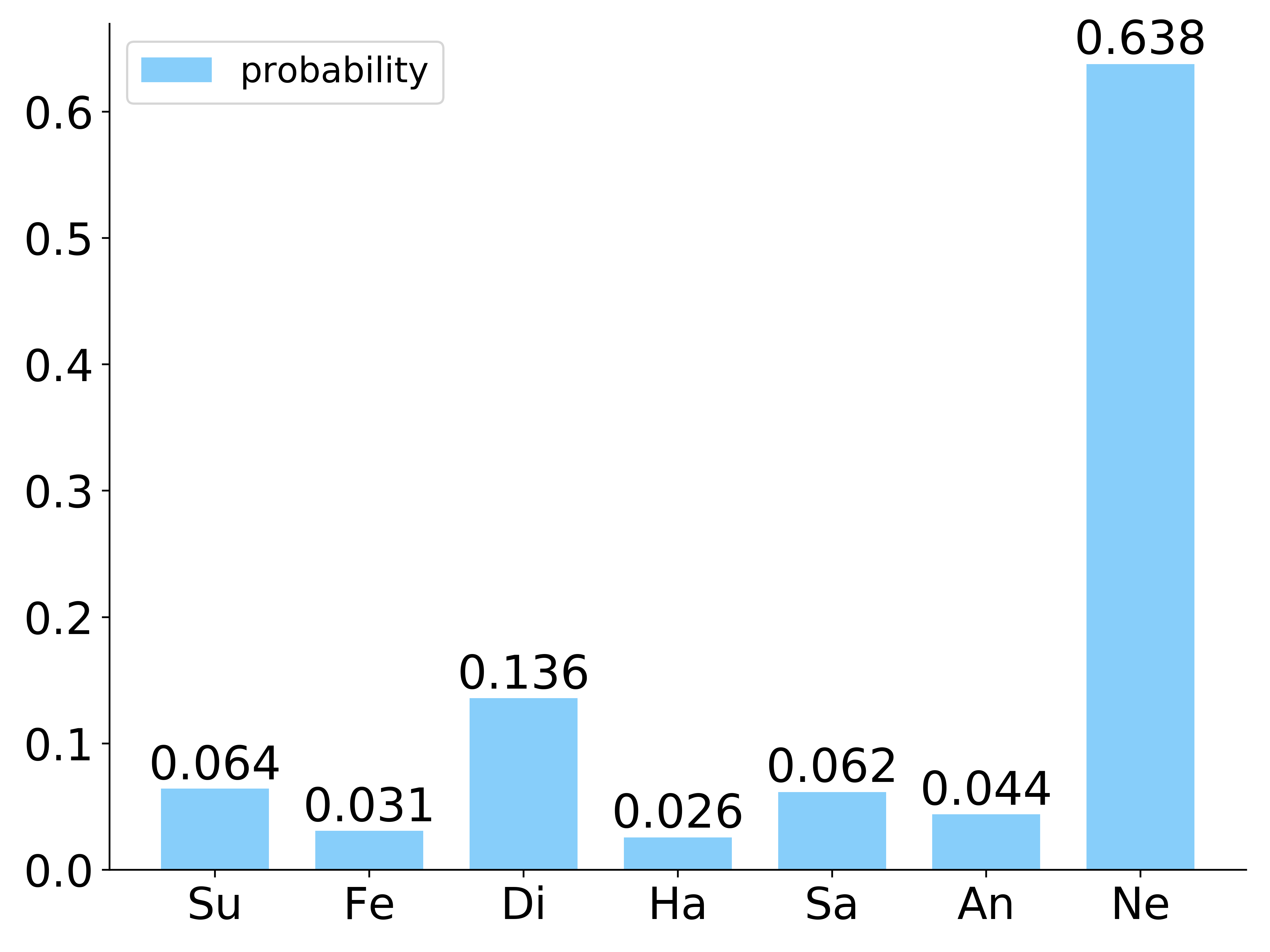}\\
	\includegraphics[width=0.28\columnwidth]{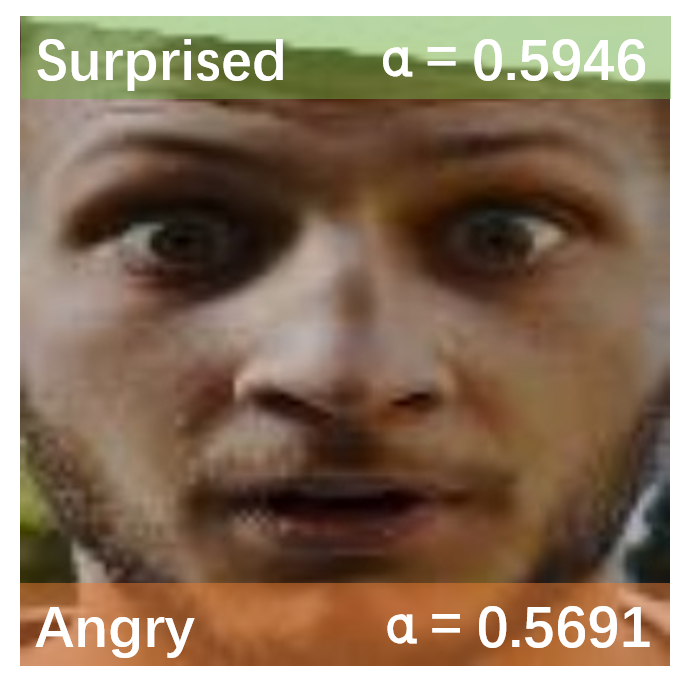}
	\includegraphics[width=0.28\columnwidth]{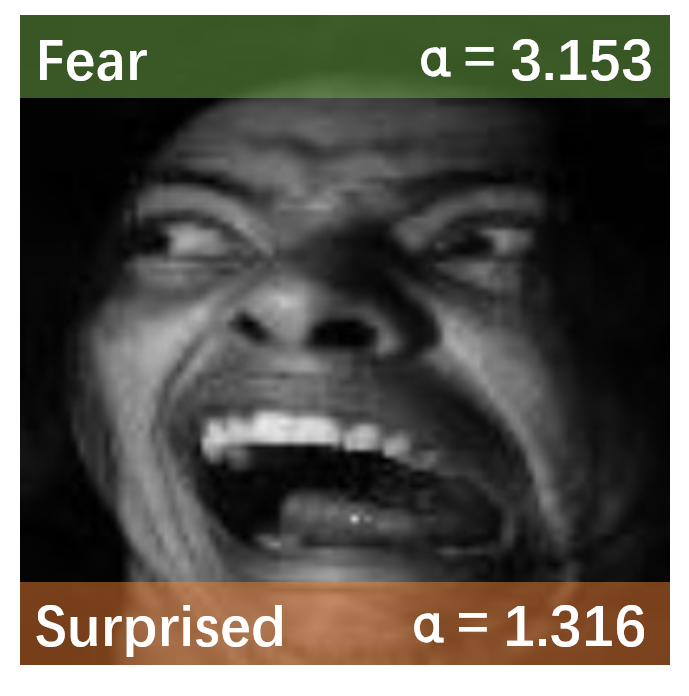}
	\includegraphics[width=0.28\columnwidth]{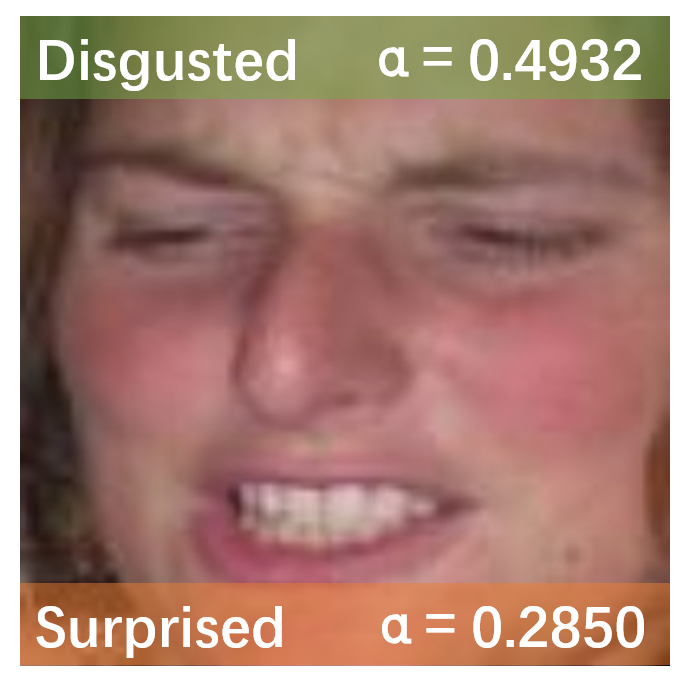}
	\includegraphics[width=0.28\columnwidth]{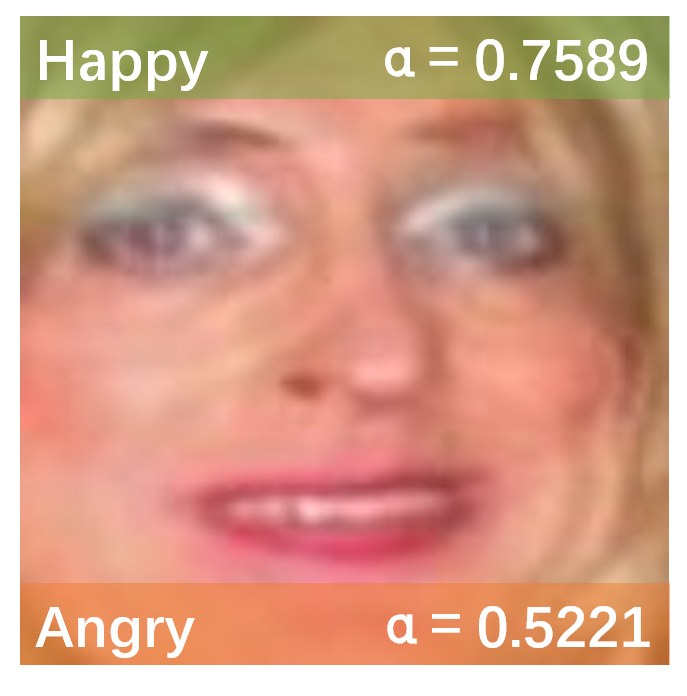}
	\includegraphics[width=0.28\columnwidth]{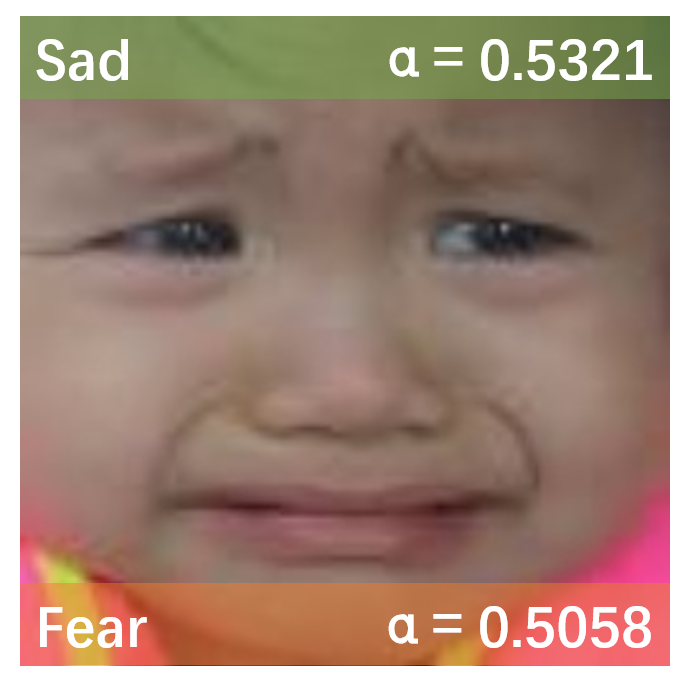}
	\includegraphics[width=0.28\columnwidth]{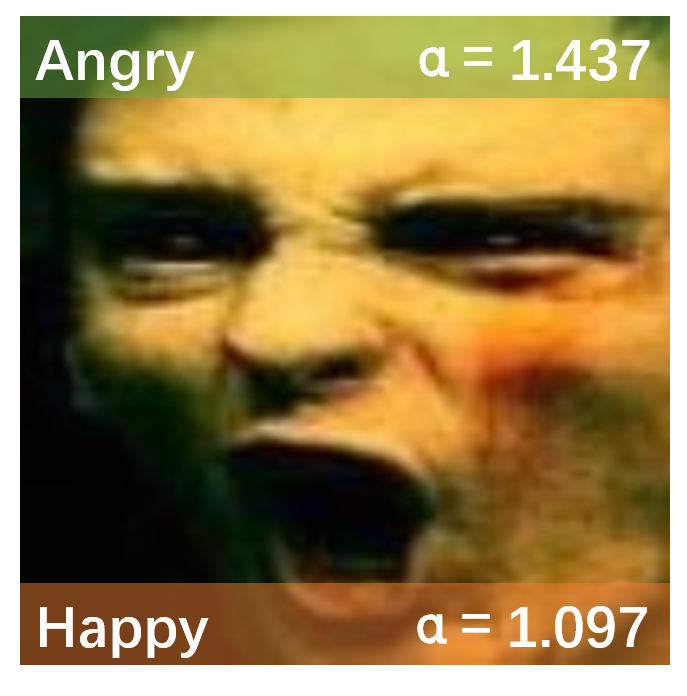}
	\includegraphics[width=0.28\columnwidth]{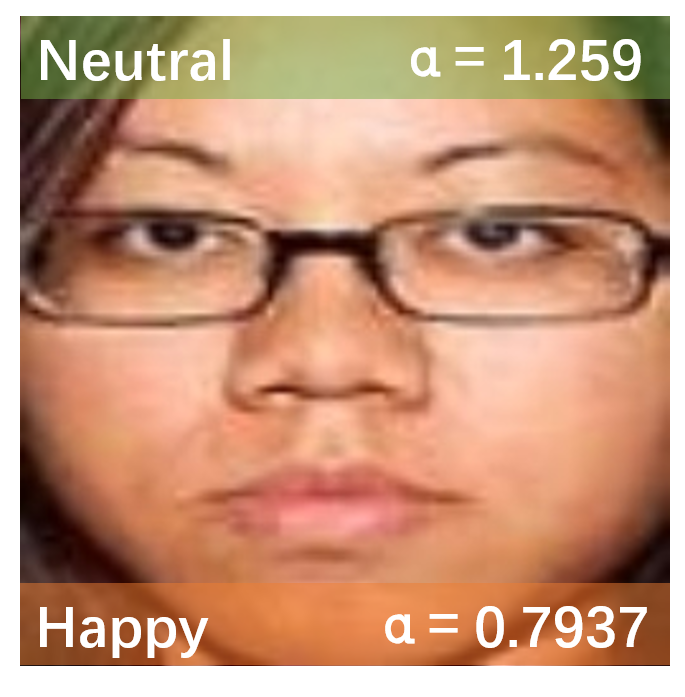}\\
	\includegraphics[width=0.28\columnwidth]{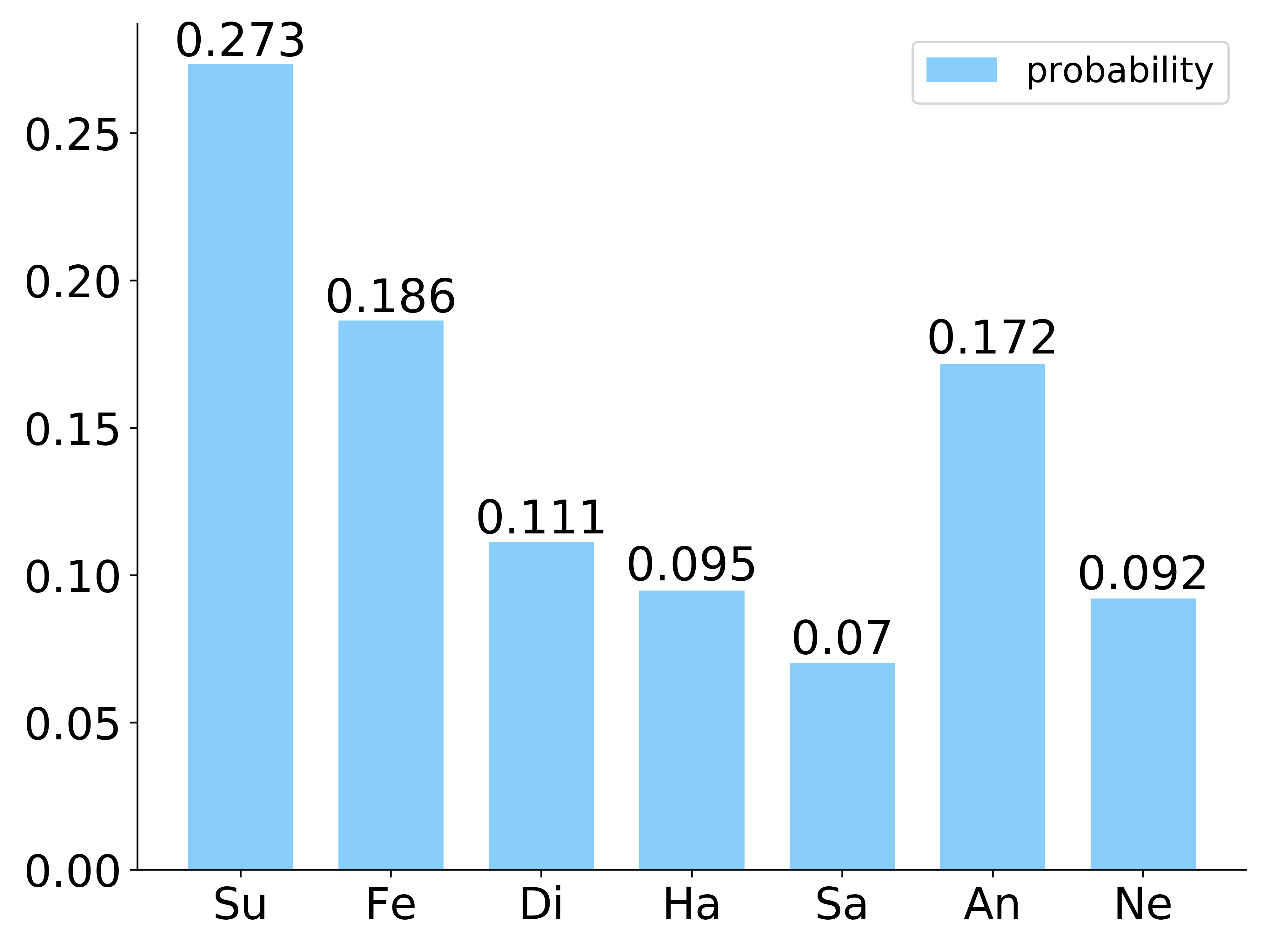}
	\includegraphics[width=0.28\columnwidth]{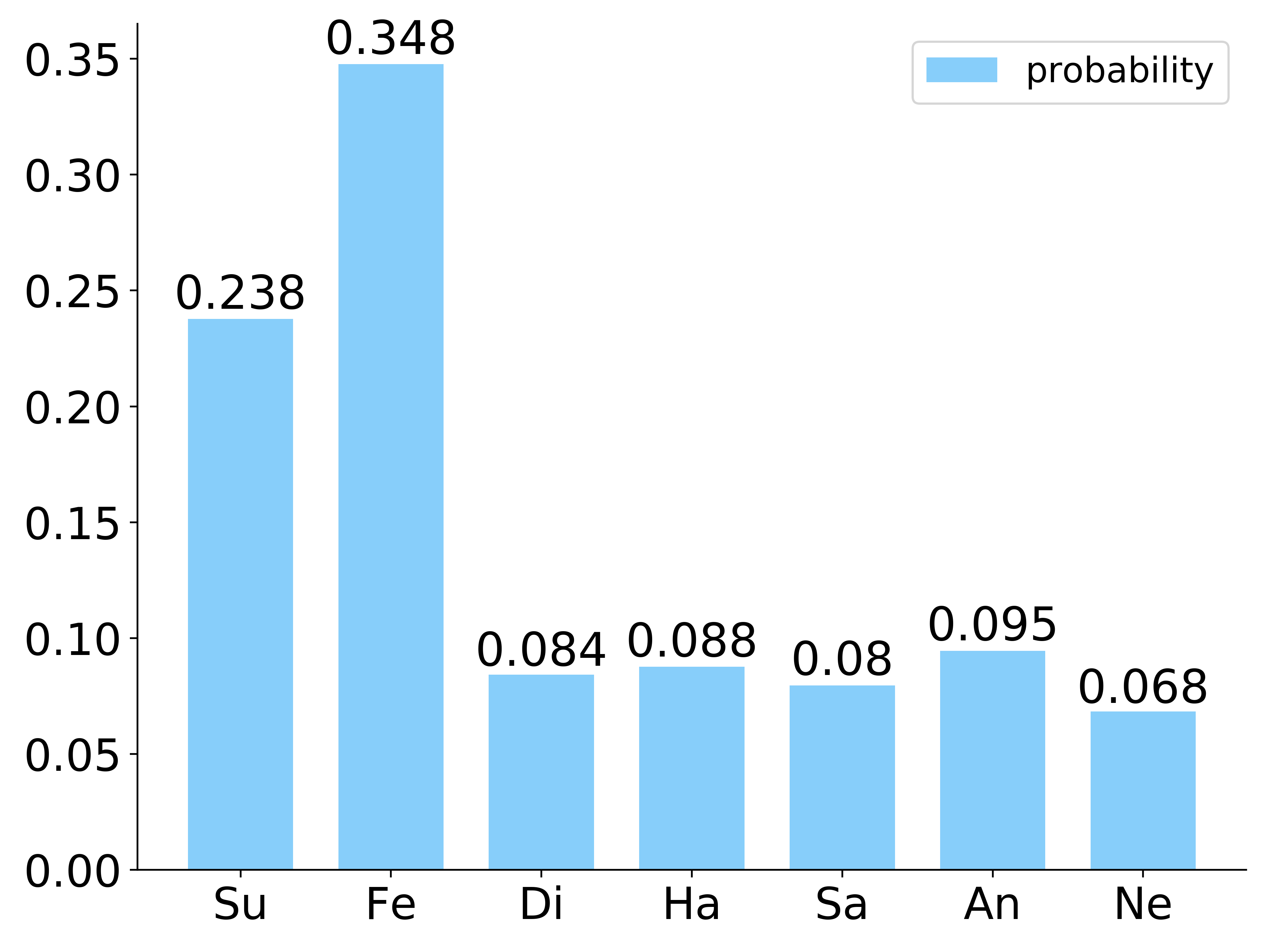}
	\includegraphics[width=0.28\columnwidth]{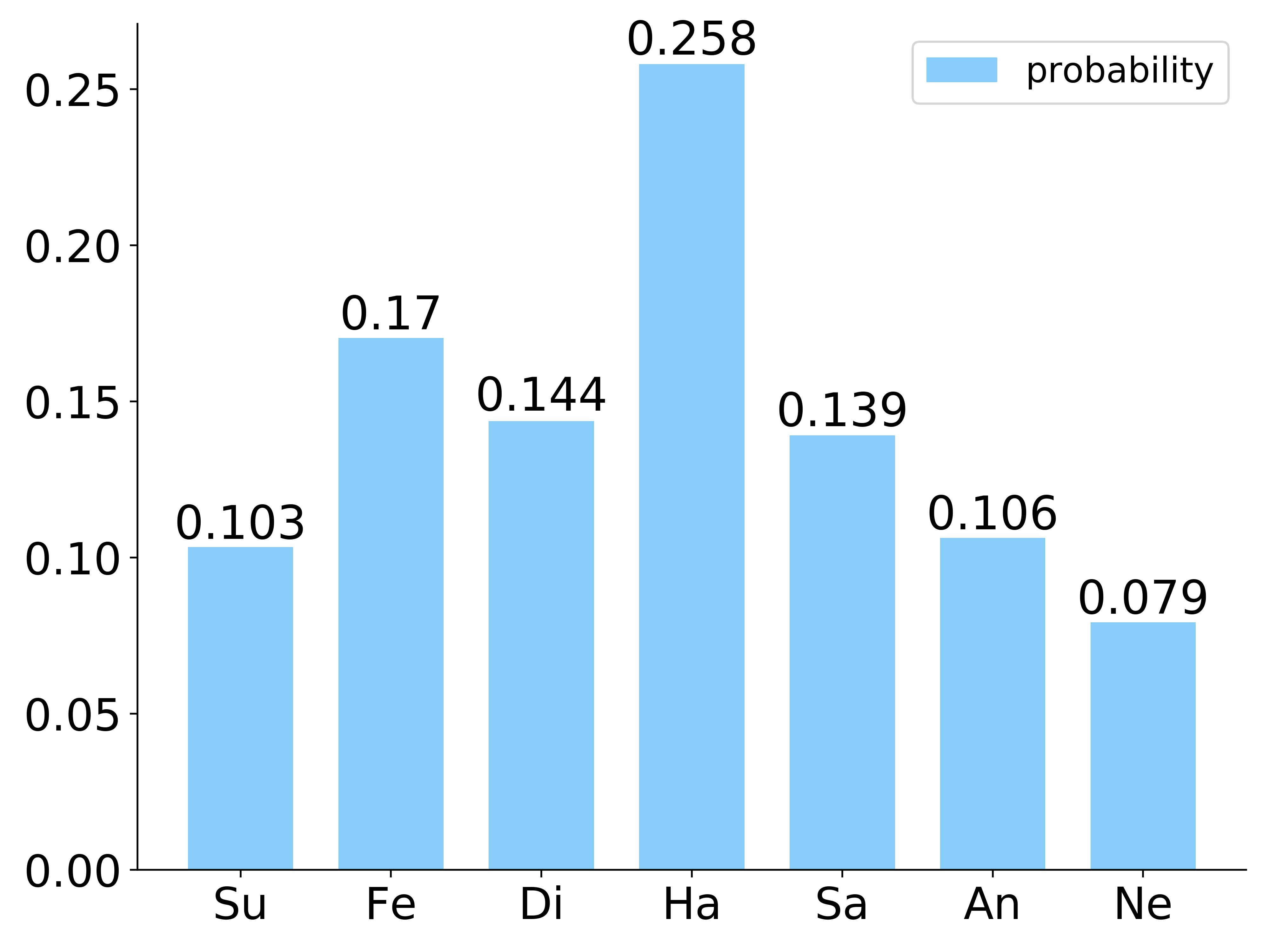}
	\includegraphics[width=0.28\columnwidth]{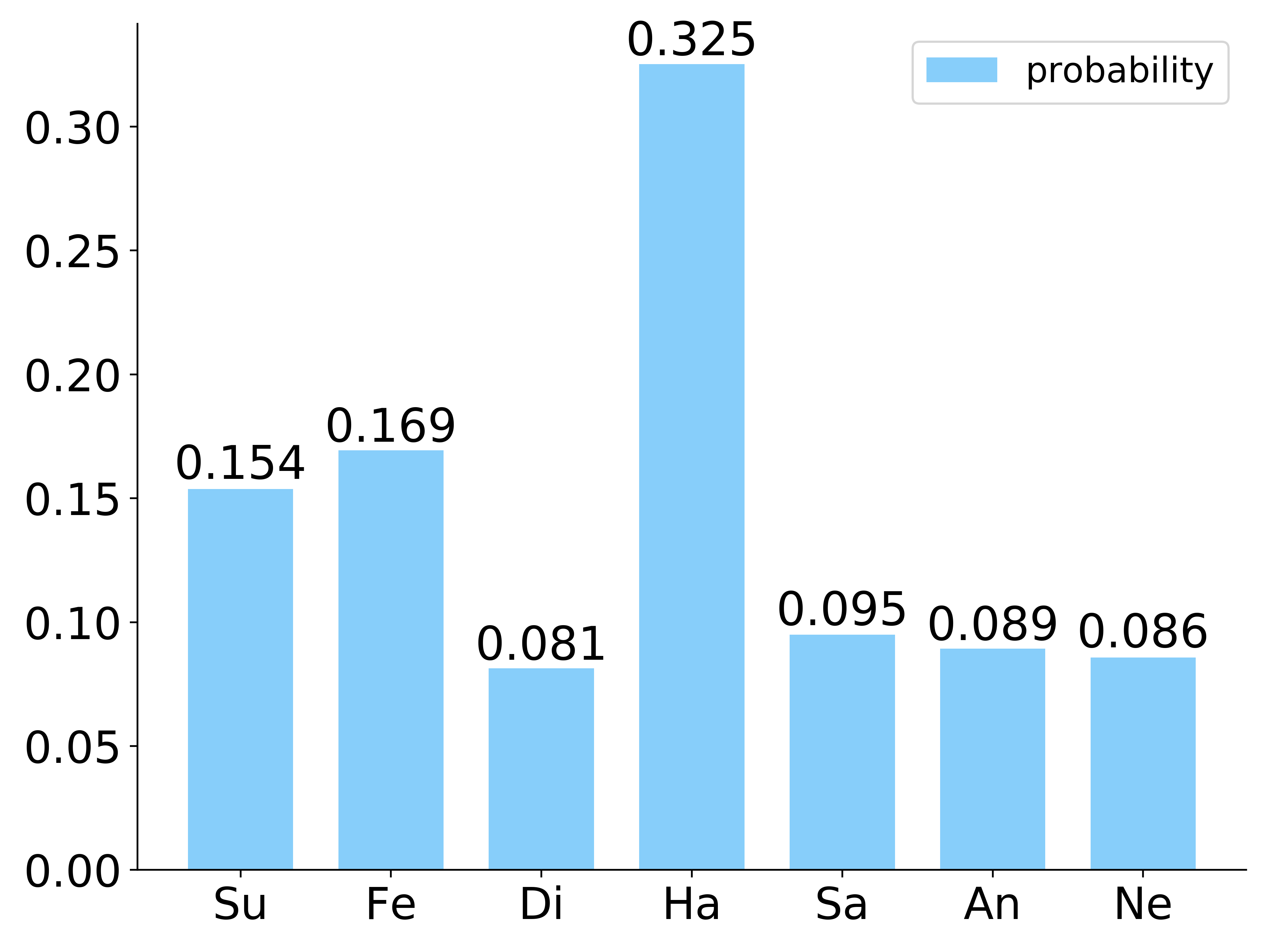}
	\includegraphics[width=0.28\columnwidth]{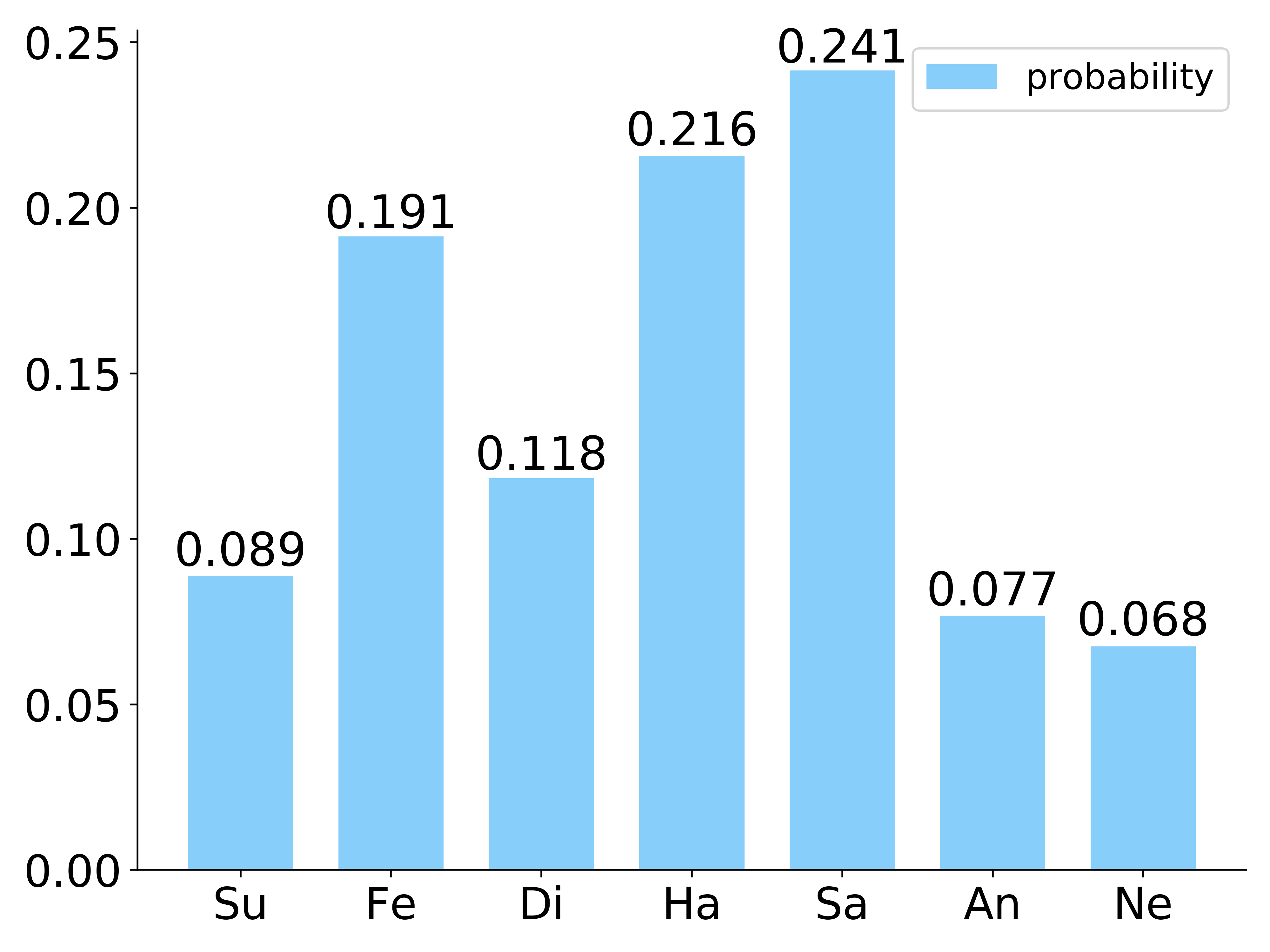}
	\includegraphics[width=0.28\columnwidth]{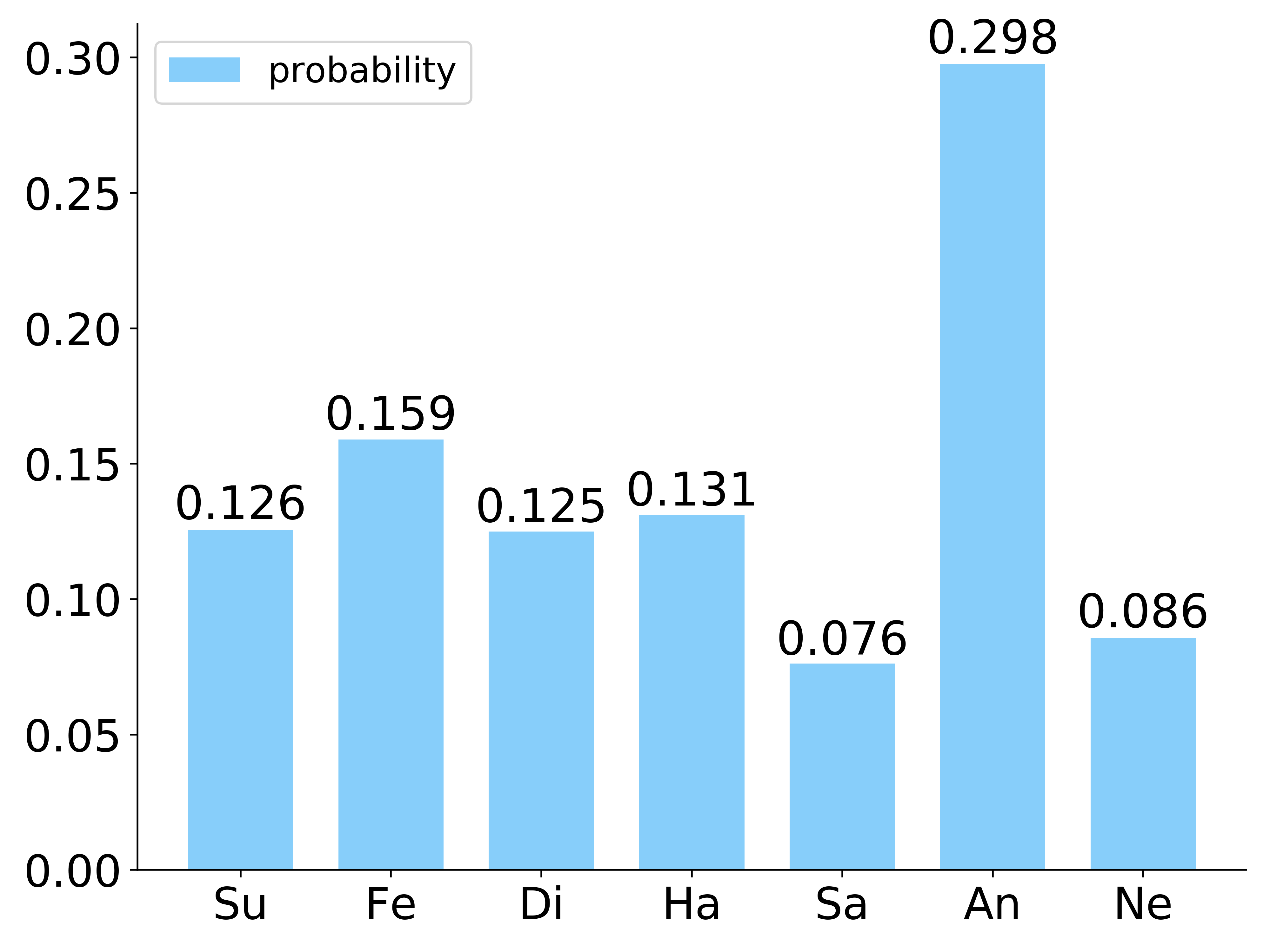}
	\includegraphics[width=0.28\columnwidth]{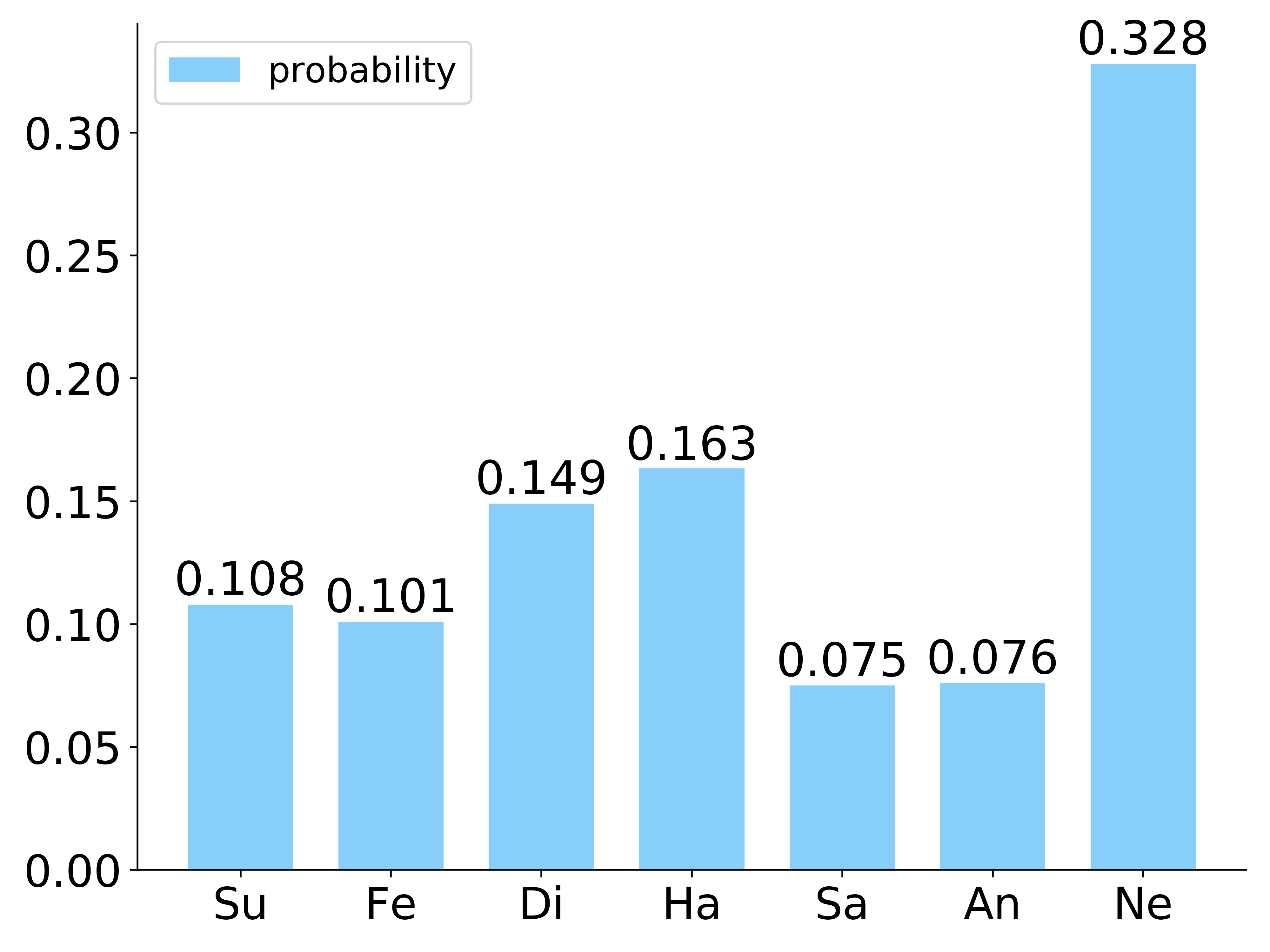}
	%\\
	%\rule[0.2\baselineskip]{17.5cm}{1pt}
	\caption{Visualization of confidences and label distributions obtained by SCE-ALD. The second row shows seven facial images from each category, and the first and third rows show label distributions corresponding to original labels and noisy labels, respectively.}
	\label{fig-vis}
\end{figure*}

\begin{figure*}[t]
	\centering		
%	\rule[0.2\baselineskip]{17.2cm}{1pt}
%	\vspace{0.5cm}	
    \rule[0.2\baselineskip]{18cm}{1pt}
	\includegraphics[width=0.28\columnwidth]{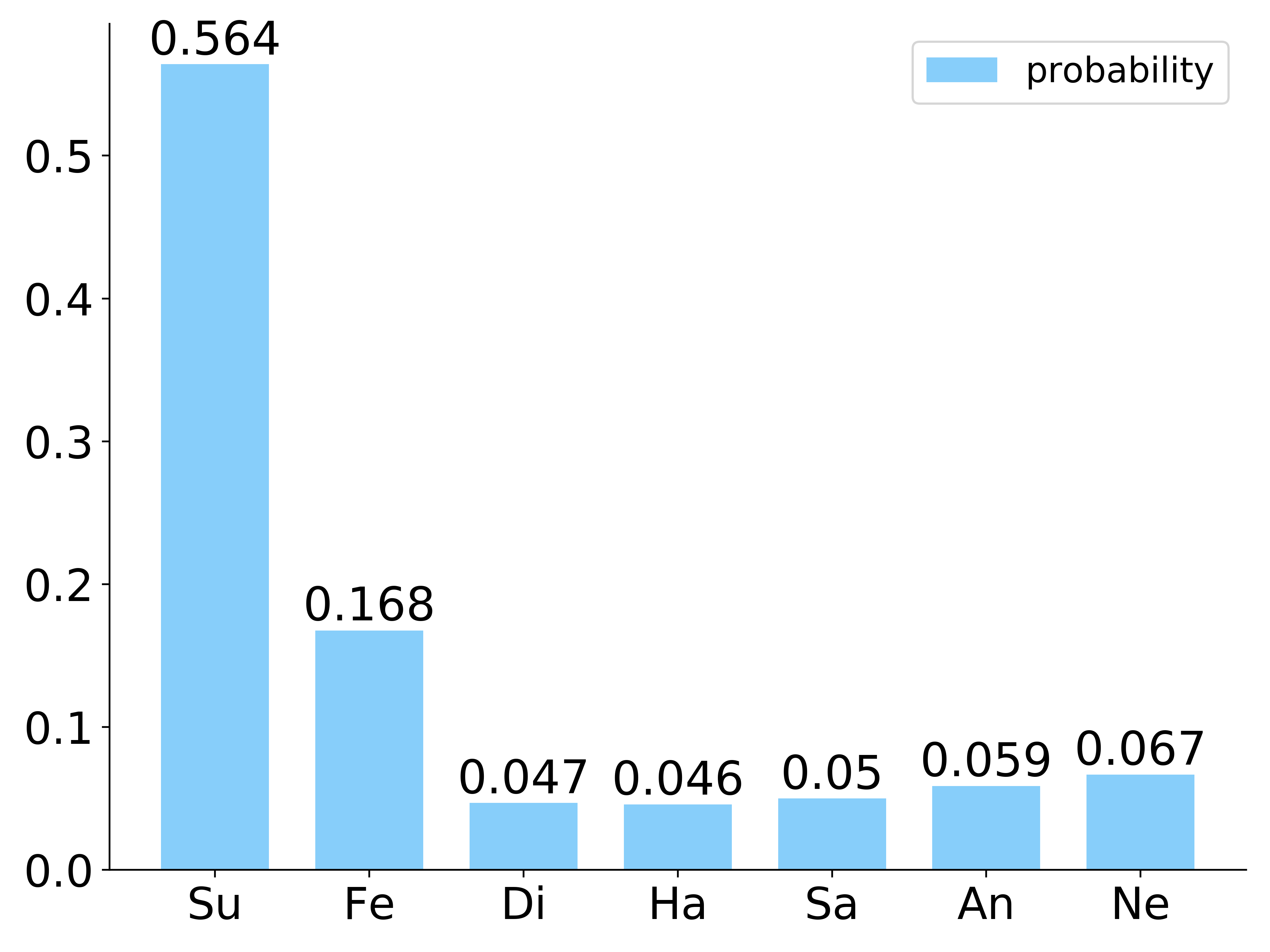}
	\includegraphics[width=0.28\columnwidth]{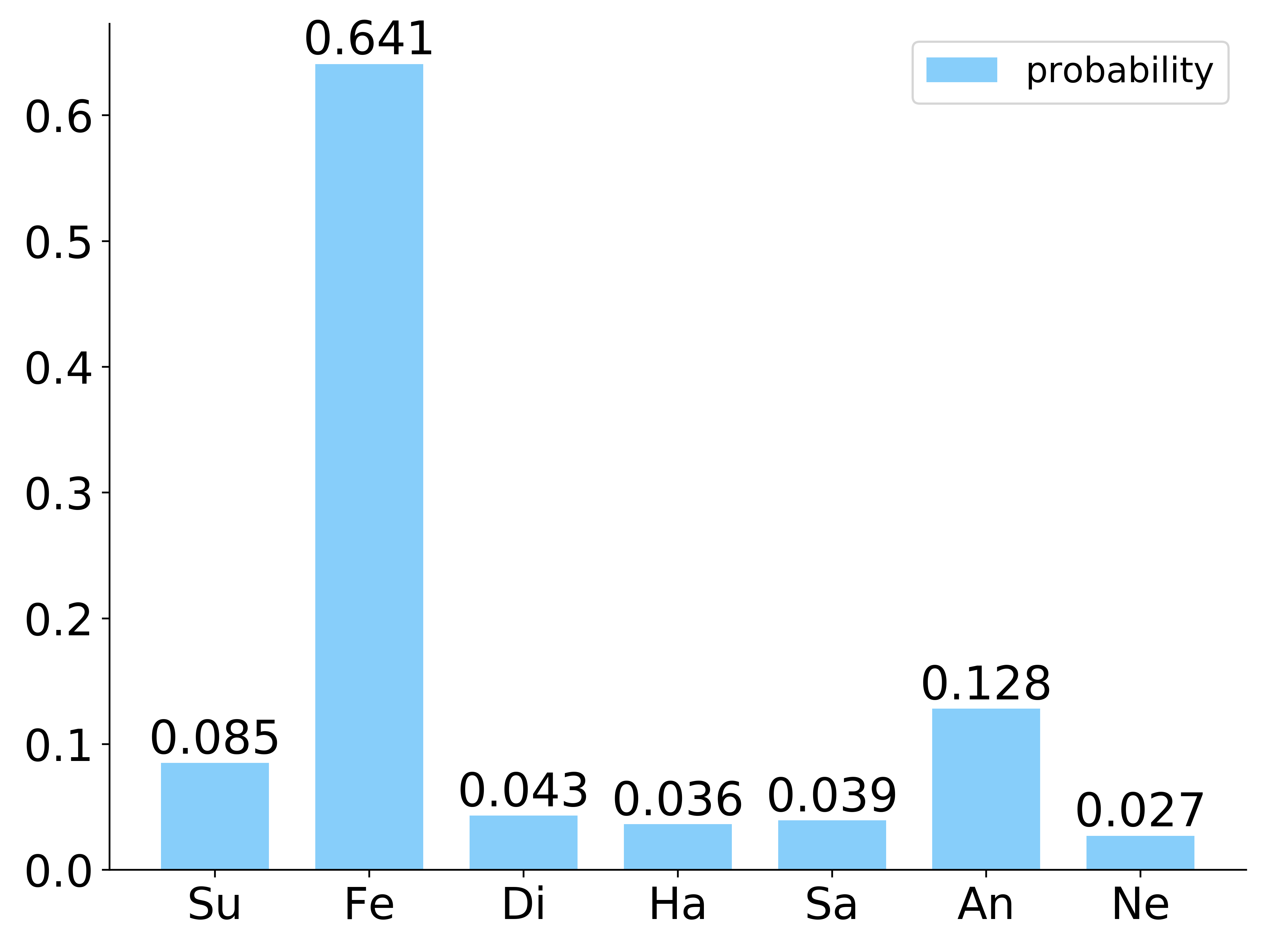}
	\includegraphics[width=0.28\columnwidth]{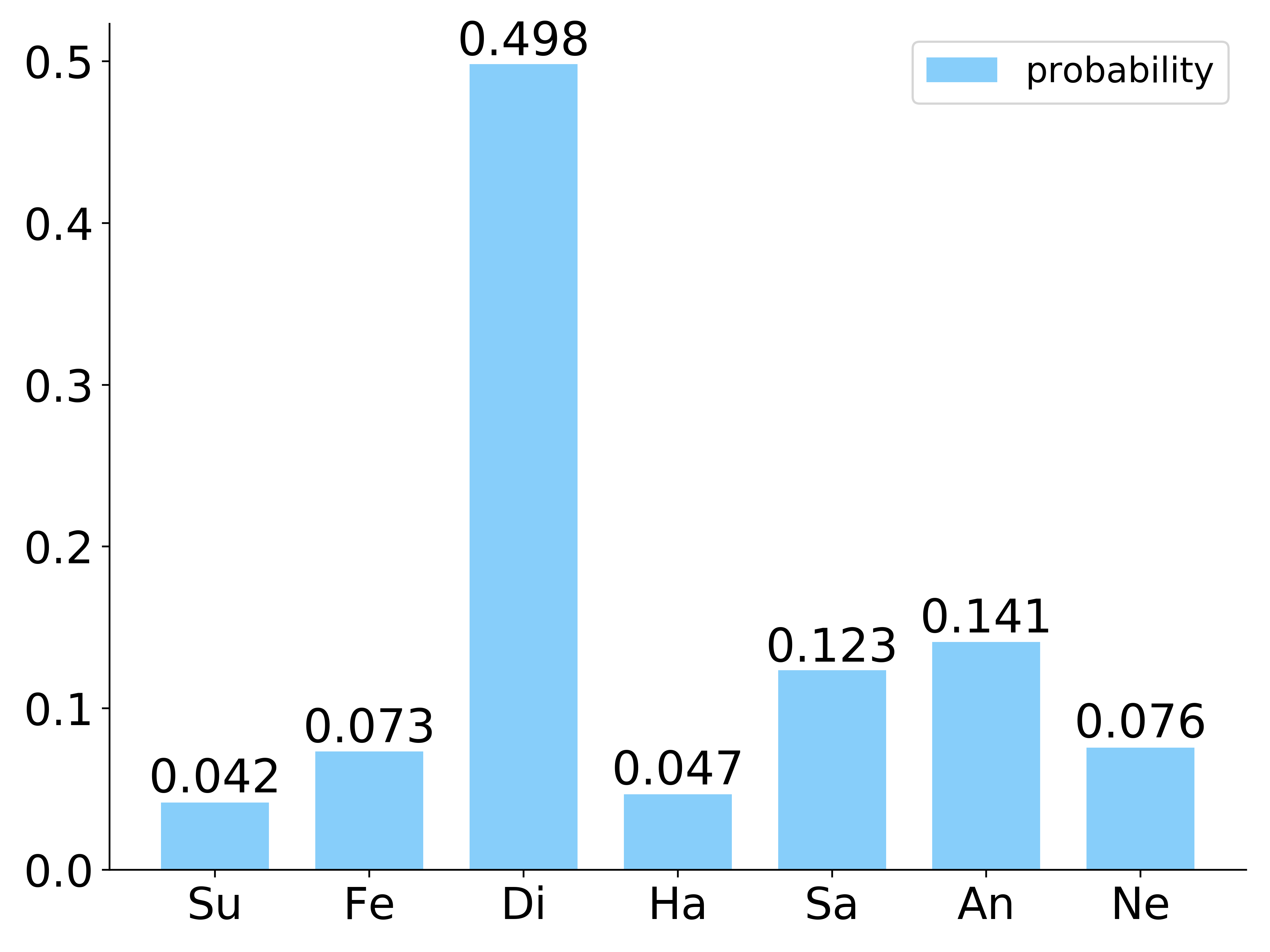}
	\includegraphics[width=0.28\columnwidth]{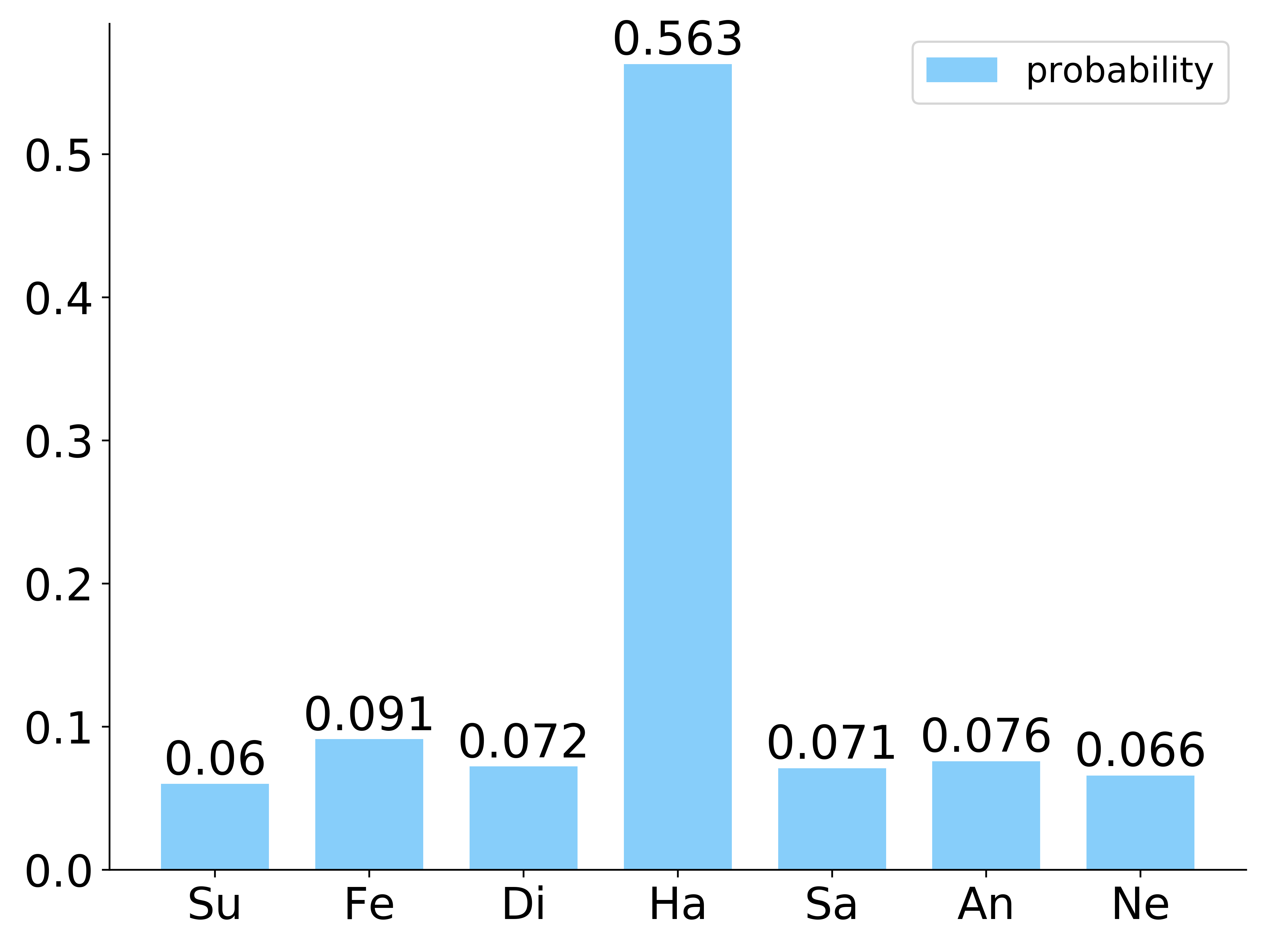}
	\includegraphics[width=0.28\columnwidth]{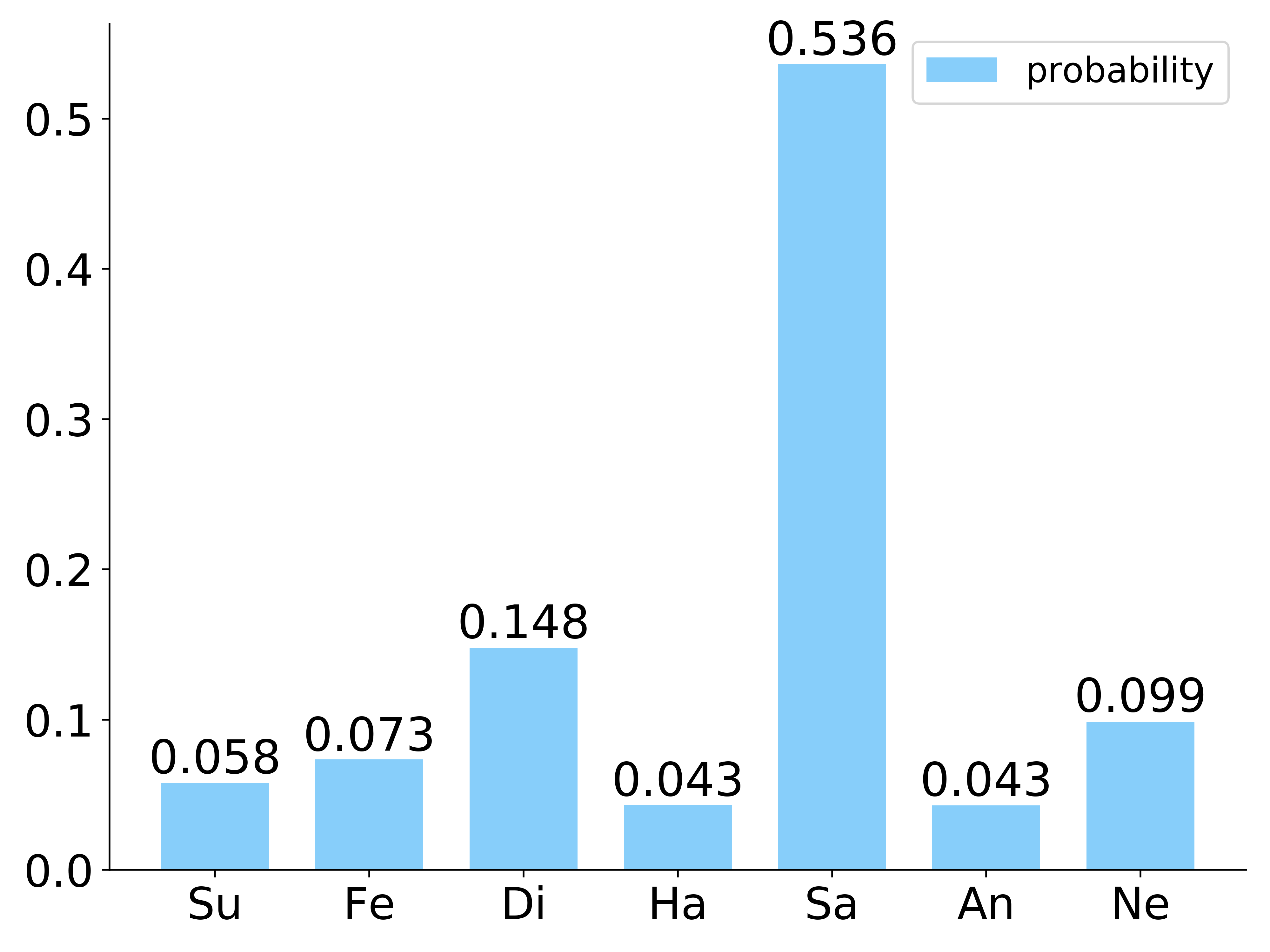}
	\includegraphics[width=0.28\columnwidth]{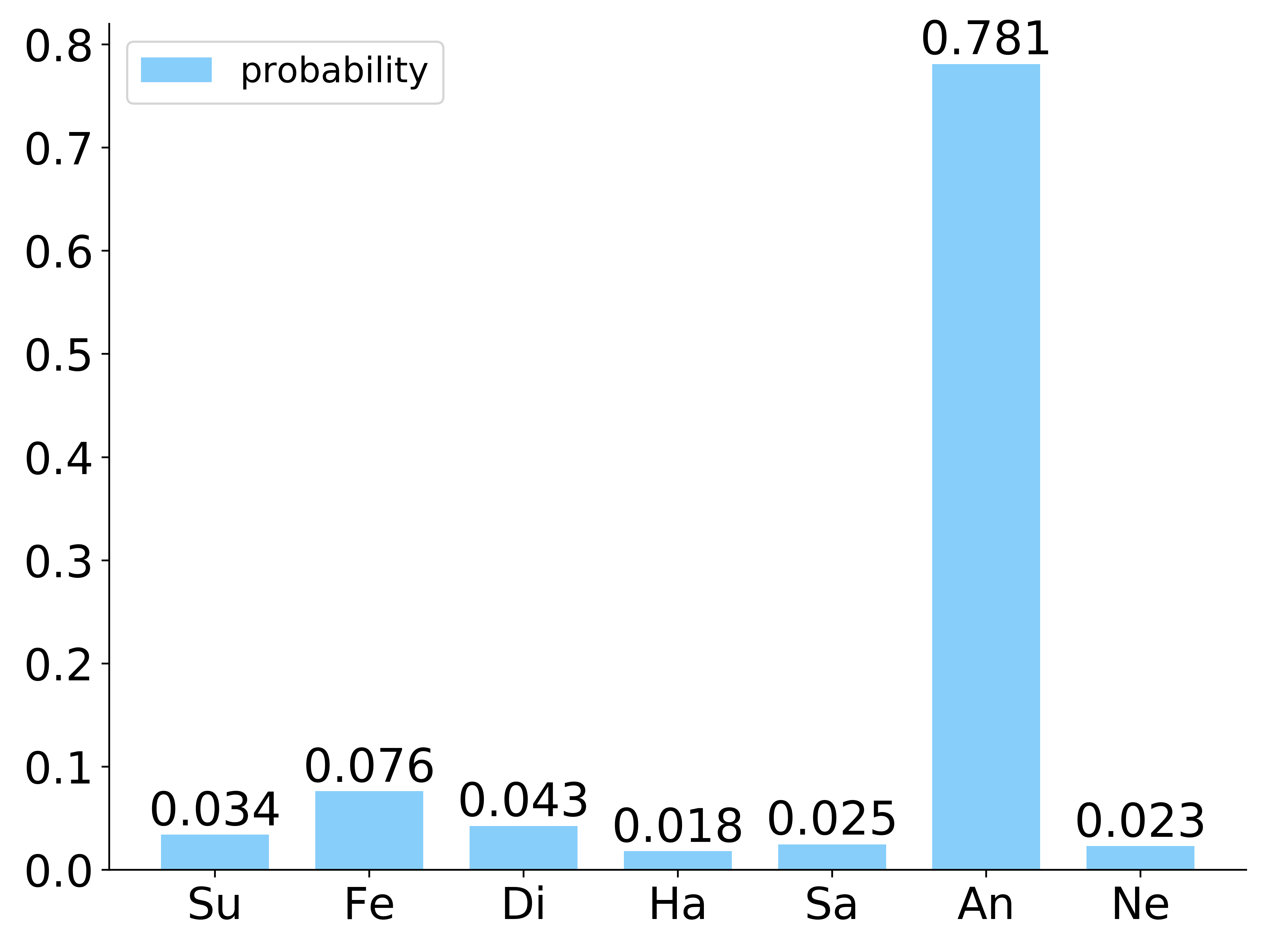}
	\includegraphics[width=0.28\columnwidth]{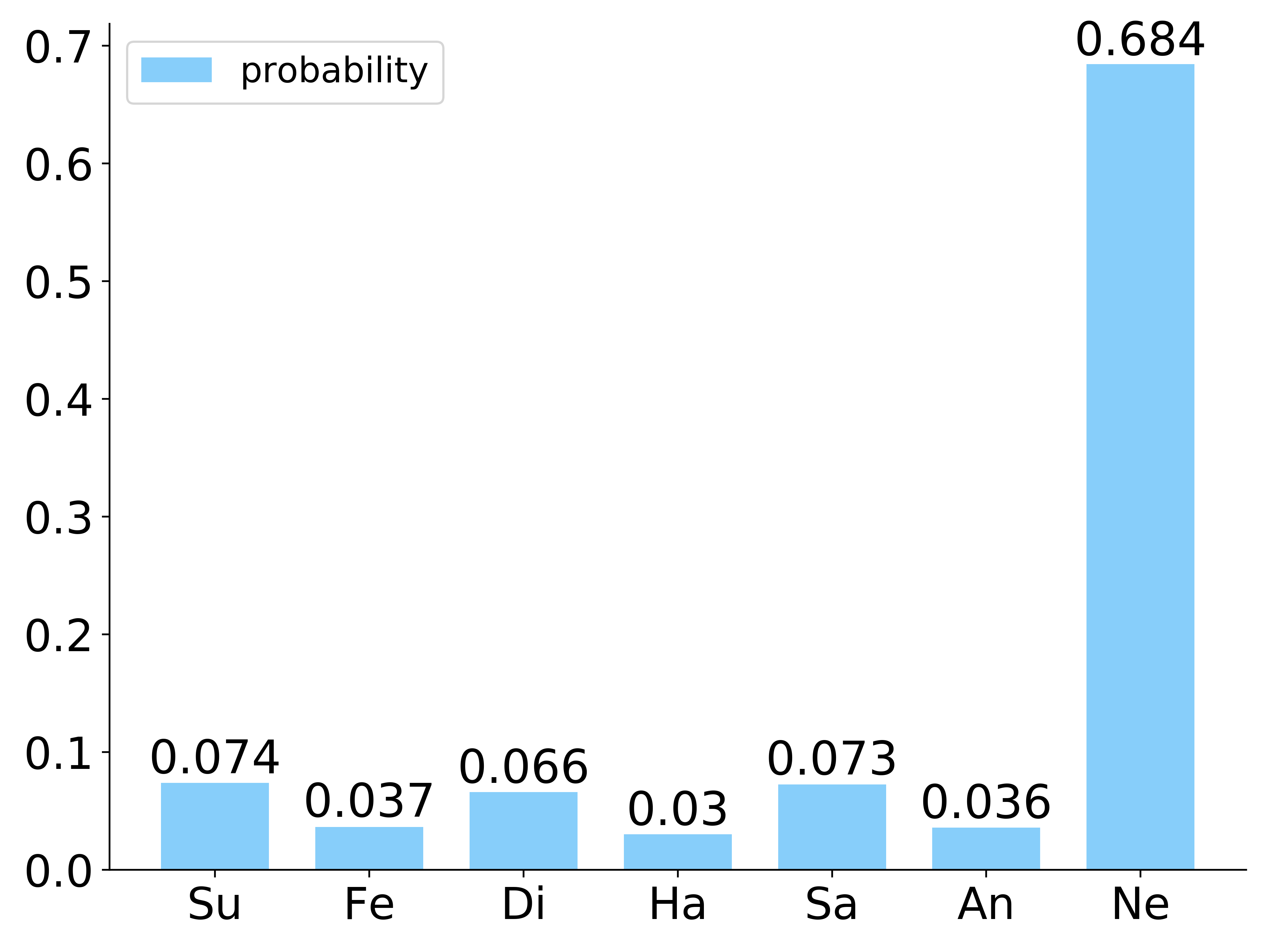}
	\\
	\includegraphics[width=0.28\columnwidth]{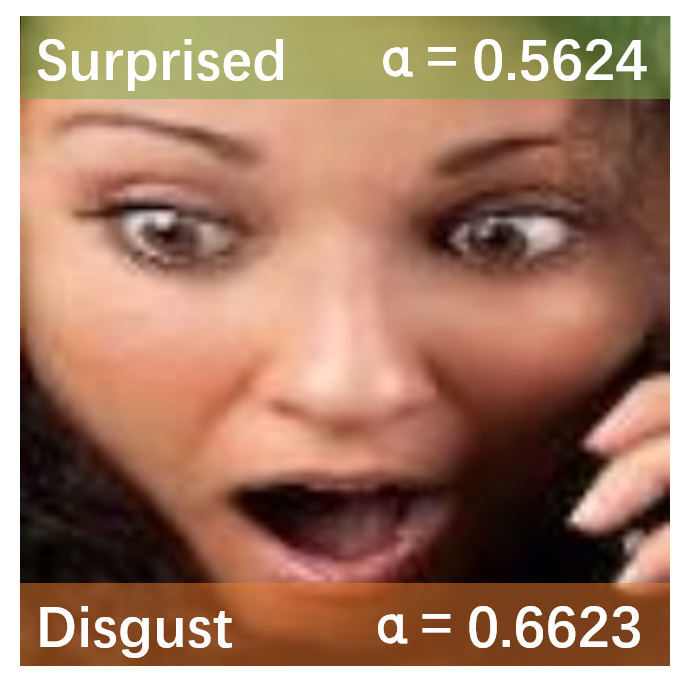}
	\includegraphics[width=0.28\columnwidth]{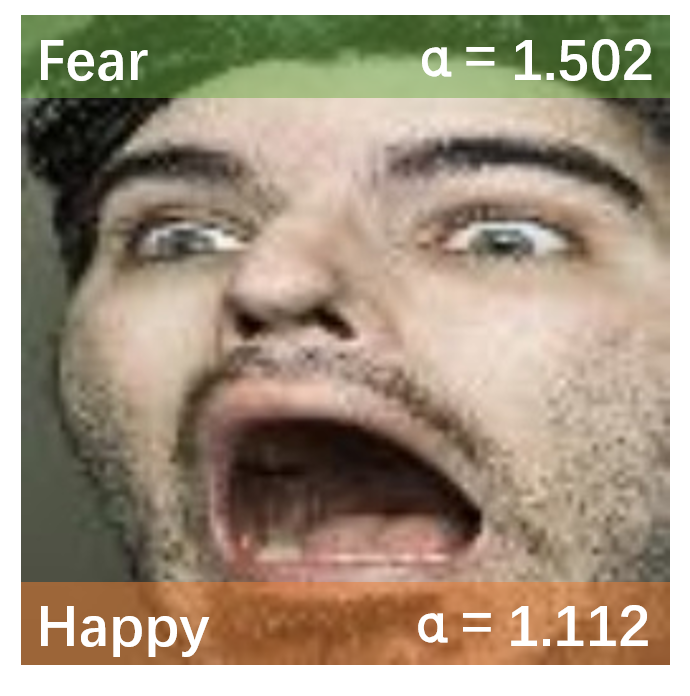}
	\includegraphics[width=0.28\columnwidth]{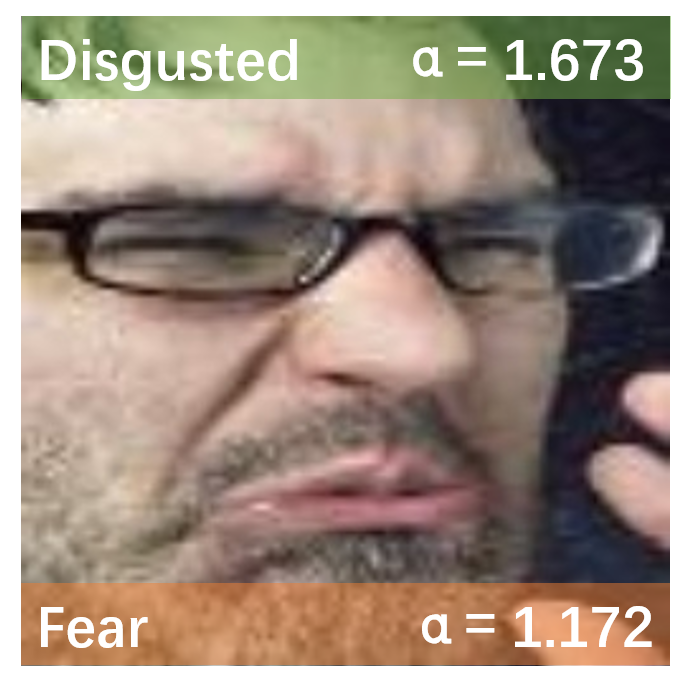}
	\includegraphics[width=0.28\columnwidth]{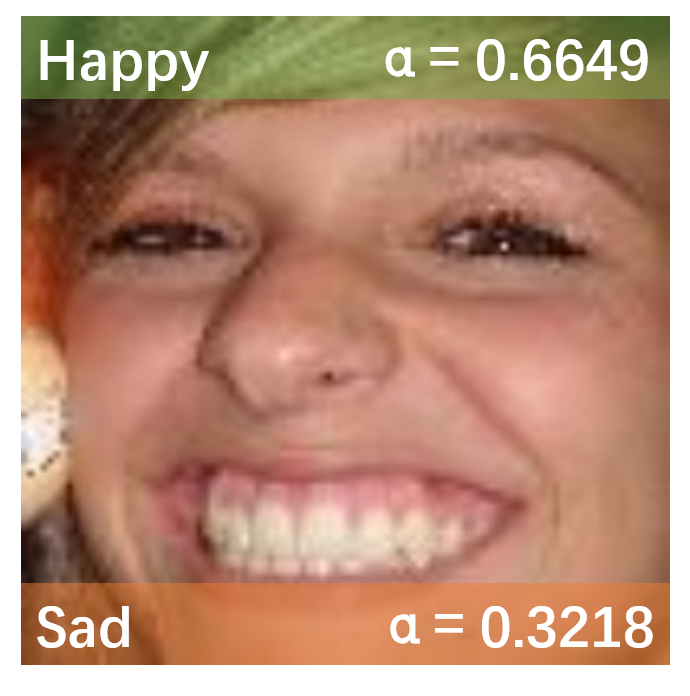}
	\includegraphics[width=0.28\columnwidth]{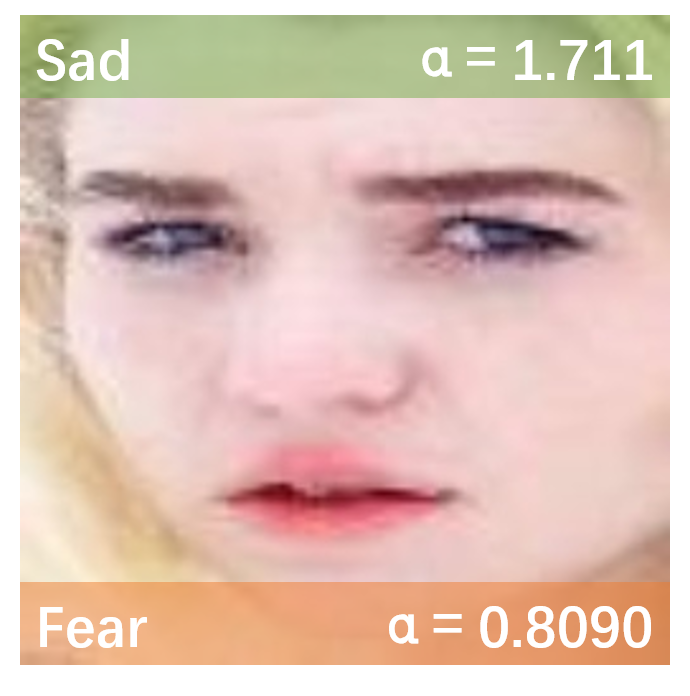}
	\includegraphics[width=0.28\columnwidth]{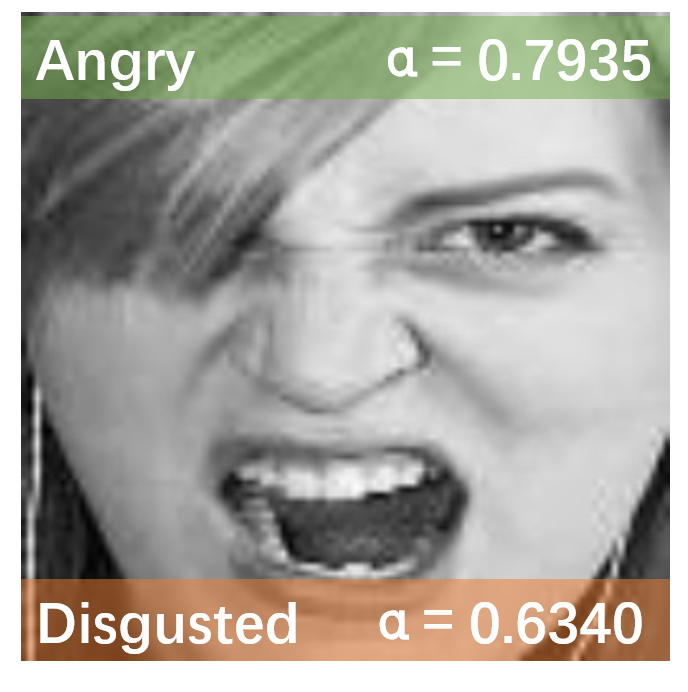}
	\includegraphics[width=0.28\columnwidth]{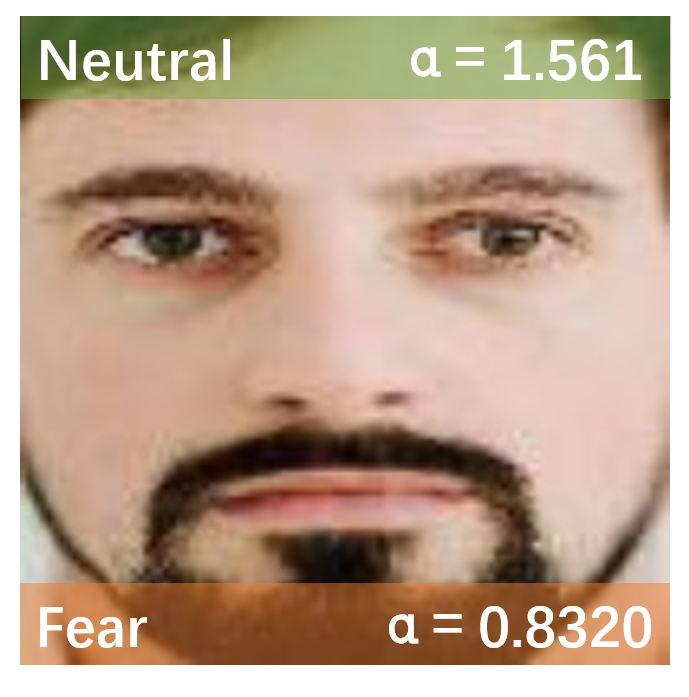}\\
	\includegraphics[width=0.28\columnwidth]{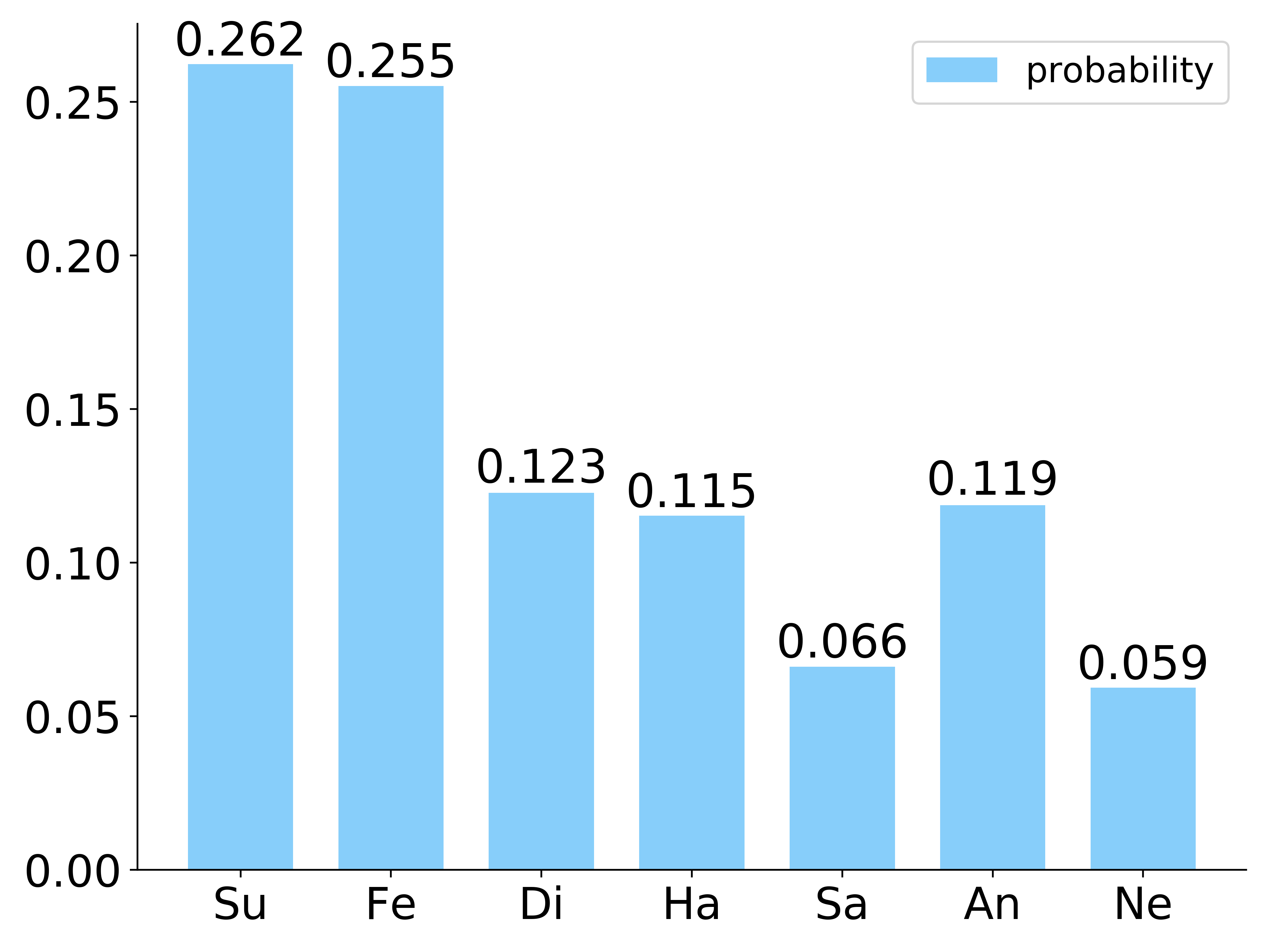}
	\includegraphics[width=0.28\columnwidth]{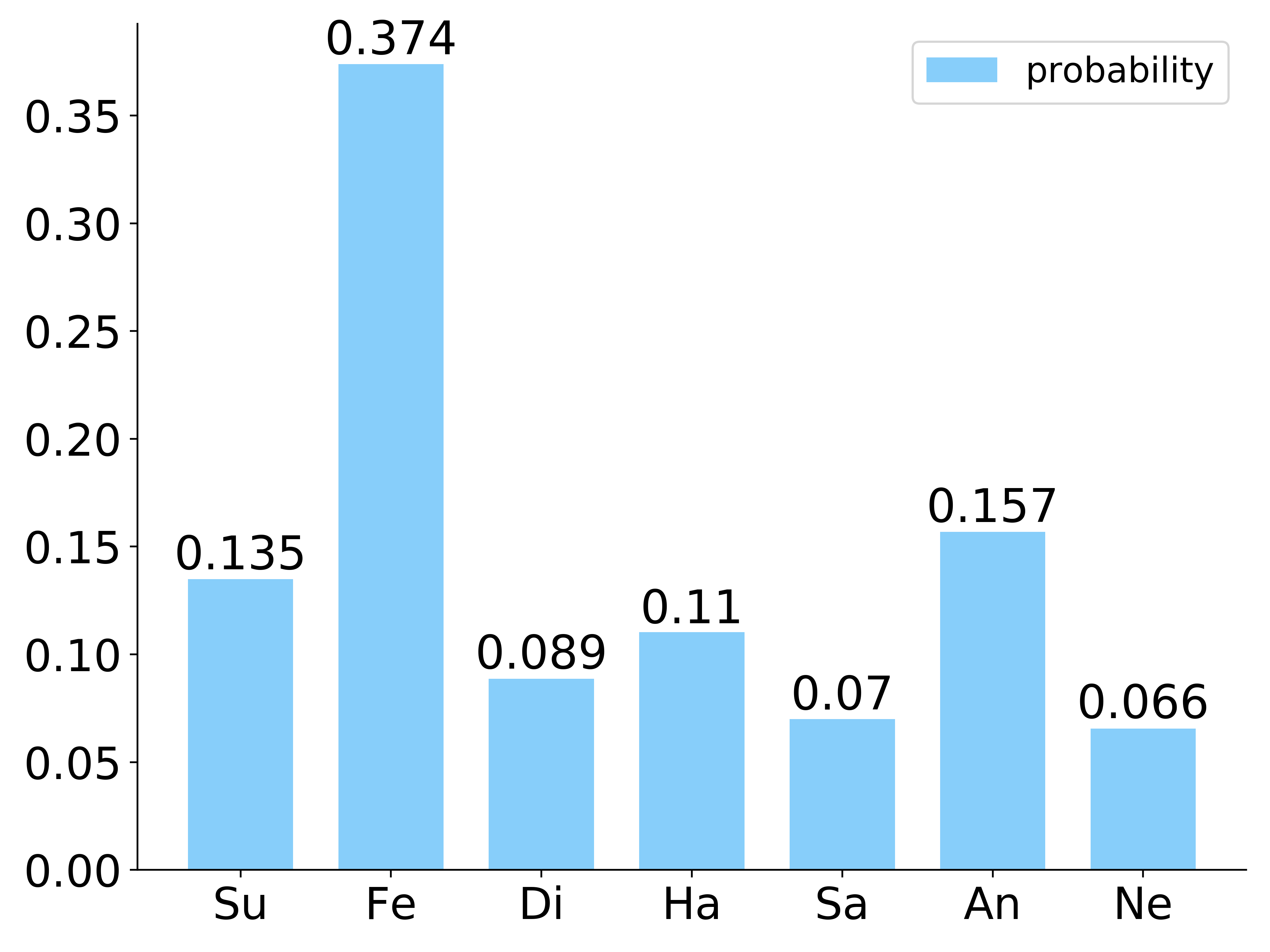}
	\includegraphics[width=0.28\columnwidth]{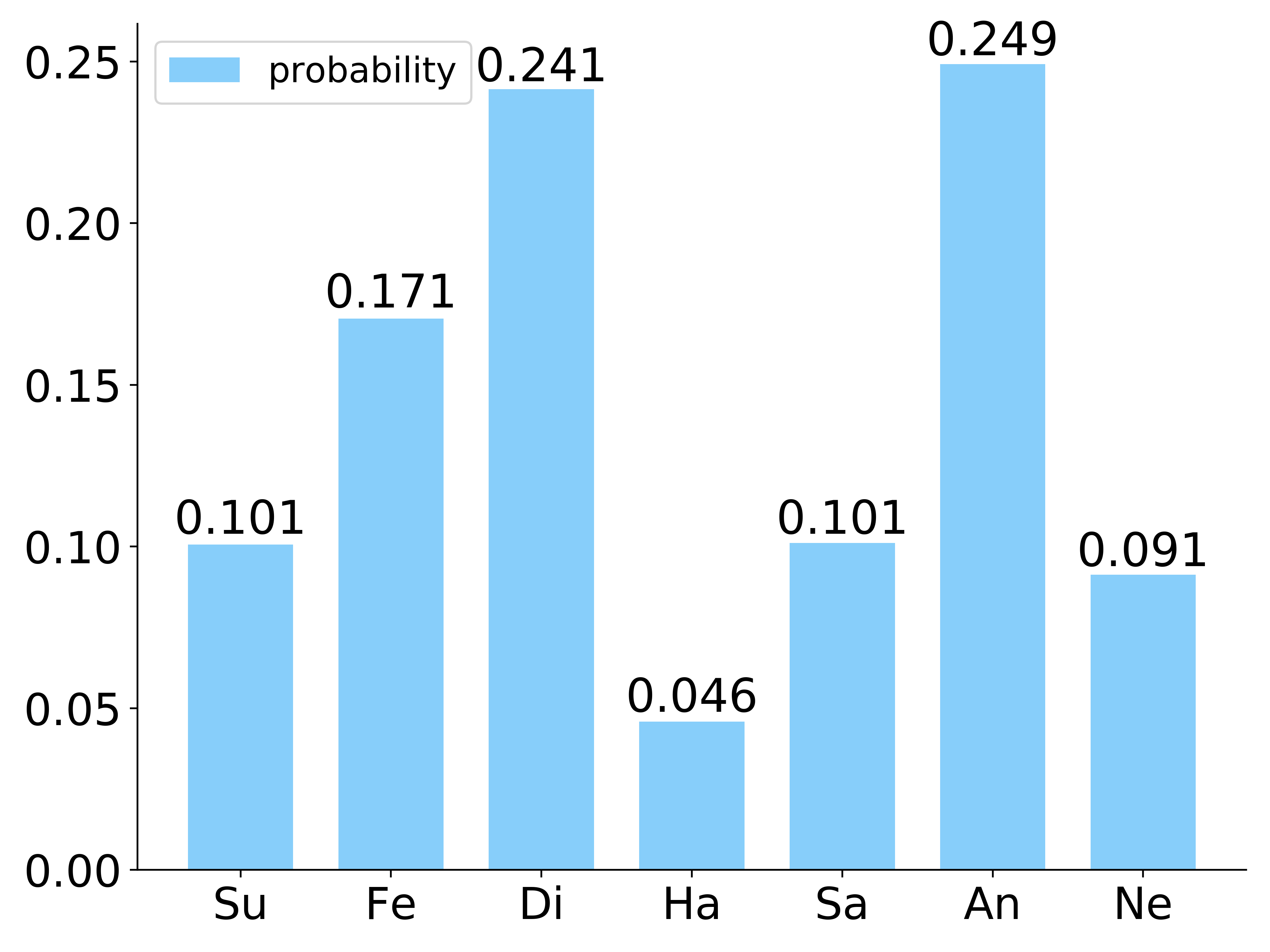}
	\includegraphics[width=0.28\columnwidth]{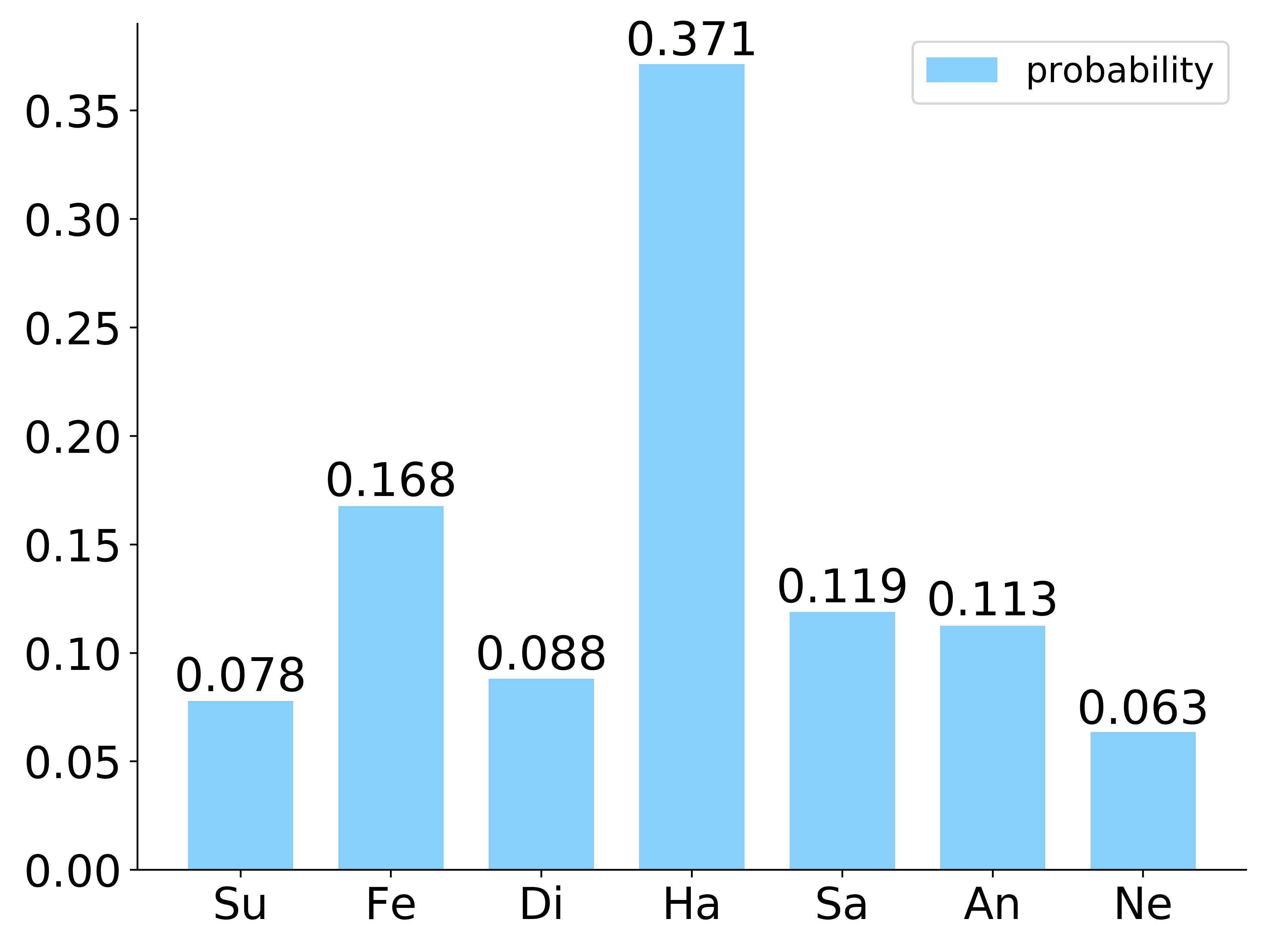}
	\includegraphics[width=0.28\columnwidth]{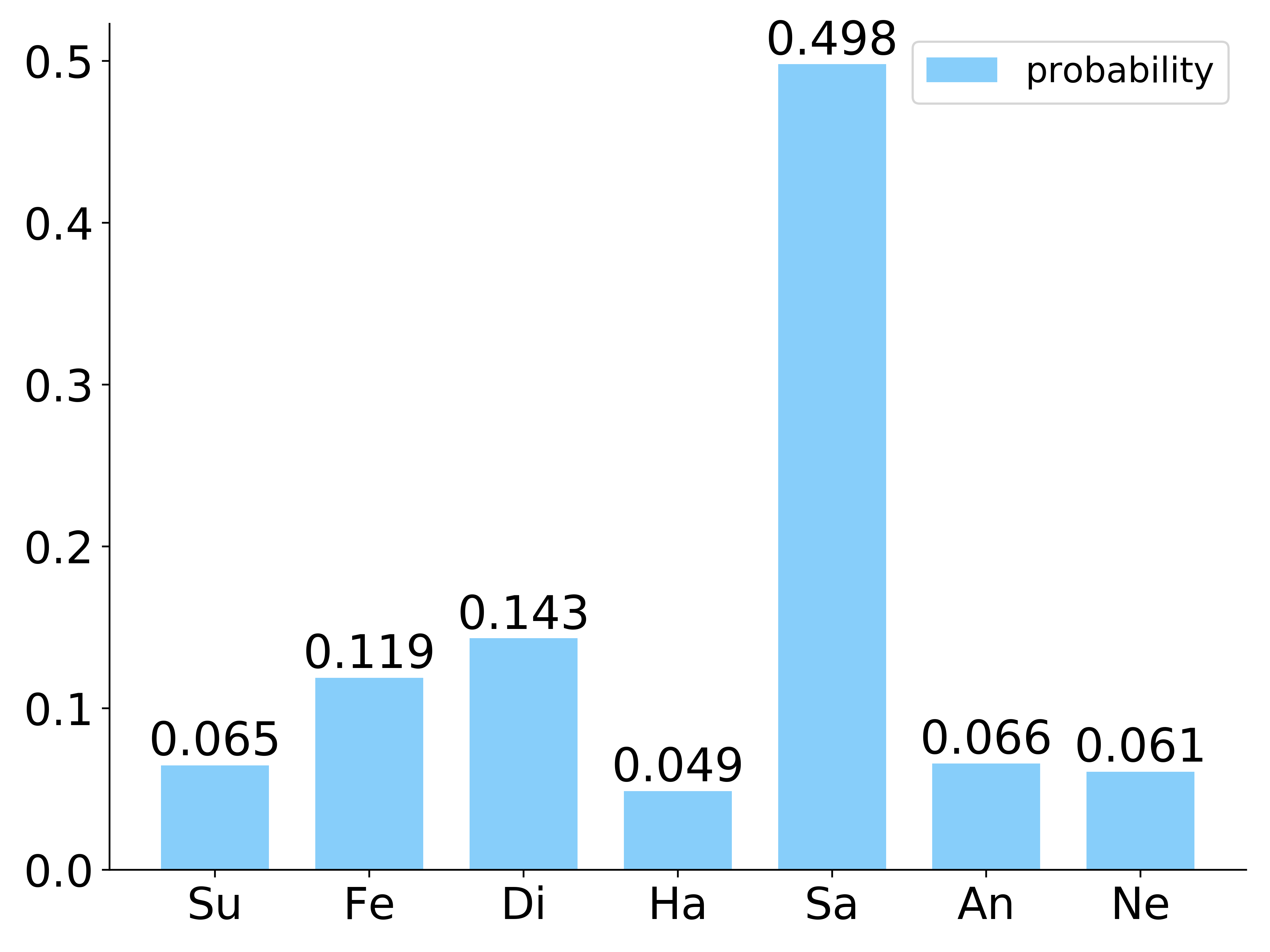}
	\includegraphics[width=0.28\columnwidth]{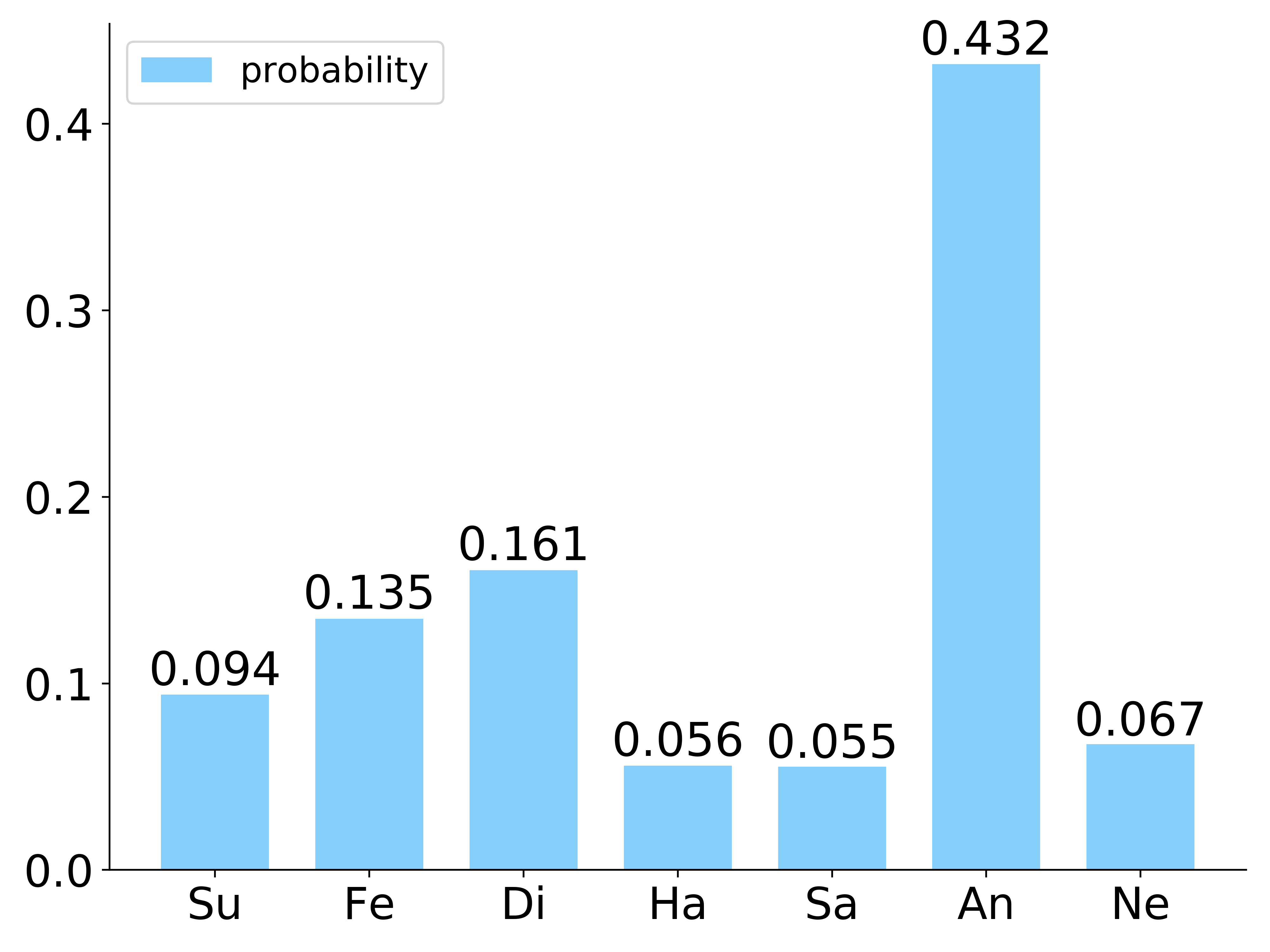}
	\includegraphics[width=0.28\columnwidth]{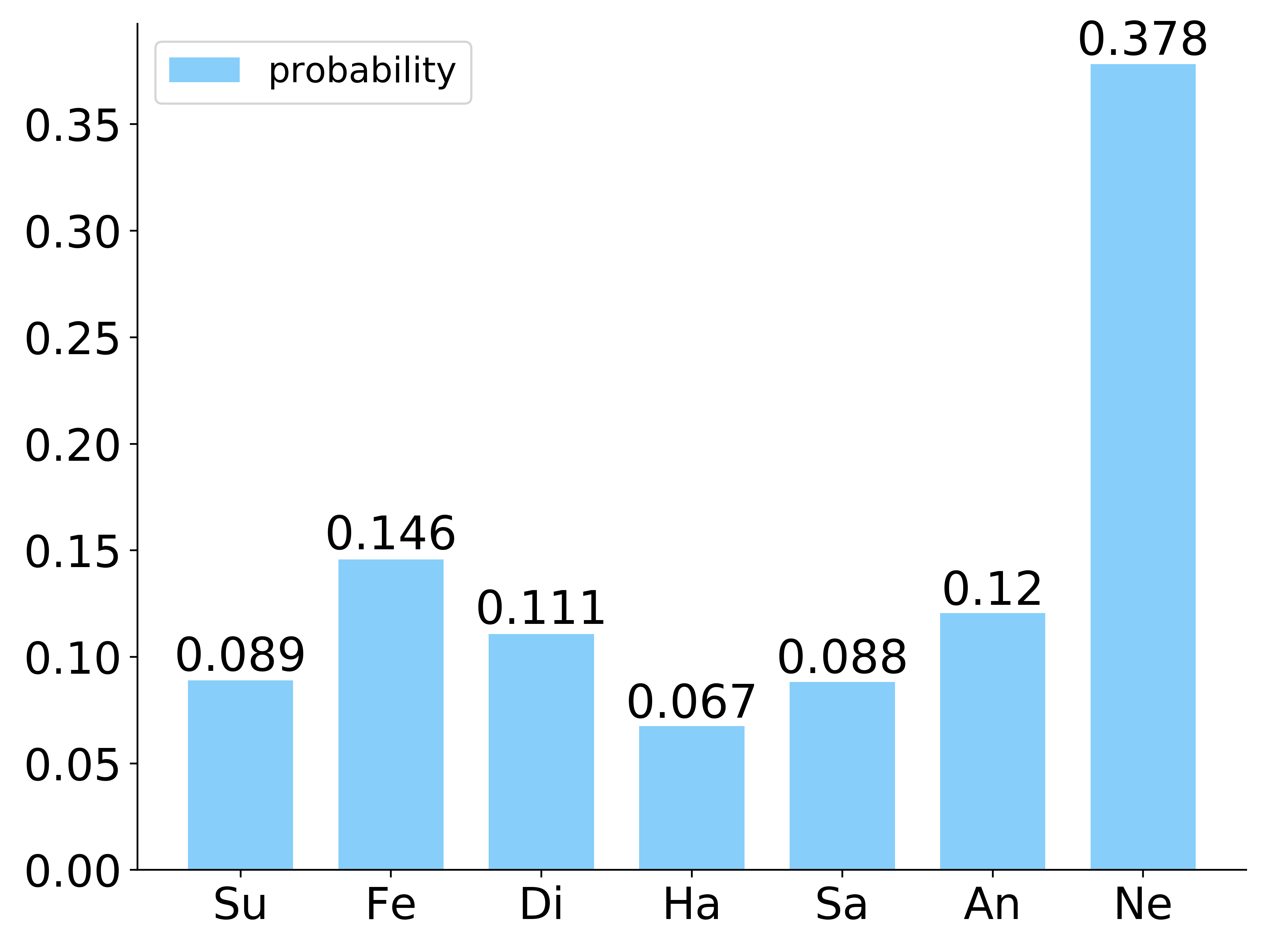}\\	
	\rule[0.2\baselineskip]{18cm}{1pt}
	\vspace{0.5cm}	
	\includegraphics[width=0.28\columnwidth]{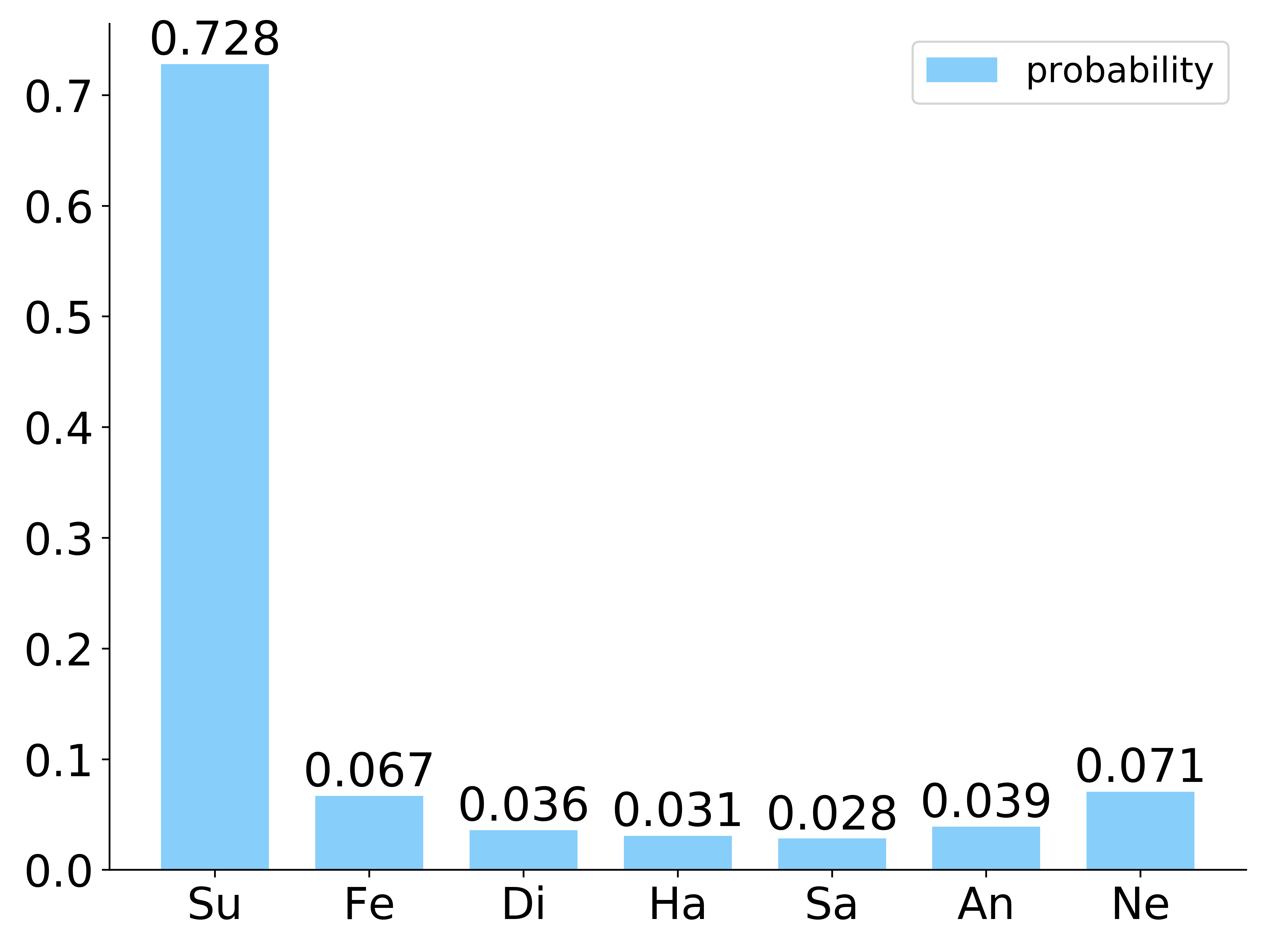}
	\includegraphics[width=0.28\columnwidth]{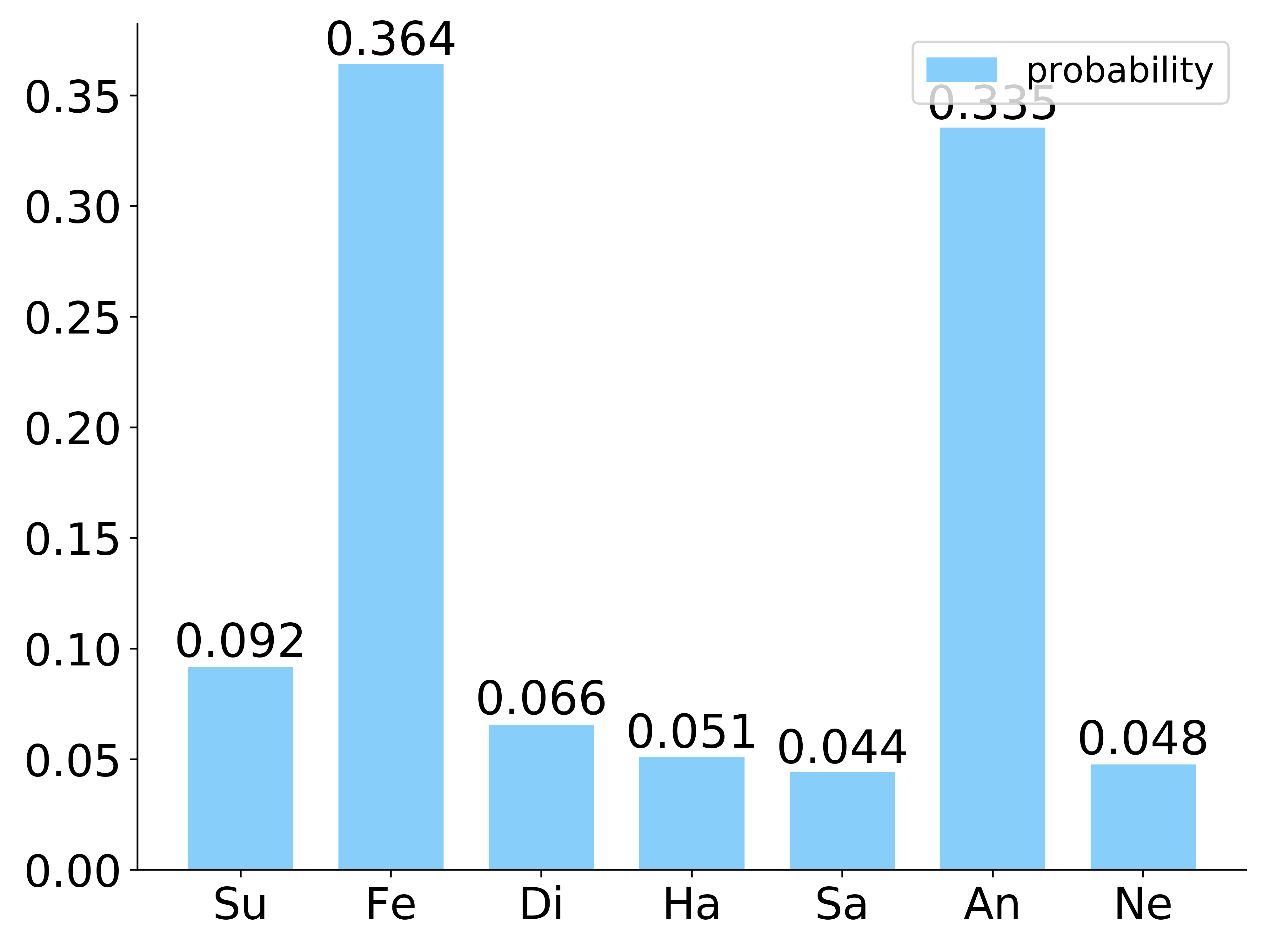}
	\includegraphics[width=0.28\columnwidth]{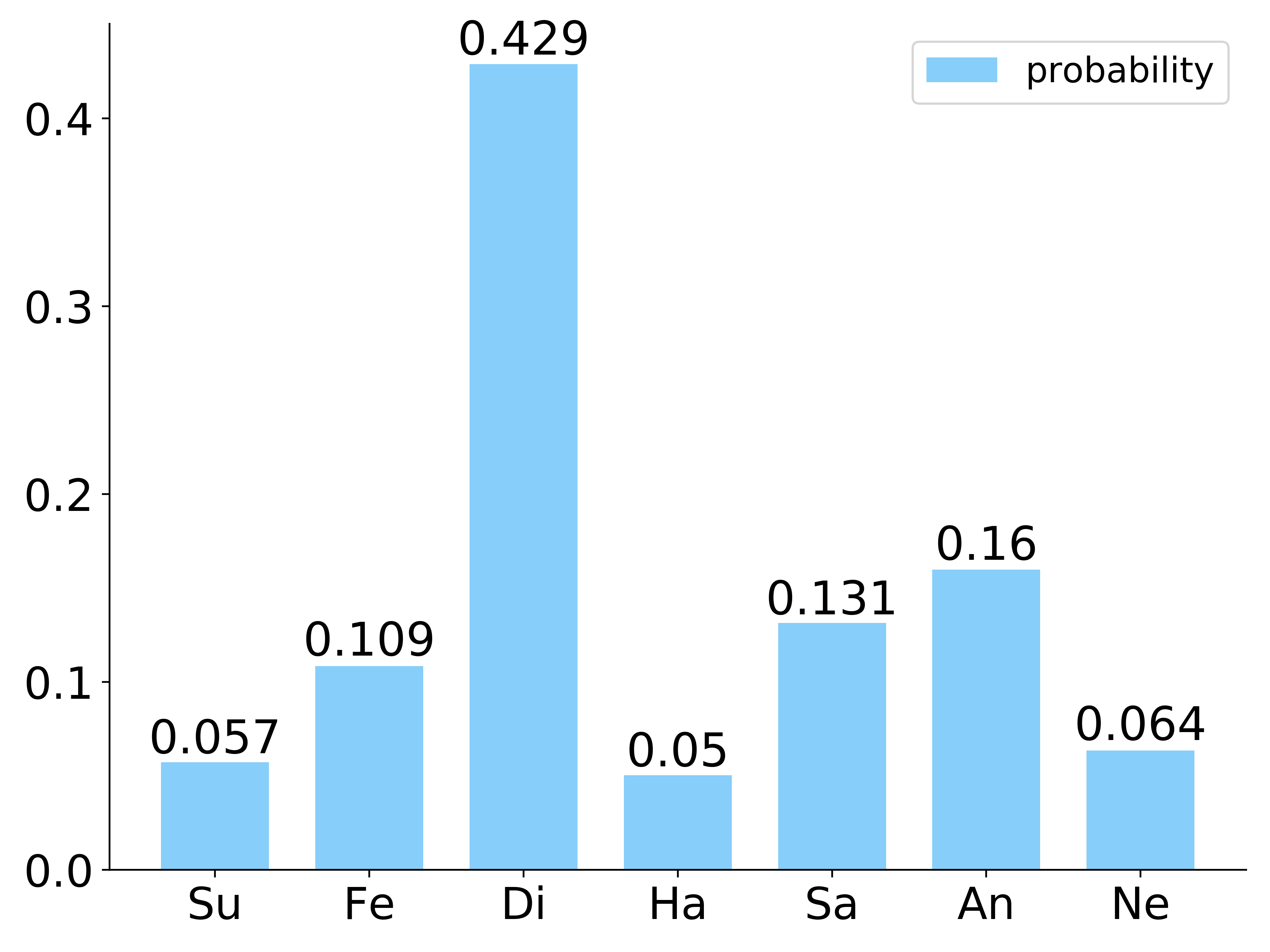}
	\includegraphics[width=0.28\columnwidth]{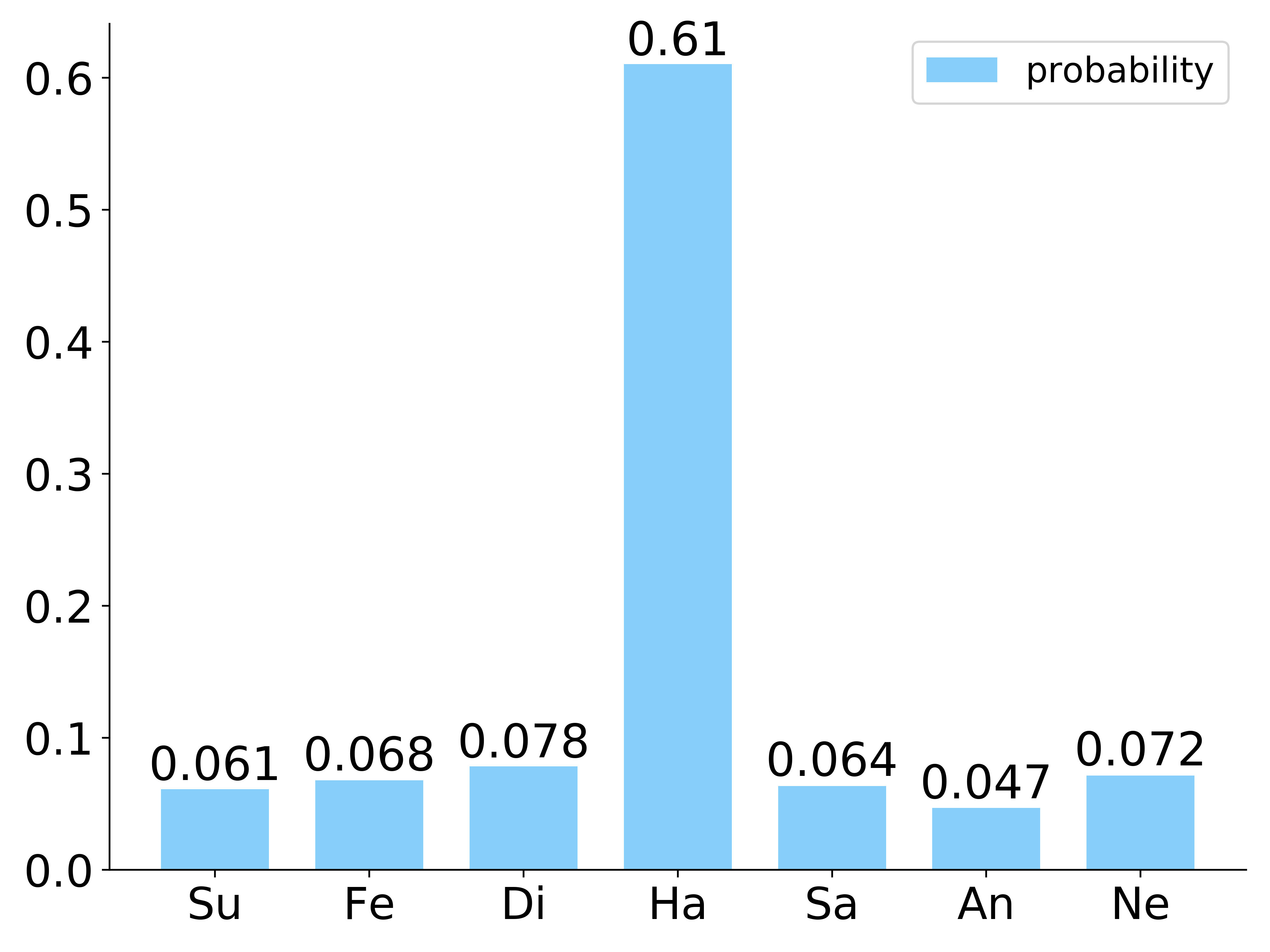}
	\includegraphics[width=0.28\columnwidth]{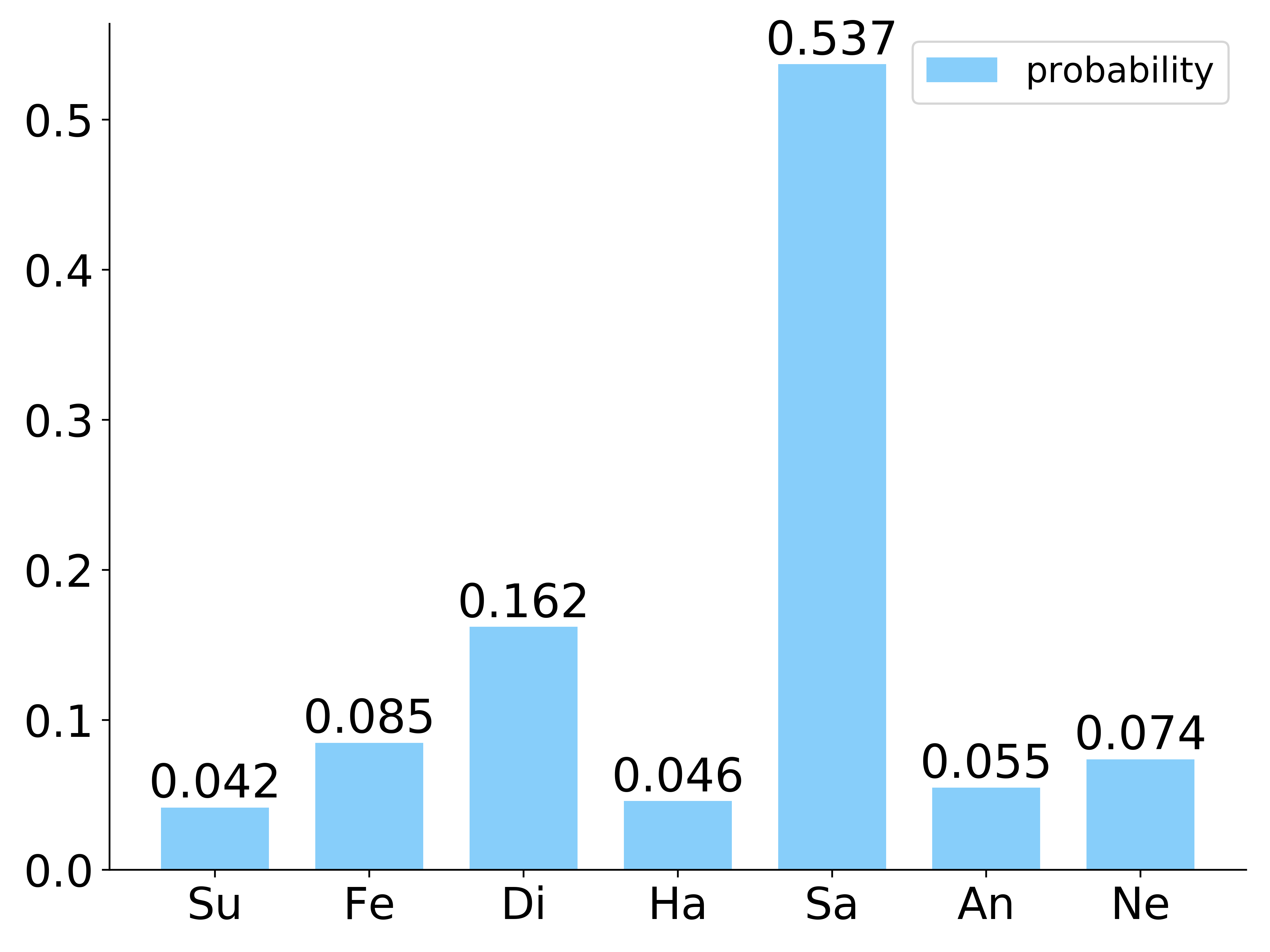}
	\includegraphics[width=0.28\columnwidth]{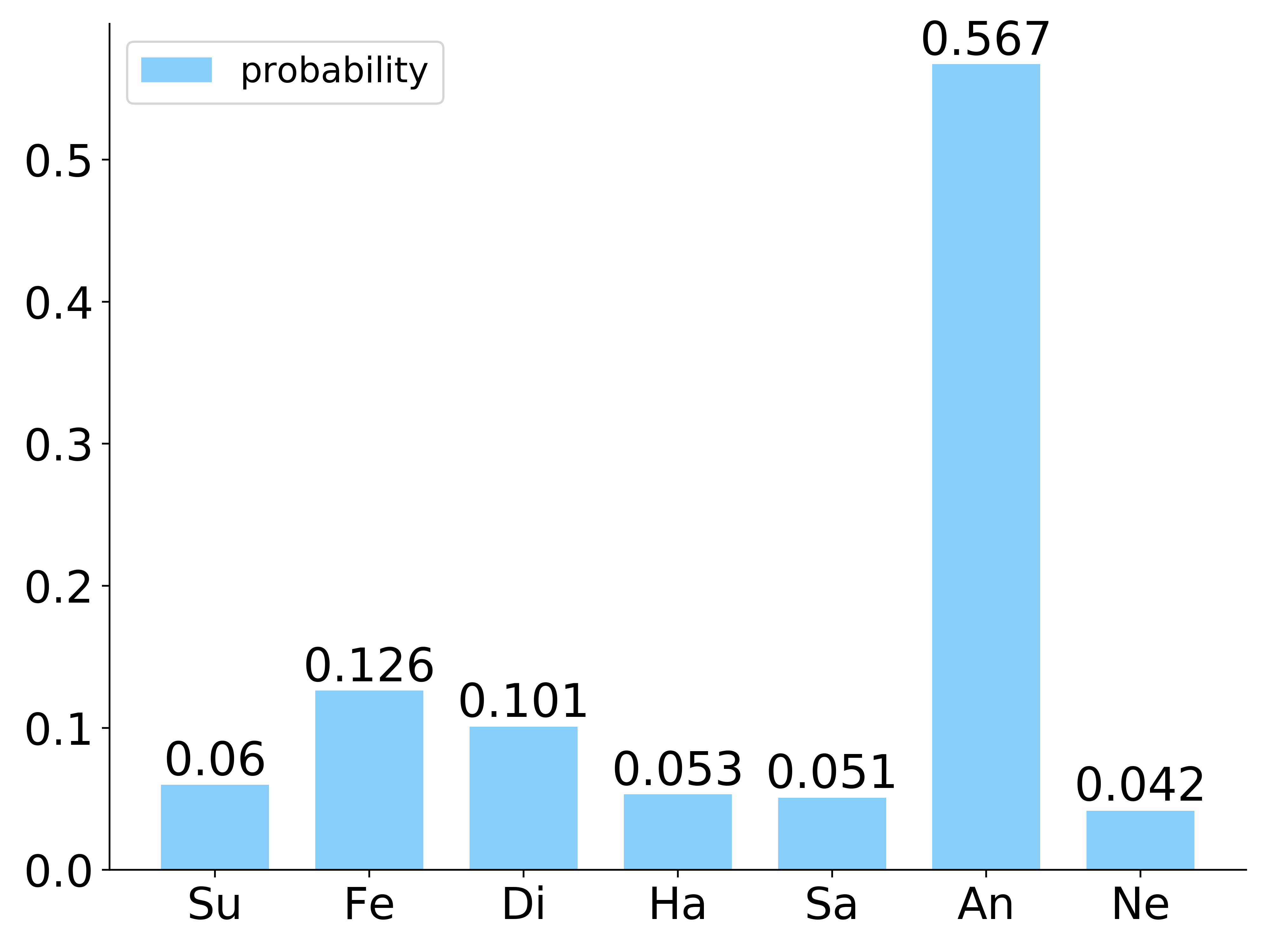}
	\includegraphics[width=0.28\columnwidth]{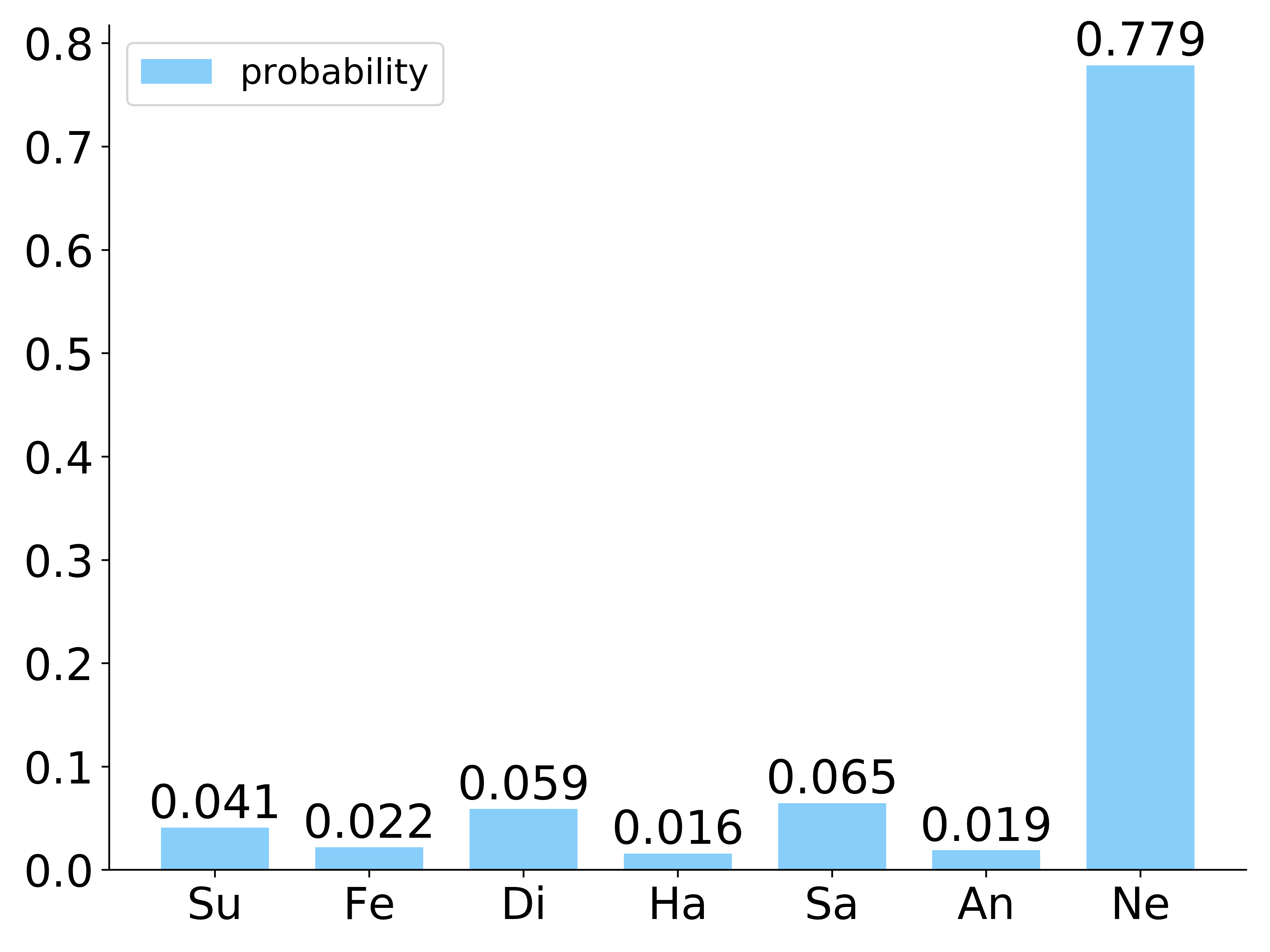}	\\
	\includegraphics[width=0.28\columnwidth]{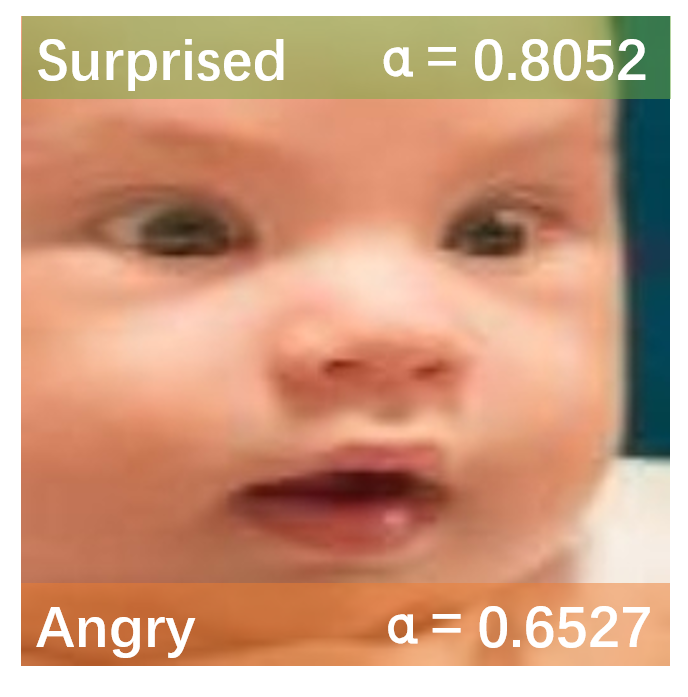}
	\includegraphics[width=0.28\columnwidth]{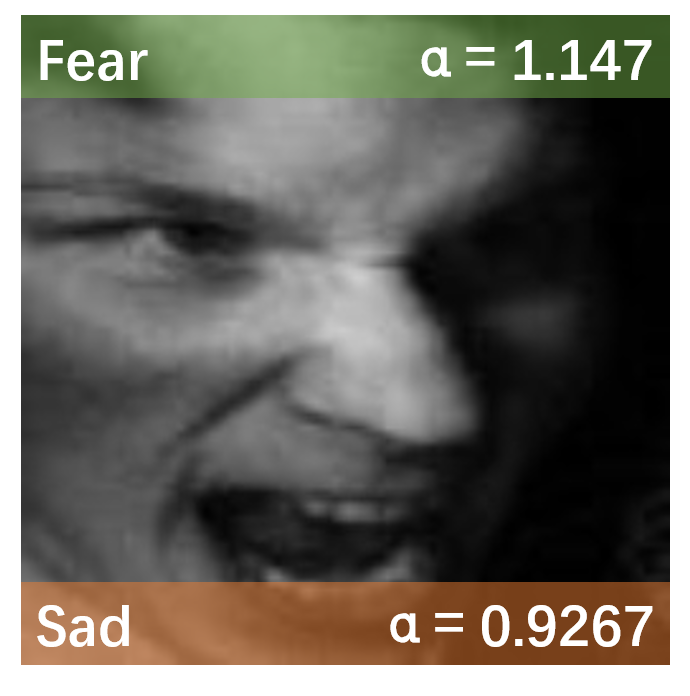}
	\includegraphics[width=0.28\columnwidth]{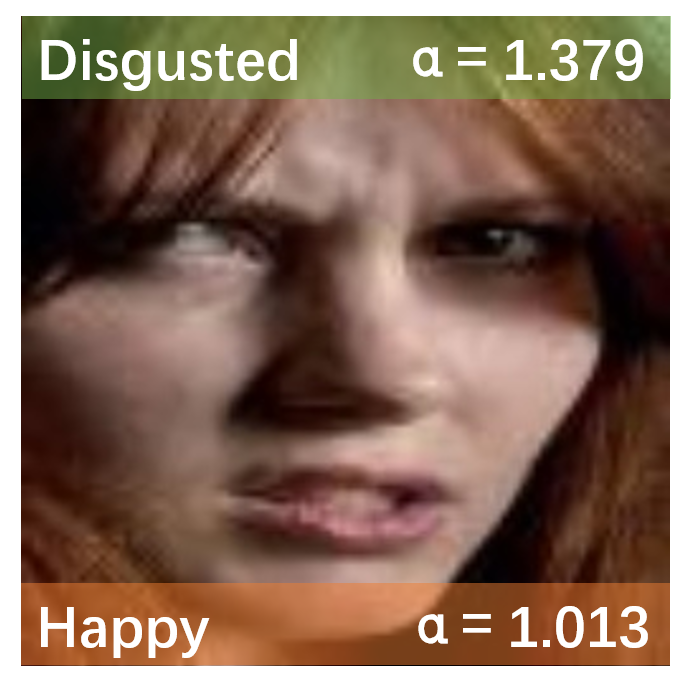}
	\includegraphics[width=0.28\columnwidth]{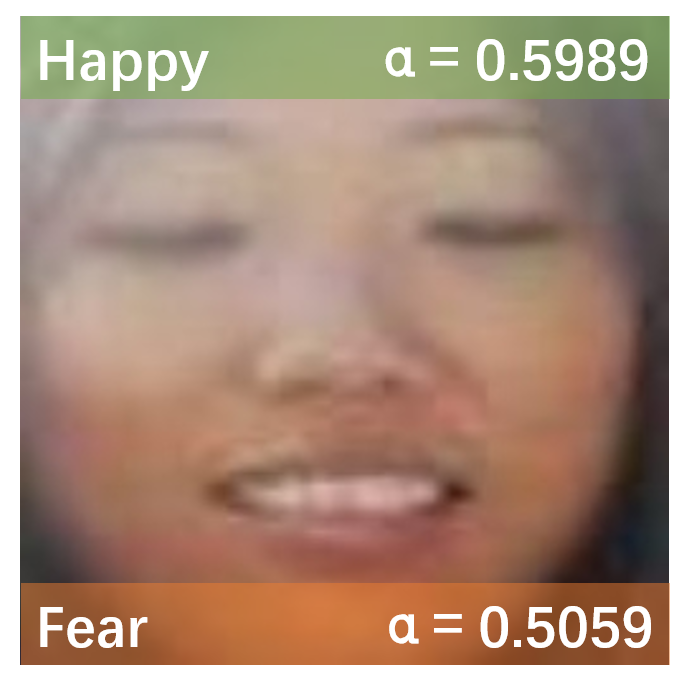}
	\includegraphics[width=0.28\columnwidth]{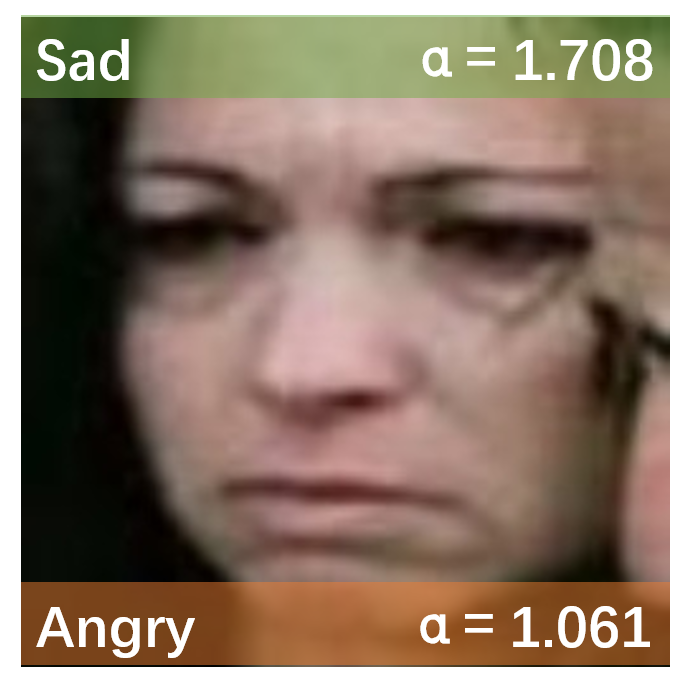}
	\includegraphics[width=0.28\columnwidth]{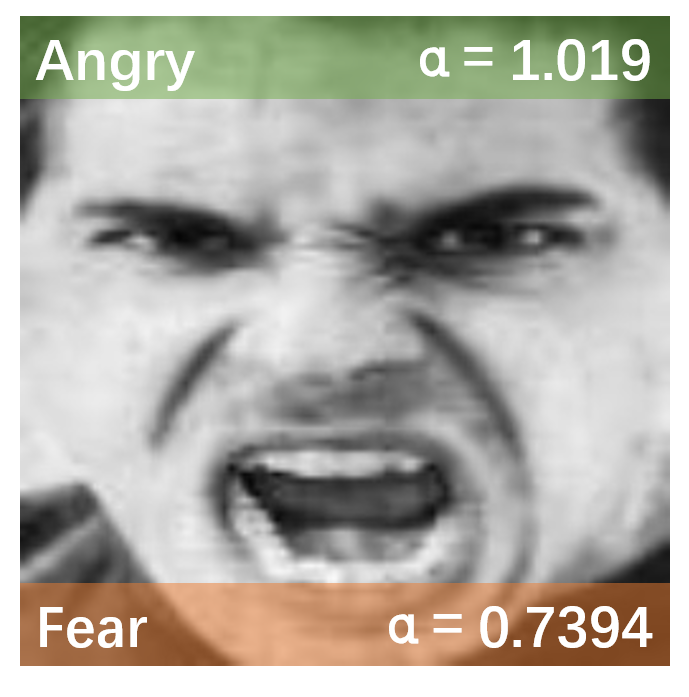}
	\includegraphics[width=0.28\columnwidth]{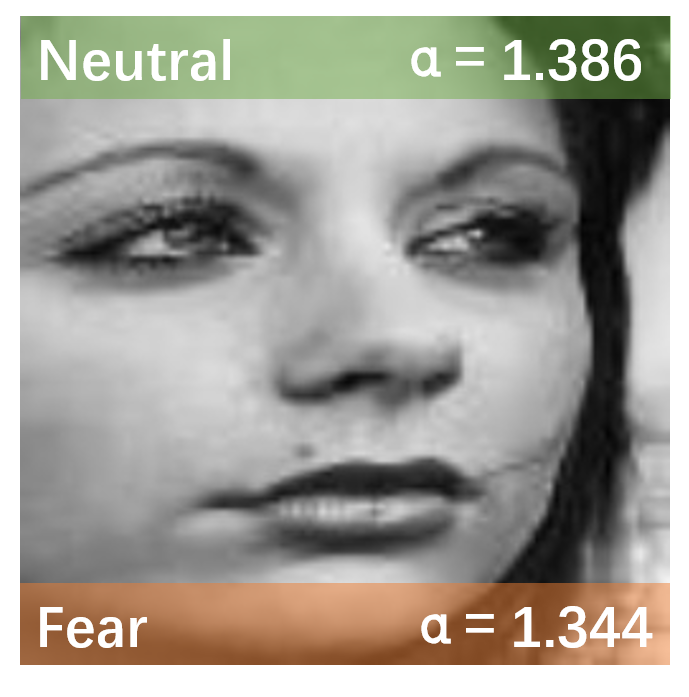}	\\
	\includegraphics[width=0.28\columnwidth]{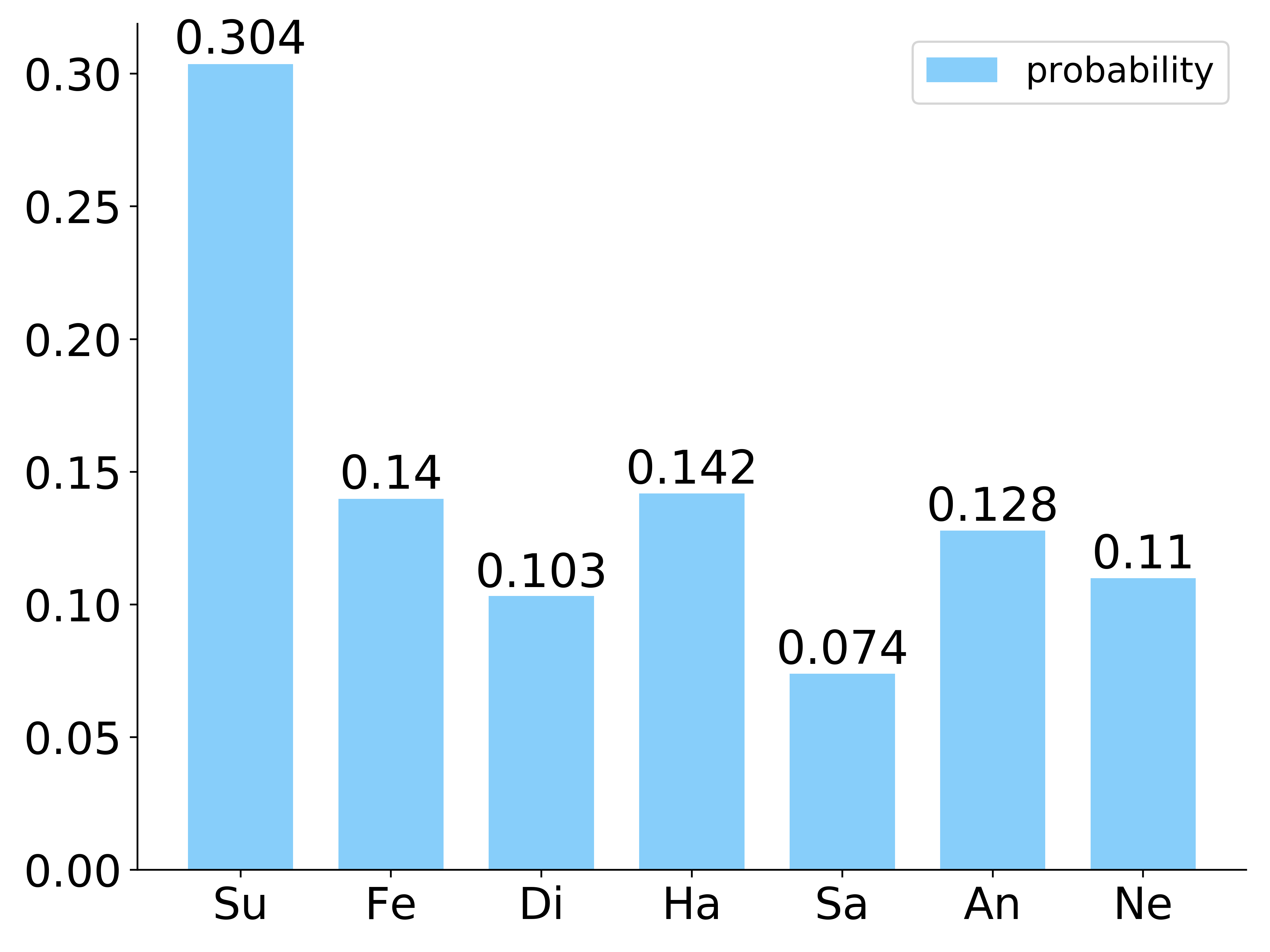}
	\includegraphics[width=0.28\columnwidth]{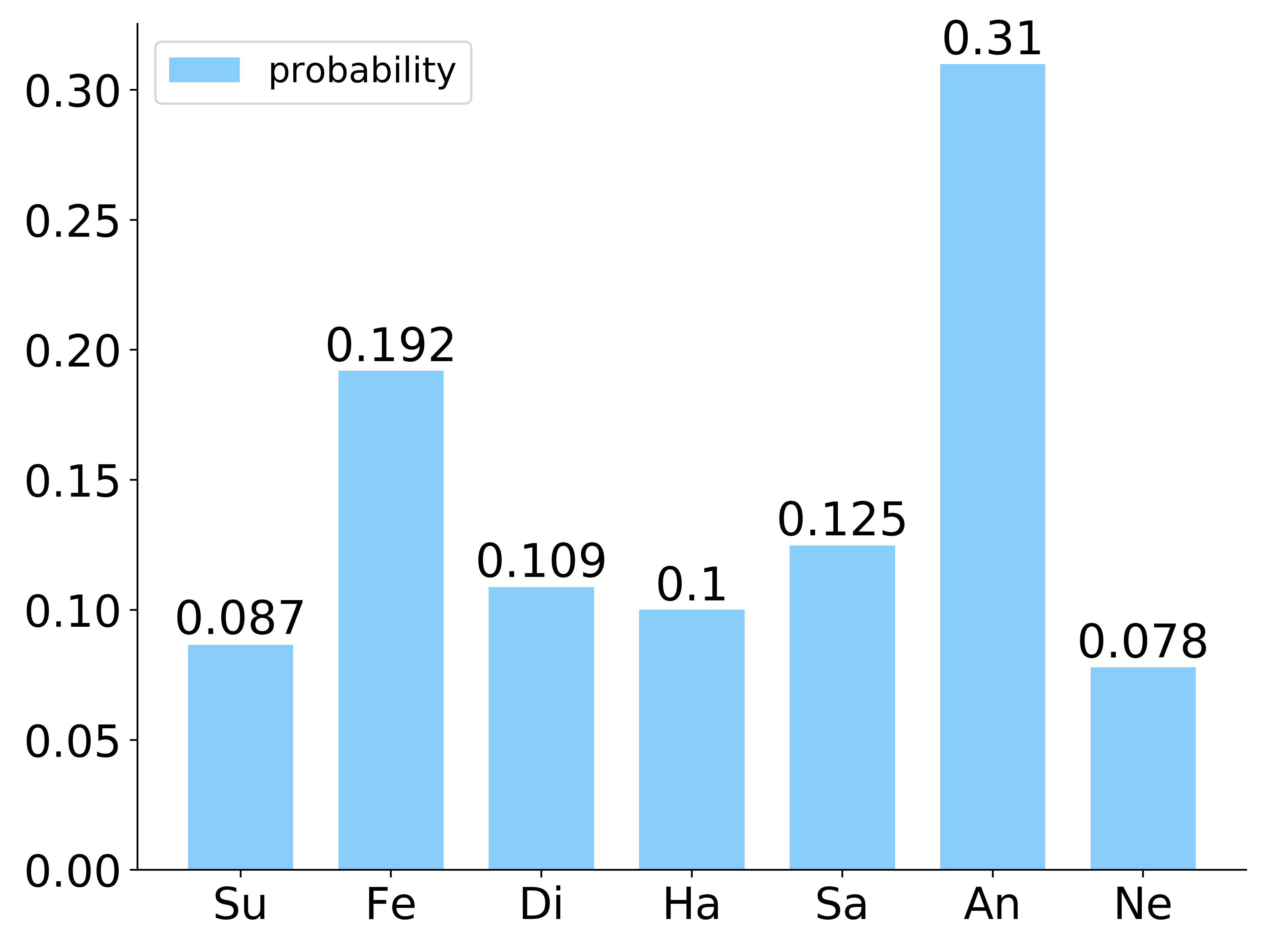}
	\includegraphics[width=0.28\columnwidth]{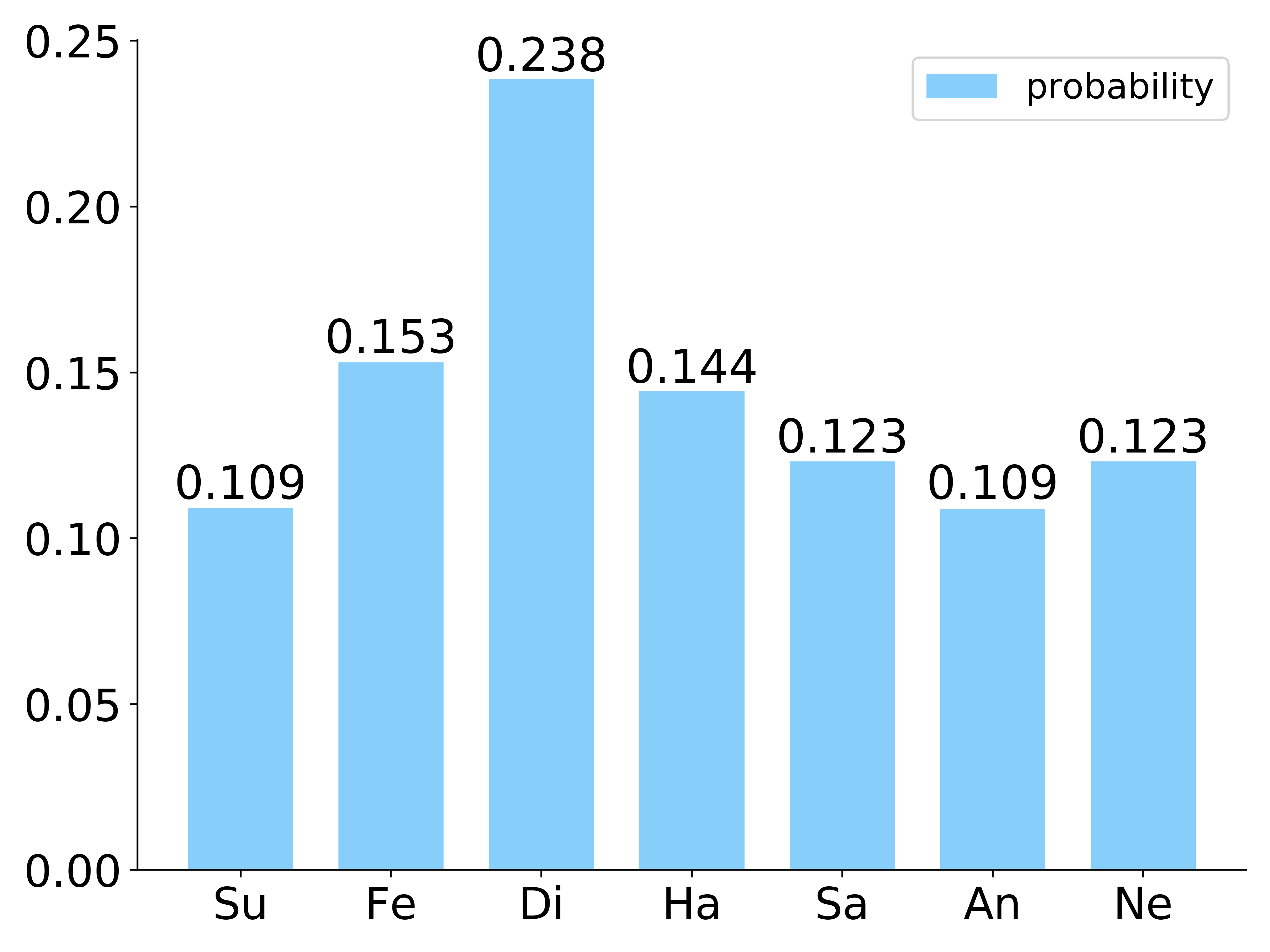}
	\includegraphics[width=0.28\columnwidth]{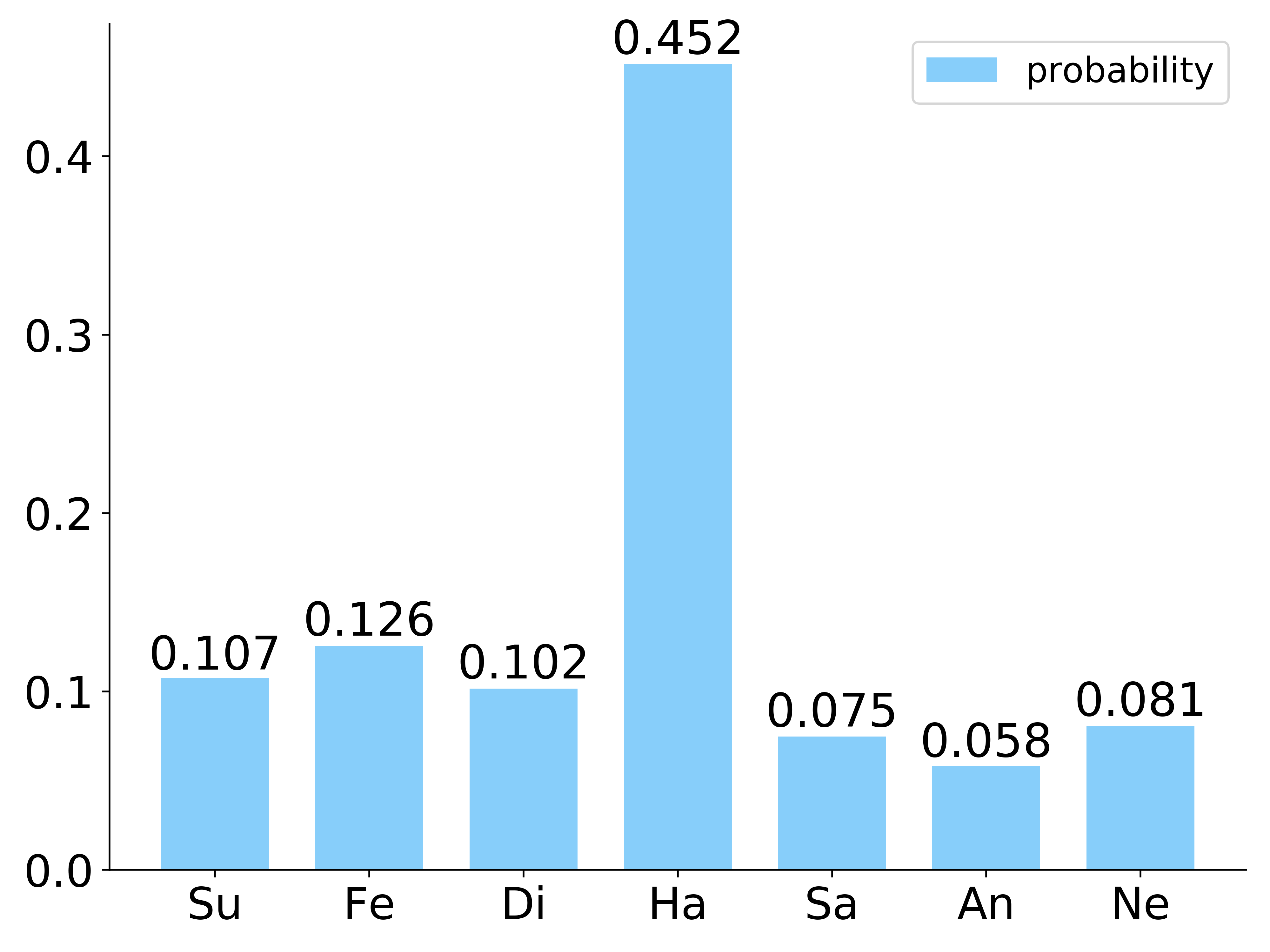}
	\includegraphics[width=0.28\columnwidth]{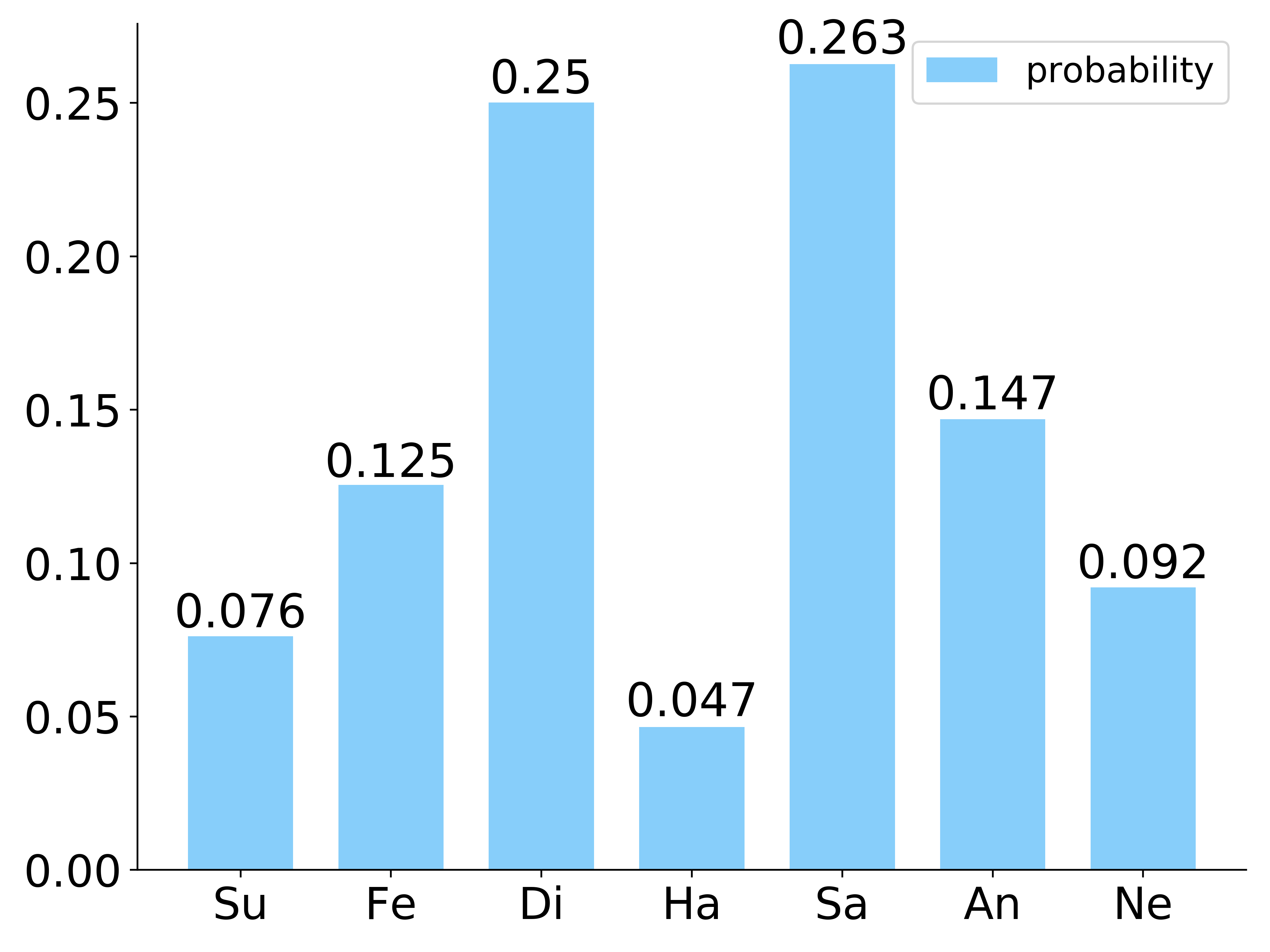}
	\includegraphics[width=0.28\columnwidth]{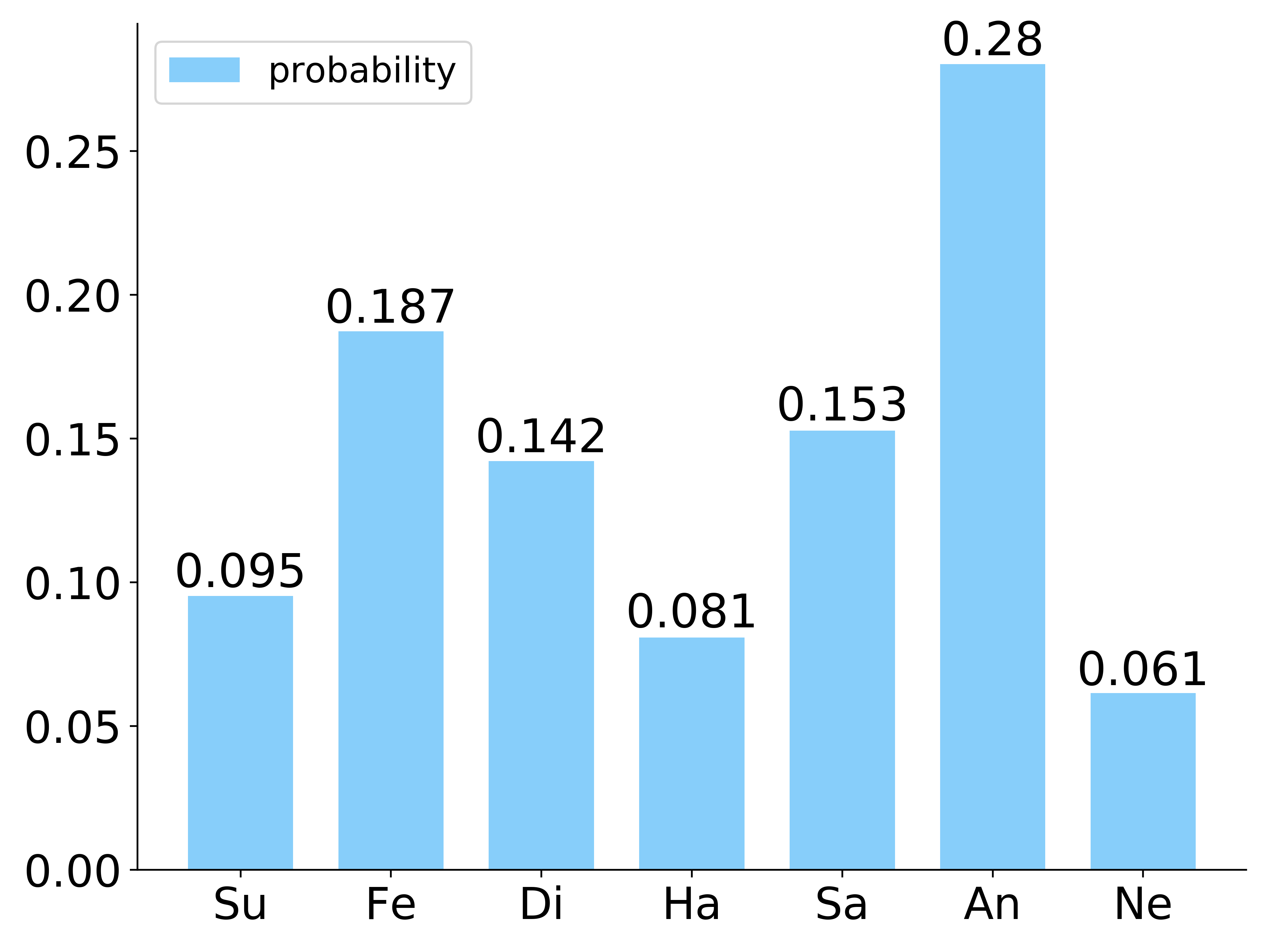}
	\includegraphics[width=0.28\columnwidth]{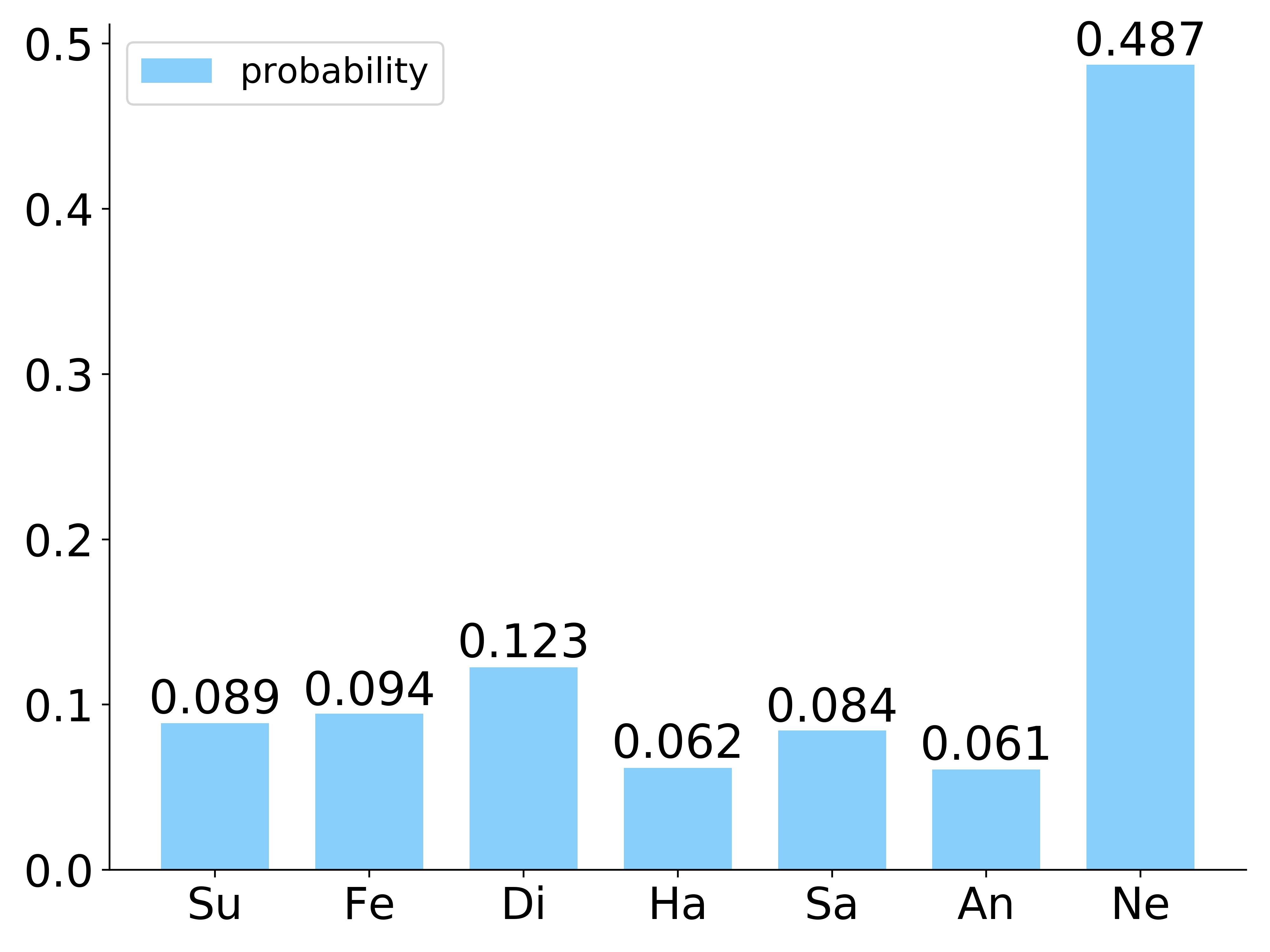}
	
	\caption{Visualization of confidences and label distributions obtained by SCE-ALD. The second row shows seven facial images from each category, and the first and third rows show label distributions corresponding to original labels and noisy labels, respectively.}
	\label{fig-vis-1}
\end{figure*}
%%%%%%%%%%%%%%%%%

As shown in Tabel \ref{tab1}, it is observed that SCE-ALD obviously improves the performance compared with the baseline and five compared methods.  
With training from scratch, SCE-ALD improves the baseline with 2.7\%, 3.4\% and 2.4\% accuracies on RAF-DB, AffectNet and FERPlus, respectively. 
For RAF-DB and AffectNet, SCE-ALD outperforms six compared methods with increasing 2.12\% and 1.53\% (ACNN), 0.29\% and 0.81\% (RAN), 0.42\% and 3\% (IPA2LT), 1.66\% and 0.96\% (LDL-ALSG), and 0.16\% and 0.08\% (SCN), respectively. 
For FERPlus, SCE-ALD is superior to compared methods with increasing 0.74\% (RAN), 1.88\% (IPA2LT) and 0.58\% (SCN), respectively. 
Furthermore, we find that SCE-ALD is slightly superior to SCN, which may be the reason that SCN also relabels facial images with low confidence similarly to ours. It indicates that amending labels of images with low confidences is effective for FER. 
In summary, experimental results demonstrates that the proposed method is more effective for improving the performance of FER by amending label distributions with semantic correlations among expressions compared with existing methods. 

%%%%%%%%%%%%%%%%%
\begin{table}
	\caption{Accuracy(\%) on three FER databases with synthetic noises.}
	\label{tab2}
	\small
	\begin{center}
		\begin{tabular}{|l|c|c|c|c|}
			\hline
			Methods           & Noise       & RAF-DB      & AffectNet    & FERPlus       \\
			\hline\hline
			Baseline          &  10\%             &  79.79      &  57.60       &   83.01         \\
			IPA2LT            &  10\%             &  80.25      &  57.42       &   83.26                \\
			SCN               &  10\%             &  82.18      &  58.58       &   84.28          \\
			SCE-ALD           &  10\%             & \bf{82.27}  &  \bf{58.97}  &  \bf{84.80}      \\
			\hline
			Baseline          &  20\%             &  73.11      &  56.29       &   81.64      \\
			IPA2LT            &  20\%             & 78.26       &   56.60      &   82.80            \\
			SCN               &  20\%             &  80.10      &  57.25       &   83.17      \\
			SCE-ALD           &  20\%             & \bf{80.38}  &  \bf{57.60}  &  \bf{83.24}  \\
			\hline
			Baseline          &  30\%             &  69.59      &  54.14       &   79.38         \\
			IPA2LT            &  30\%             &  72.26      &   54.29      &   80.95          \\
			SCN               &  30\%             &  77.46      &  55.05       &   \bf{82.47}  \\
			SCE-ALD           &  30\%             & \bf{77.61}  &  \bf{55.54}  &   82.39      \\
			\hline
		\end{tabular}
	\end{center}	
\end{table}
%%%%%%%%%%%%%%%%%

\subsection{Evaluation for Synthetic Noise Labels}    
Due to inconsistent or incorrect annotations, label noises easily occur in FER datasets.  
Based on this, we make experiments on three datasets with synthetic nosiy labels, to observe the robustness of the proposed method.   
In this experiment, we will test three level noise labels with 10\%, 20\% and 30\%, respectively, where synthetic noise labels are generated by randomly revising the given one-hot label to another category with a certain probability.
Table \ref{tab2} shows the experimental results on different synthetic noises. 

As shown in Table \ref{tab2}, it is seen that SCE-ALD is superior to Baseline, IPA2LT and SCN for RAF-DB and AffectNet, when the ratios of noisy labels are 10\%, 20\% and 30\%. 
For FERPlus, SCE-ALD is still better than others except 30\% noise. 
Higher accuracies of SCE-ALD illustrate that amending label distributions by introducing semantic correlations among expressions is more effective for FER.     
Moreover, the performance of SCE-ALD is obviously higher than Baseline for different noise ratios, which indicates label distributions effectively mitigates the problem of label noise. 
Specifically, it is also found that the proportion of noise labels is the larger in the training data, the improvement achieved by our model is the greater.  

%%%%%%%%%%%%%%%%%%%%%%%%%%%%%%%%%%%%%%%%%%%%%%%%%%%%%%%%%%%%%%%%%%%%%%%%%%%%%%%%%%%%%%%%
\begin{table*} % 4.5
	\caption{Accuracy(\%) of SCE-ALD with different $\beta$ on RAF-DB database.}
	\label{tab3}
	\small 
	\begin{center}
		\begin{tabular}{|c|c|c|c|c|c|c|c|c|c|c|}
			\hline
			{\multirow{2}*{Noises}}
			& \multicolumn{10}{c|}{Different values of $\beta$} \\
			\cline{2-11}
			\multicolumn{1}{|c|}{} & 0.1     & 0.2     & 0.3     & 0.4     & 0.5     & 0.6     & 0.7     & 0.8     & 0.9   &  1.0\\
			% \diagbox {ratio of \\ synthetic noisy label}{$\beta$}  & 0.1     & 0.2     & 0.3     & 0.4     & 0.5     & 0.6     & 0.7     & 0.8     & 0.9        \\
			\hline\hline
			0\%                                         & 71.06   & 79.63      & 85.58      & 85.96   & 86.70   & 86.96      & \bf{87.19} & 86.83   & 86.30    & 84.49      \\
			\hline
			10\%                                        & 68.45   & 80.25      & 82.13      & 81.94   & 82.27   & \bf{82.86} & 81.91      & 80.83   & 80.11    & 79.79      \\
			\hline
			20\%                                        & 67.37   & 79.27      & \bf{81.00} & 80.38   & 79.79   & 78.97      & 78.26      & 77.47   & 76.69    & 73.11      \\
			\hline
			30\%                                        & 65.16   & \bf{77.61} & 77.51      & 77.18   & 76.50   & 76.83      & 75.10      & 74.09   & 72.26    & 69.59       \\
			\hline
		\end{tabular}
	\end{center}	
\end{table*} 
%%%%%%%%%%%%%%%%%%%%%%%%%%%%%%

\subsection{Visualization of Label Distributions} 
In our model, the label distribution of each image is amended based on its confidence obtained by transferring semantic correlations among expressions into the task space. 
Hence, we exhibit some visualized results of label distributions and the confidence of facial images during the training process, to further investigate the effectiveness of our method. 
This experiment is implemented on RAF-DB, and other experimental settings are same to above.     
The visualization results are shown in Fig.~\ref{fig-vis} and Fig.~\ref{fig-vis-1}. In Figs.~\ref{fig-vis} and \ref{fig-vis-1}, the second row shows seven facial images from each category. In each image, the upper green area exhibits its original label and its confidence $\alpha$, and the lower orange area exhibits its noisy label and the corresponding confidence $\alpha$.  
The first and third rows express the obtained label distributions corresponding to original labels and noisy labels with 10\% ratio, respectively. 

From Figs.~\ref{fig-vis} and \ref{fig-vis-1}, it is found that the confidence of images with noisy labels is obviously lower than with original labels, except the third image, which dose demonstrate our model can well indentify the outliers. Based on our analyses, the reason may be that the label cannot precisely express the facial emotion for the third image.  
According to results shown in the third row, it is found that the category corresponding to the highest value in the label distribution is close to the original one-hot label, which indicates that the noisy label of each image can be  amended to a reasonable label distribution. 
For instances, 'happy' is given the highest value in the amended label distribution for the fourth image, and 'neutral' is given highest values in label distributions for the seventh image. 
It is worth noting that the obtained label distributions based on noisy labels are not inconsistent with based on original labels for seven images, but they are also subjectively reasonable, which implies that the proposed method is more effective for improving the label-side problem.

\subsection{Ablation Studies}   
In this part, we make an analysis to investigate the affect of $\beta$ in Eq.(7), and the experimental settings are same to the above experiments. Table \ref{tab3} shows all results of different $\beta =\{0.1,...,1\}$ in cases of 0\% (no noise), 10\%, 20\% and 30\% noises on RAF-DB, where the case of $\beta =1$ is equivalent to the Baseline method. 

From Fig.~\ref{fig-vis}, it is seen that the highest accuracy is given based on $\beta=0.6$ ($\beta=0.7$) when the ratio of synthetic noises is 10\% (0\%). When the ratio is 20\% (30\%), the highest accuracy is obtained based on $\beta=0.3$ ($\beta=0.2$). 
Based on these, it is found that the best performance is given based on a small value of $\beta$ when the noisy ratio is higher. In contrast, setting a big $\beta$ is more easily to obtain higher accuracy when the noisy ratio is lower, but label distribution is also essential for FER.  
Additionally, experimental results indicate that the amended label distribution of SCE-ALD is indispensable and more crucial for training a better deep network for FER, especially for the case with big label noises. 

%%%%%%%%%%%%%%----Figure 5
\begin{figure*}[t]
	\centering
	\includegraphics[width=0.95\textwidth]{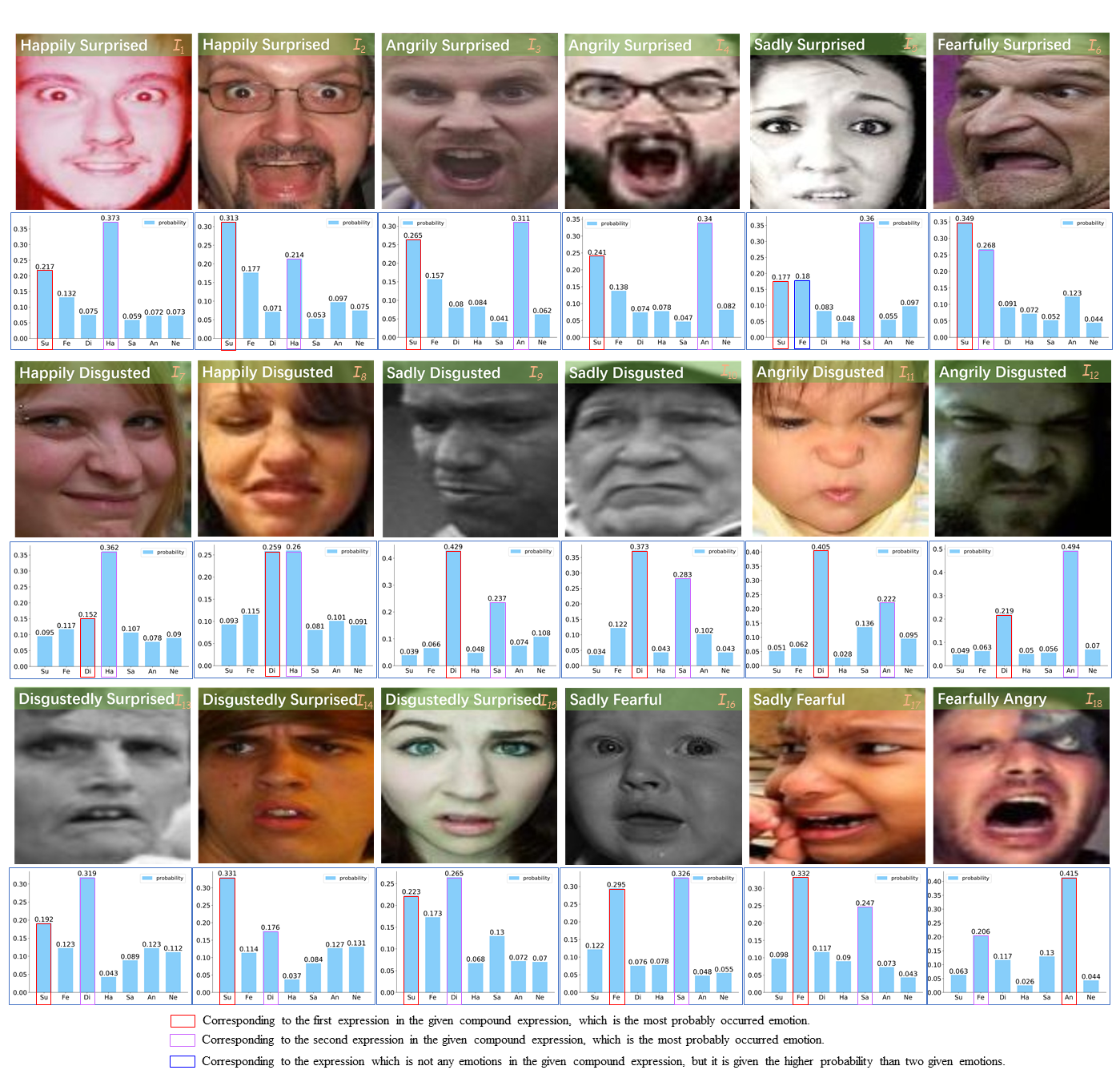}
	\caption{Visualization of the label distributions of the compound expressions images on RAF-DB, in which our model is trained with the single expression images on RAF-DB. }
	\label{fig-ldl-compound}
\end{figure*}	
%%%%%%%%%%%%%

\subsection{Results on Compound RAF-DB}
For FER, most public databases are given only with one-hot label. Thus, we mainly show the experimental results on three databases with one-hot label in above parts. Whereas, it is known that RAF-DB includes not a basic data with one-hot label but a data with compound expressions. Hence, in order to illustrate the performance of our method, we also tested SCE-ALD on compound RAF-DB and then compared the given compound labels with the label distribution obtained by our method, which is trained on the basic RAF-DB. 
In this part, all experimental settings are same to our previous experiments.  
Fig.~\ref{fig-ldl-compound} shows the visualization results of label distribution obtained by our method. 

From Fig.~\ref{fig-ldl-compound}, it is observed that the categories corresponding to two highest probabilities in the label distribution obtained by our method are consistent with the multi-label given in compound RAF-DB, which indicates that SCE-ALD can obtain a reliable label distribution of the compound expression. 

Furthermore, it is seen that two categories in mixture labels is assigned to a different probability by our method. Specially, the assigned distributions are visually reasonable for expressing different emotions in an facial images. For instances, for the $3$rd facial image ($I_3$) located in the first row, its original label is the equal mixture of happily and disgusted emotions, but the happy emotion is assigned a higher value than the disgusted emotion by our method, obviously which is more suitable than the given compound label. For the $1$th image ($I_7$) located in the second row, happily and disgusted emotions are assigned the equal values, since its emotion is too ambiguous to visually confirm which emotion is primary.  
Moreover, for the $4$st image ($I_{16}$) located on the third row, it is found that the label distribution obtained by our method more reasonably represents its compound emotion, considering that the sadly emotion is visually more obvious than the fear emotion. 
In short, all experimental results illustrate that the proposed method can effectively obtain more reliable label distributions by leveraging the intrinsic correlation among emotions. 

\section{Conclusion}\label{Conclusion}
In this paper, we propose a new FER method that adaptively amends the label distribution by leveraging the correlation among different emotions in the semantic space. 
Inspired by diverse correlations among emotions, we employ an Auto-Encoder to explore the topological information among facial expressions in the semantic space, and then a semantic class-relation graph is constructed as a prior information for FER. 
By transferring the semantic class-relation graph into the task space, the confidence of each facial image is assessed by comparing semantic and task class-relation graphs, and the prototype of categories is regenerated by boosting the feature of samples with higher confidence, and meanwhile the label distribution is amended with the prototype. 
Experimental results illustrate the proposed method is superior to compared state-of-the-art methods on three wild databases, and the visualization of label distributions and confidences of facial images demonstrates the proposed method can effectively amend the label distribution by leveraging semantic correlations among facial expressions. 

%\appendices
%\section{Proof of the First Zonklar Equation}
%Appendix one text goes here.

% you can choose not to have a title for an appendix
% if you want by leaving the argument blank
%\section{}
%Appendix two text goes here.

%% use section* for acknowledgment
%\section*{Acknowledgment}
% The authors would like to thank...

% Can use something like this to put references on a page
% by themselves when using endfloat and the captionsoff option.
\ifCLASSOPTIONcaptionsoff
  \newpage
\fi

\end{document}